\documentclass[conference]{IEEEtran}
\IEEEoverridecommandlockouts
\usepackage{cite}
\usepackage{booktabs}
\usepackage{amsmath,amssymb,amsfonts}
\usepackage{algorithmic}
\usepackage{graphicx}
\usepackage{caption}
\usepackage{subcaption}
\usepackage{xifthen,bm,tikz,tikz-cd}
\usetikzlibrary{positioning}
\usepackage{textcomp}
\usepackage{xcolor}
\def\BibTeX{{\rm B\kern-.05em{\sc i\kern-.025em b}\kern-.08em
    T\kern-.1667em\lower.7ex\hbox{E}\kern-.125emX}}
\begin{document}

\title{Concept Drift Detection from Multi-Class Imbalanced Data Streams\\
}

\author{\IEEEauthorblockN{1\textsuperscript{st} \L{}ukasz Korycki}
\IEEEauthorblockA{\textit{Department of Computer Science} \\
\textit{Virginia Commonwealth University}\\
Richmond VA, USA \\
koryckil@vcu.edu}
\and
\IEEEauthorblockN{2\textsuperscript{nd} Bartosz Krawczyk}
\IEEEauthorblockA{\textit{Department of Computer Science} \\
\textit{Virginia Commonwealth University}\\
Richmond VA, USA \\
bkrawczyk@vcu.edu}
}

\maketitle

\begin{abstract}
Continual learning from data streams is among the most important topics in contemporary machine learning. One of the biggest challenges in this domain lies in creating algorithms that can continuously adapt to arriving data. However, previously learned knowledge may become outdated, as streams evolve over time. This phenomenon is known as concept drift and must be detected to facilitate efficient adaptation of the learning model. While there exists a plethora of drift detectors, all of them assume that we are dealing with roughly balanced classes. In the case of imbalanced data streams, those detectors will be biased towards the majority classes, ignoring changes happening in the minority ones. Furthermore, class imbalance may evolve over time and classes may change their roles (majority becoming minority and vice versa). This is especially challenging in the multi-class setting, where relationships among classes become complex. In this paper, we propose a detailed taxonomy of challenges posed by concept drift in multi-class imbalanced data streams, as well as a novel trainable concept drift detector based on Restricted Boltzmann Machine. It is capable of monitoring multiple classes at once and using reconstruction error to detect changes in each of them independently. Our detector utilizes a skew-insensitive loss function that allows it to handle multiple imbalanced distributions. Due to its trainable nature, it is capable of following changes in a stream and evolving class roles, as well as it can deal with local concept drift occurring in minority classes. Extensive experimental study on multi-class drifting data streams, enriched with a detailed analysis of the impact of local drifts and changing imbalance ratios, confirms the high efficacy of our approach.
\end{abstract}


\section{Introduction}
\label{sec:int}

Continual learning from streaming data is a well-established, yet still rapidly developing field of modern machine learning \cite{Parisi:2019}. Contemporary data sources generate information characterized by both volume and velocity, thus continuously flooding learning systems. This makes traditional classification methods too slow and unable to handle the ever-changing properties of arriving instances \cite{Sahoo:2018}. Therefore, new algorithms are being developed with their efficacy and adaptive capabilities in mind \cite{Ditzler:2015}. Learning methods for streaming data must be capable of working under strictly limited memory and time consumption while offering the ability to continually incorporate new instances, and swiftly adapting to evolving data stream characteristics \cite{Krawczyk:2017}. This phenomenon is known as concept drift and may influence the properties of a stream in a multitude of ways, from changing class distributions \cite{Chandra:2018} to new features or classes emerging \cite{Wang:2019}. Timely detection of concept drift and using this information to adapt the classifier is of crucial importance to any continual learning system. 

One of the biggest challenges in learning from data streams is the non-stationary class imbalance \cite{Krawczyk:2016}. Skewed data distributions are a very challenging topic in standard machine learning, being present there for over 25 years. In the continual learning domain, class imbalance becomes even more difficult, as we not only need to deal with disproportion among classes, but also with their evolving nature \cite{Wang:2018}. Class roles and imbalance ratios are subject to change over time and we need to monitor them closely to understand which class and why poses the biggest problem for the classifier in a given moment \cite{Fernandez:2018}. When we extend this to a multi-class imbalance scenario, we get a complex and perplexing scenario that actually occurs in many real-life applications. Concept drift detection in such problems becomes extremely demanding, as we need to factor in both evolving nature of multiple classes, as well as their non-stationary skewed distributions.  

\smallskip 
\noindent \textbf{Research goal.} To propose a fully trainable drift detector that is capable of handling multi-class imbalanced data streams with special focus on evolving minority classes, and with changes appearing at both global (all classes affected) and local (some classes affected) levels.

\smallskip 
\noindent \textbf{Motivation.} While there exist a plethora of drift detectors proposed in the literature, most of them share two limitations: (i) they assume roughly balanced data distributions and thus are likely to omit concept drift happening in minority classes; and (ii) they monitor global data stream characteristics, thus detecting concept drifts that affect the entire stream, not particular classes or decision regions. This makes the state-of-the-art drift detectors unsuitable for mining imbalanced data streams, especially when multiple classes are involved. There is a need to develop a new drift detector that is skew-insensitive, can monitor multiple classes at once, and can rapidly adapt to changing imbalance ratios and classes switching roles, as none of the existing methods is capable of this.

\smallskip 
\noindent \textbf{Summary.} In this paper we propose RBM-IM, a trainable concept drift detector for continual learning from multi-class imbalanced data streams. It is realized as a Restricted Boltzmann Machine neural network with skew-insensitive modifications of the training procedure. We use it to track reconstruction error for each class independently and signal if any of them has been subject to a significant change over the most recent mini-batch of data. Our drift detector re-trains itself in an online fashion, allowing it to handle dynamically changing imbalance ratio, as well as evolving class roles (minority classes becoming majority and vice versa). RBM-IM is capable of detecting drifts occurring at both global and local levels, allowing for complex monitoring of multi-class imbalanced data streams and understanding the nature of each change that takes place.

\smallskip 
\noindent \textbf{Main contributions.} We offer the following novel contributions to the field of continual learning from data streams.

\begin{itemize}

\item \textbf{RBM-IM:} a novel and trainable concept drift detector realized as a recurrent neural network with skew-insensitive loss function that is capable of monitoring multi-class imbalanced data streams with dynamic imbalance ratio. 

\item \textbf{Robustness to class imbalance:} RBM-IM provides robustness to multi-class skewed distributions, offering excellent detection rates of drifts appearing in minority classes without being biased towards majority concepts.

\item \textbf{Detecting local and global changes:} RBM-IM is capable of detecting concept drifts affecting only a subset of minority classes (even when only a single class is affected), offering a better understanding of the nature of changes than any state-of-the-art drift detector. 

\item \textbf{Taxonomy of multi-class imbalanced data streams:} we propose a systematic view on possible challenges that can be encountered in continual learning from multi-class imbalanced data streams and formulate three scenarios that allow us to model such changes.

\item \textbf{Extensive experimental study:} we evaluate the efficacy of RBM-IM on a thoroughly designed experimental test bed using both real-world and artificial benchmarks. We introduce a novel approach towards evaluating concept drift detectors on imbalanced data streams, by measuring their reactivity to drifts occurring only in a subset of minority classes, as well as by checking their robustness to increasing imbalance ratio among multiple classes. 

\end{itemize}

	\section{Data stream mining}
	\label{sec:dsm}
	
	Data stream is defined as a sequence ${<S_1, S_2, ..., S_n,...>}$, where each element $S_j$ is a new instance. In this paper, we assume the (partially) supervised learning scenario with classification task and thus we define each instance as $S_j \sim p_j(x^1,\cdots,x^d,y) = p_j(\mathbf{x},y)$, where $p_j(\mathbf{x},y)$ is a joint distribution of the $j$-th instance, defined by a $d$-dimensional feature space and assigned to class $y$. Each instance is independent and drawn randomly from a probability distribution $\Psi_j (\mathbf{x},y)$.
	
	\smallskip
	\noindent \textbf{Concept drift.} When all instances come from the same distribution, we deal with a stationary data stream. In real-world applications, data very rarely falls under stationary assumptions \cite{Masegosa:2020}. It is more likely to evolve over time and form temporary concepts, being subject to concept drift \cite{Lu:2019}. This phenomenon affects various aspects of a data stream and thus can be analyzed from multiple perspectives. One cannot simply claim that a stream is subject to the drift. It needs to be analyzed and understood in order to be handled adequately to specific changes that occur \cite{Goldenberg:2020}. Let us now discuss the major aspects of concept drift and its characteristics. 
	
	\smallskip
	\noindent \textbf{Influence on decision boundaries}. Firstly, we need to take into account how concept drift impacts the learned decision boundaries, distinguishing between real and virtual concept drifts \cite{Oliveira:2019}. The former influences previously learned decision rules or classification boundaries, decreasing their relevance for newly incoming instances. Real drift affects posterior probabilities $p_j(y|\mathbf{x})$ and additionally may impact unconditional probability density functions. It must be tackled as soon as it appears, since it negatively impacts the underlying classifier. Virtual concept drift affects only the distribution of features $\mathbf{x}$ over time:
	
	\begin{equation}
	\widehat{p}_j(\mathbf{x}) = \sum_{y \in Y} p_j(\mathbf{x},y),
	\label{eq:cd2}
	\end{equation}
	
	\noindent where $Y$ is a set of possible values taken by $S_j$. While it seems less dangerous than real concept drift, it cannot be ignored. Despite the fact that only the values of features change, it may trigger false alarms and thus force unnecessary and costly adaptations. 
	
	\smallskip
	\noindent \textbf{Locality of changes}. It is important to distinguish between global and local concept drifts~\cite{Gama:2006}. The former affects the entire stream, while the latter affects only certain parts of it (e.g., individual clusters of instances, or subsets of classes). 
	
	\smallskip
	\noindent \textbf{Speed of changes}. Here we distinguish between sudden, gradual, and incremental concept drifts~\cite{Lu:2019}. 
	
	\begin{itemize}
		\item \textbf{Sudden concept drift} is a case when instance distribution abruptly changes with $t$-th example arriving from the stream:
		
		\begin{equation}
		p_j(\mathbf{x},y) =
		\begin{cases}
		D_0 (\mathbf{x},y),       & \quad \text{if } j < t\\
		D_1 (\mathbf{x},y),  & \quad \text{if } j \geq t.
		\end{cases}
		\label{eq:cd3}
		\end{equation}
		
		\smallskip
		\item \textbf{Incremental concept drift} is a case when we have a continuous progression from one concept to another (thus consisting of multiple intermediate concepts in between), such that the distance from the old concept is increasing, while the distance to the new concept is decreasing:
		
		\begin{equation}
		p_j(\mathbf{x},y) =
		\begin{cases}
		D_0 (\mathbf{x},y),&\text{if } j < t_1\\
		(1 - \alpha_j) D_0 (\mathbf{x},y) + \alpha_j D_1 (\mathbf{x},y),&\text{if } t_1 \leq j < t_2\\
		D_1 (\mathbf{x},y),&\text{if } t_2 \leq j
		\end{cases}
		\label{eq:cd4}
		\end{equation}
		
		\noindent where

		\begin{equation}
		\alpha_j = \frac{j - t_1}{t_2 - t_1}.
		\label{eq:cd5}
		\end{equation}
		
		\smallskip
		\item \textbf{Gradual concept drift} is a case where instances arriving from the stream oscillate between two distributions during the duration of the drift, with the old concept appearing with decreasing frequency:
		
		\begin{equation}
		p_j(\mathbf{x},y) =
		\begin{cases}
		D_0 (\mathbf{x},y),       &  \text{if } j < t_1\\
		D_0 (\mathbf{x},y),       &  \text{if } t_1 \leq j < t_2 \wedge \delta > \alpha_j\\
		D_1 (\mathbf{x},y),       &  \text{if } t_1 \leq j < t_2 \wedge \delta \leq \alpha_j\\
		D_1 (\mathbf{x},y),  &  \text{if } t_2 \leq j,
		\end{cases}
		\label{eq:cd4}
		\end{equation}
		\noindent where $\delta \in [0,1]$ is a random variable. 
	\end{itemize}
	
%	\smallskip
%	\noindent \textbf{Recurrence}. In many scenarios it is possible that a previously seen concept from $k$-th iteration may reappear D$_{j+1}$ = D$_{j-k}$ over time. One may store models specialized in previously seen concepts in order to speed up recovery rates after a known concept re-emerges \cite{Guzy:2020}. 

\section{Related works}	

\noindent \textbf{Drift detectors.} In order to be able to adapt to evolving data streams, classifiers must either have explicit information on when to update their model or use continuous learning to follow the progression of a stream. Concept drift detectors are external tools that can be paired with any classifier and used to monitor a state of the stream \cite{Barros:2018}. Usually, this is based on tracking the error of the classifier or measuring the statistical properties of data. One of the first and most popular drift detectors is Drift Detection Method (DDM) \cite{Gama:2004} which analyzes the standard deviation of errors coming from the underlying classifier. DDM assumes that the increase in error rates directly corresponds to changes in the incoming data stream and thus can be used to signal the presence of drift. This concept was extended by Early Drift Detection Method (EDDM) \cite{Garcia:2006} by replacing the standard error deviation with a distance between two consecutive errors. This makes EDDM more reactive to slower, gradual changes in the stream, at the cost of losing sensitivity to sudden drifts. Reactive Drift Detection Method (RDDM) \cite{Barros:2017} is an improvement upon DDM that allows detecting sudden and local changes under access to a reduced number of instances. RDDM offers better sensitivity than DDM by implementing a pruning mechanism for discarding outdated instances. Adaptive Windowing (ADWIN) \cite{Bifet:2007} is based on a dynamic sliding window that adjusts its size according to the size of the stable concepts in the stream. ADWIN stores two sub-windows for old and new concepts, detecting a drift when mean values in these sub-windows differ more than a given threshold. Drift Detection Methods based on the Hoeffding's bounds (HDDM) \cite{Blanco:2015} uses the identical bound as SEED, but drops the idea of sub-windows and focuses on measuring both false positive and false negative rates. Fast Hoeffding Drift Detection Method (FHDDM) \cite{Pesaranghader:2016} is yet another drift detector utilizing the popular Hoeffding's inequality, but its novelty lies in measuring the probability of correct decisions returned by the underlying classifier. Wilcoxon Rank Sum Test Drift Detector (WSTD) \cite{Barros:2018w} uses the Wilcoxon rank-sum statistical test for comparing distributions in sub-windows. 

\smallskip
\noindent \textbf{Drift detectors for imbalanced data streams.} There are almost no concept drift detectors dedicated to imbalanced data streams, especially to multi-class ones. Most of the works in this domain focus mainly on making the underlying classifier skew-insensitive, while assuming it is going to adapt on its own \cite{Cano:2020}, or erroneously using standard drift detection methods. Two main dedicated drift detectors for skewed streams are PerfSim \cite{Antwi:2012} that monitors the changes in the entire confusion matrix; and Drift Detection Method for Online Class Imbalance (DDM-OCI) \cite{Wang:2020} that monitors recall for every class.

\smallskip
\noindent \textbf{Limitations of existing methods.} Existing drift detectors suffer from two major limitations: (i) lack of self-adaptation mechanisms; and (ii) lack of robustness mechanisms. The first problem is rooted in state-of-the-art drift detectors being based on monitoring the selected properties of a stream while neglecting the fact that used monitoring criteria should also be adapted over time. The second problem is rooted in a lack of research on effective drift detectors when a stream is suffering from various data-level difficulties, such as class imbalance or noise presence. In this work, we address those two limitations with our RBM-IM, a fully trainable drift detector that autonomously adapts its detection mechanisms, while offering robustness to imbalanced class distributions.

\section{Challenges in learning from multi-class imbalanced data streams}
\label{sec:imb}

In static scenarios there is a plethora of works devoted to two-class imbalanced problems, but much less attention is paid to a much more challenging multi-class imbalanced setup \cite{Krawczyk:2016}. The same carries over to the continual learning from data streams, where most of the works focused on binary streams \cite{Krawczyk:2017}. This is highly limiting for many modern real-world applications and thus there is a need to develop skew-insensitive techniques that can handle multiple classes \cite{Saadallah:2019}. 

There is no single universal approach to how to view and analyze multi-class imbalanced data streams. Therefore, we propose a taxonomy of most crucial problems that can be encountered in this setting, creating three distinctive scenarios. They cover various learning difficulties that affect one or more classes and thus pose significant challenges for both drift detectors and classifiers. 

\begin{figure}[h]
	\centering
	\begin{subfigure}{0.33\linewidth}{
			\includegraphics[width=\linewidth,trim=2cm 2cm 2cm 2cm,clip]{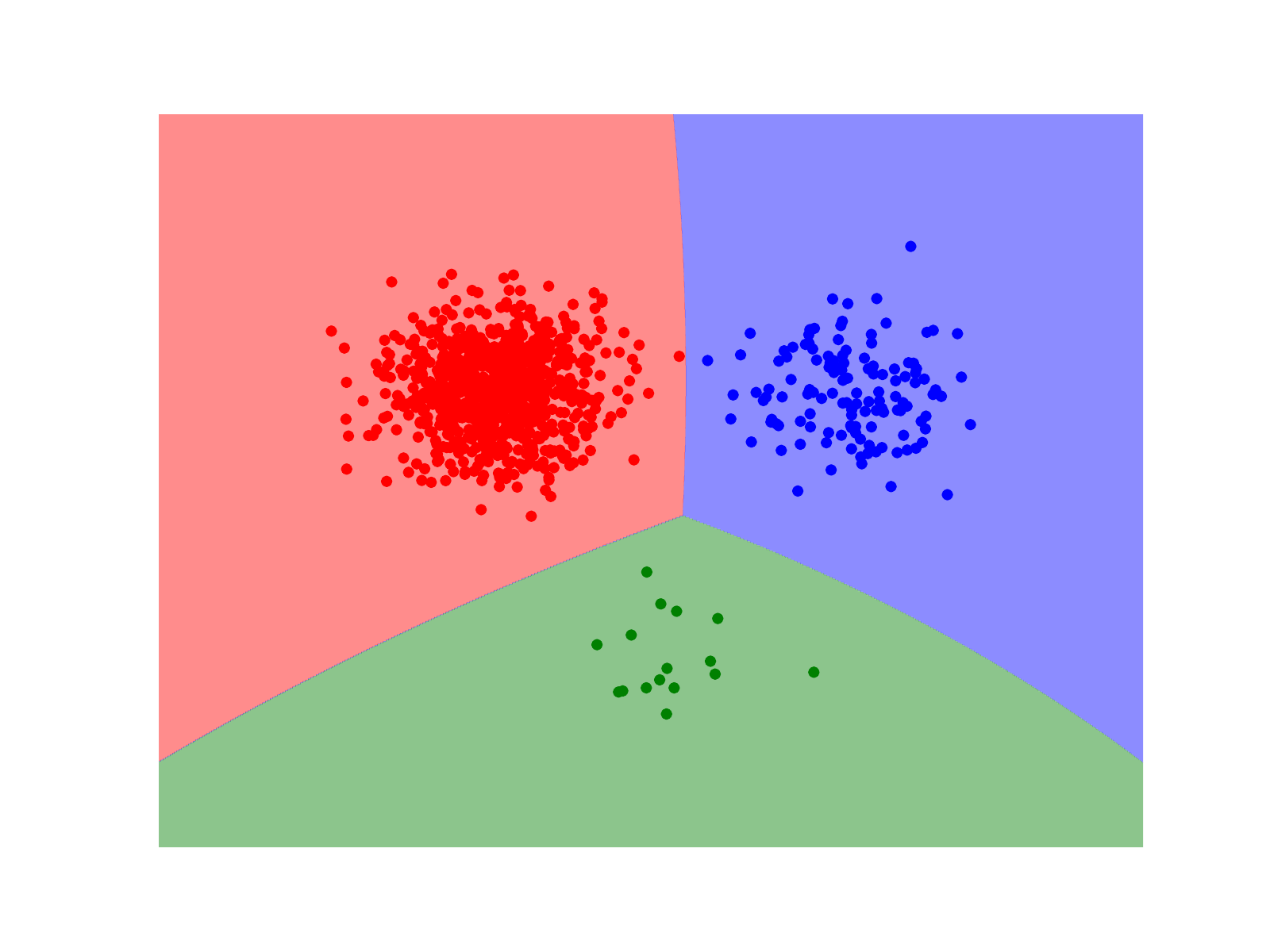}
			\subcaption{Before drift}}
	\end{subfigure}\hspace*{1pt}%
	\begin{subfigure}{0.33\linewidth}
		{\includegraphics[width=\linewidth,trim=2cm 2cm 2cm 2cm,clip]{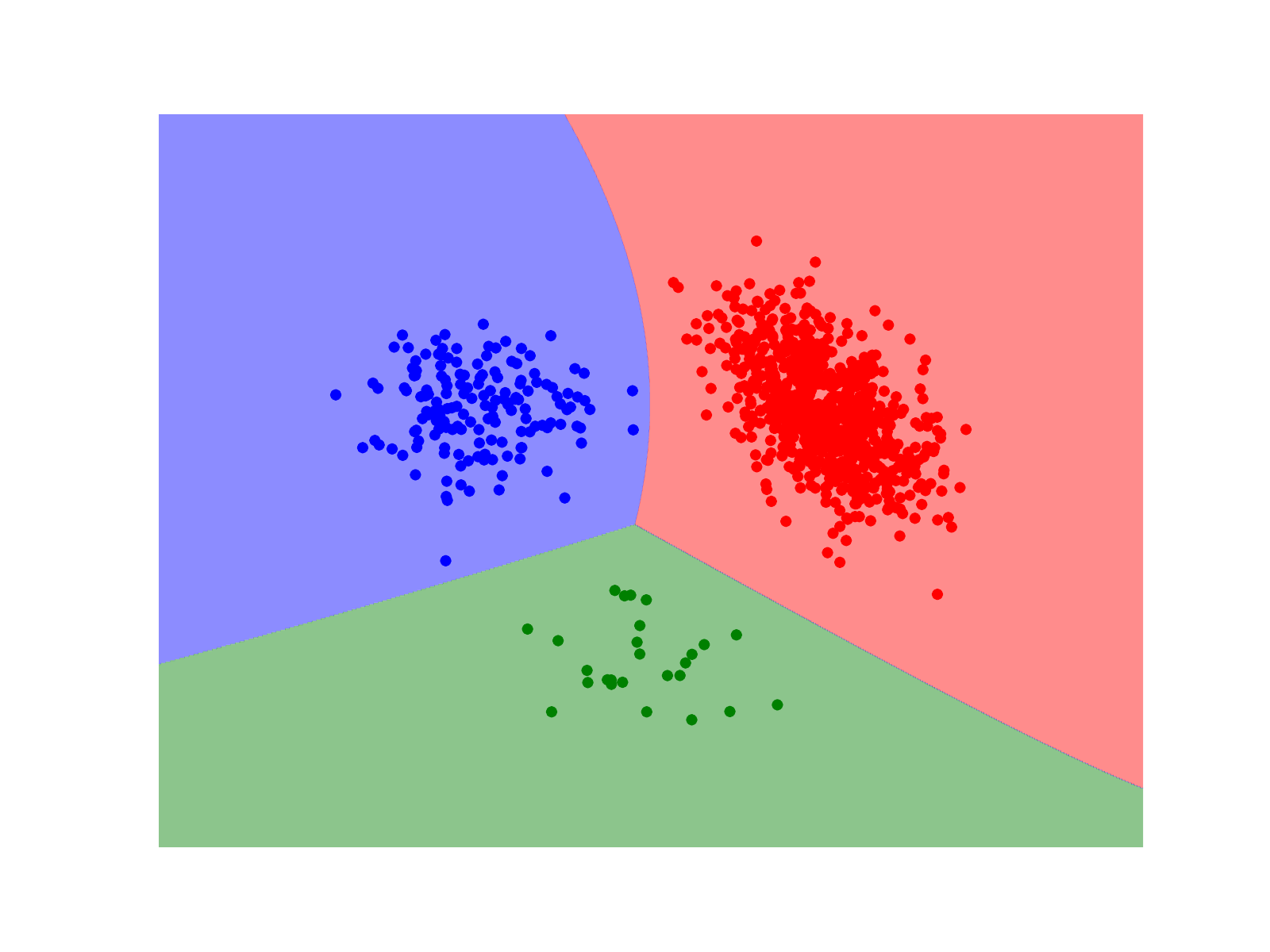}
			\subcaption{I drift}}
	\end{subfigure}\hspace*{1pt}%
	\begin{subfigure}{0.33\linewidth}
		{\includegraphics[width=\linewidth,trim=2cm 2cm 2cm 2cm,clip]{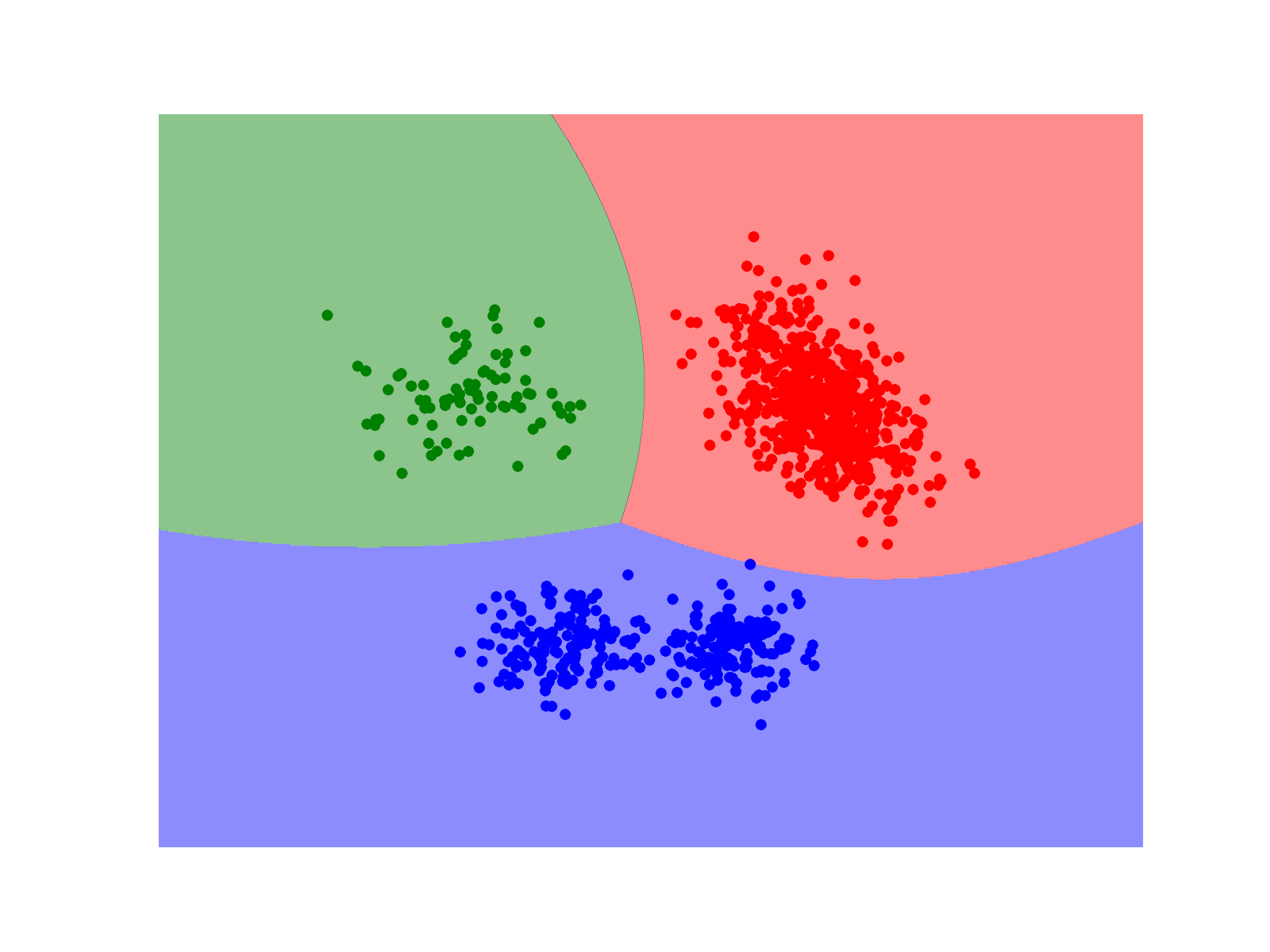}
			\subcaption{II drift}}
	\end{subfigure}\vspace*{3pt}\\
	\caption{Scenario 1 -- global concept drift and dynamic imbalance ratio.}
	\label{fig:vis1}
\end{figure}

\smallskip
\noindent \textbf{Scenario 1: Global concept drift and dynamic imbalance ratio.} Here we assume that all classes are subject to a real concept drift that will influence the decision boundaries. Additionally, imbalance ratio among the classes changes together with the drift occurrences. However, class roles remain static and classes denoted as minority stay minority during the entire stream processing. This scenario poses challenges to drift detectors by varying the degree of changes in each class and how they actually impact the decision boundaries. Changes in minority classes may get overlooked by detector bias towards the majority ones, as usually they gather statistics over the entire data stream. This is depicted in Fig.~\ref{fig:vis1}.

\begin{figure}[h]
	\centering
	\begin{subfigure}{0.33\linewidth}{
			\includegraphics[width=\linewidth,trim=2cm 2cm 2cm 2cm,clip]{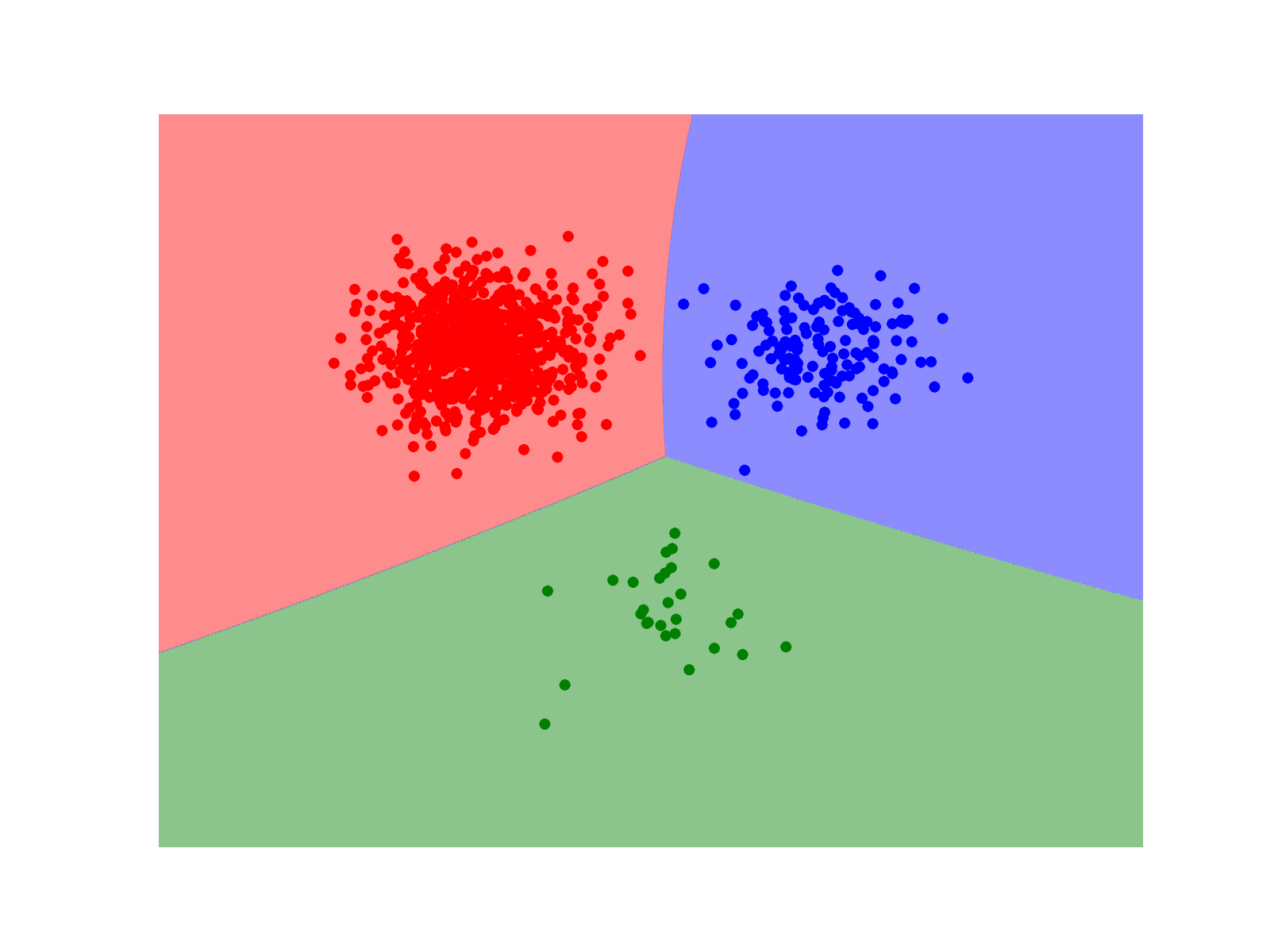}
			\subcaption{Before drift}}
	\end{subfigure}\hspace*{1pt}%
	\begin{subfigure}{0.33\linewidth}
		{\includegraphics[width=\linewidth,trim=2cm 2cm 2cm 2cm,clip]{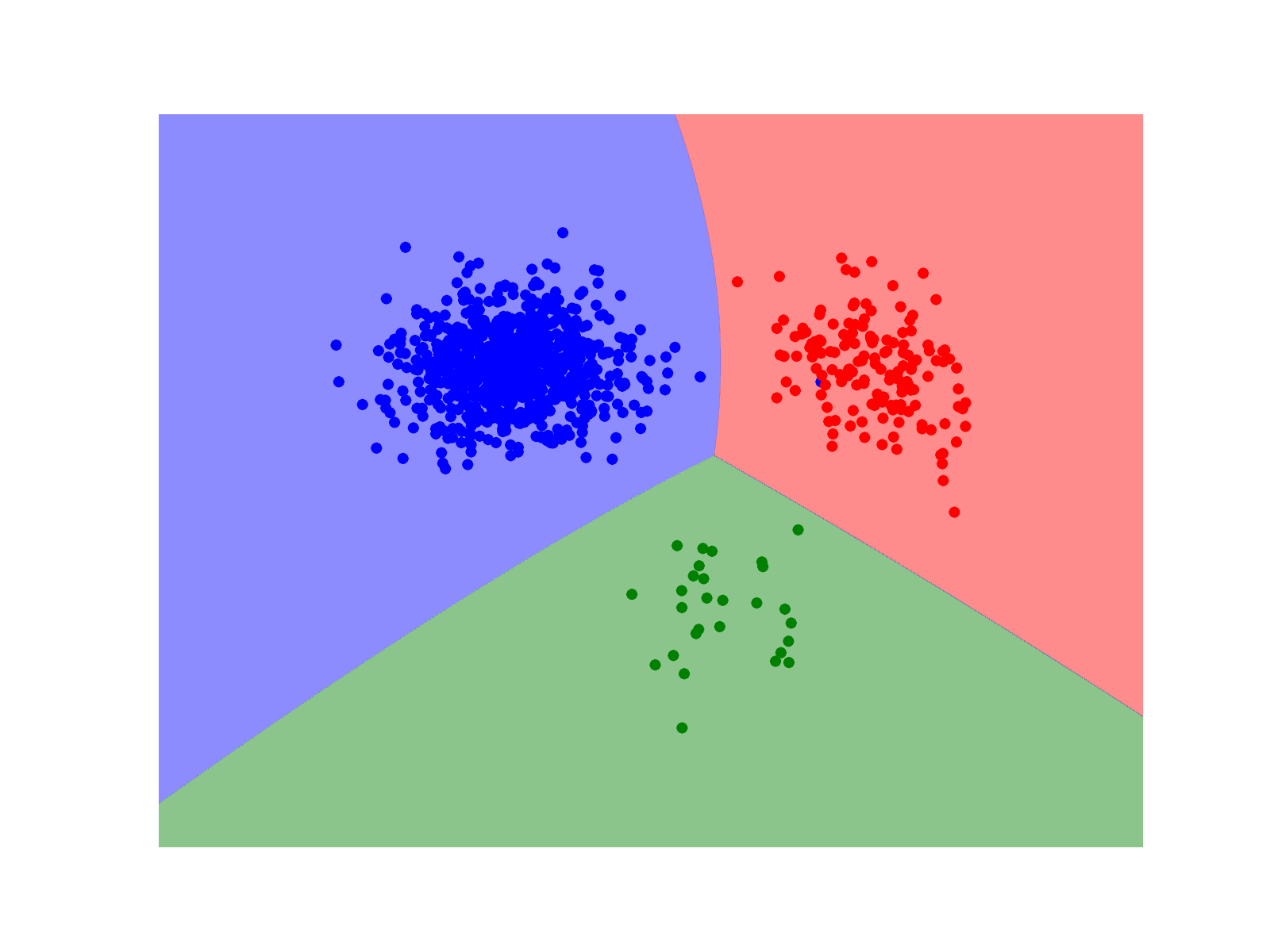}
			\subcaption{I drift}}
	\end{subfigure}\hspace*{1pt}%
	\begin{subfigure}{0.33\linewidth}
		{\includegraphics[width=\linewidth,trim=2cm 2cm 2cm 2cm,clip]{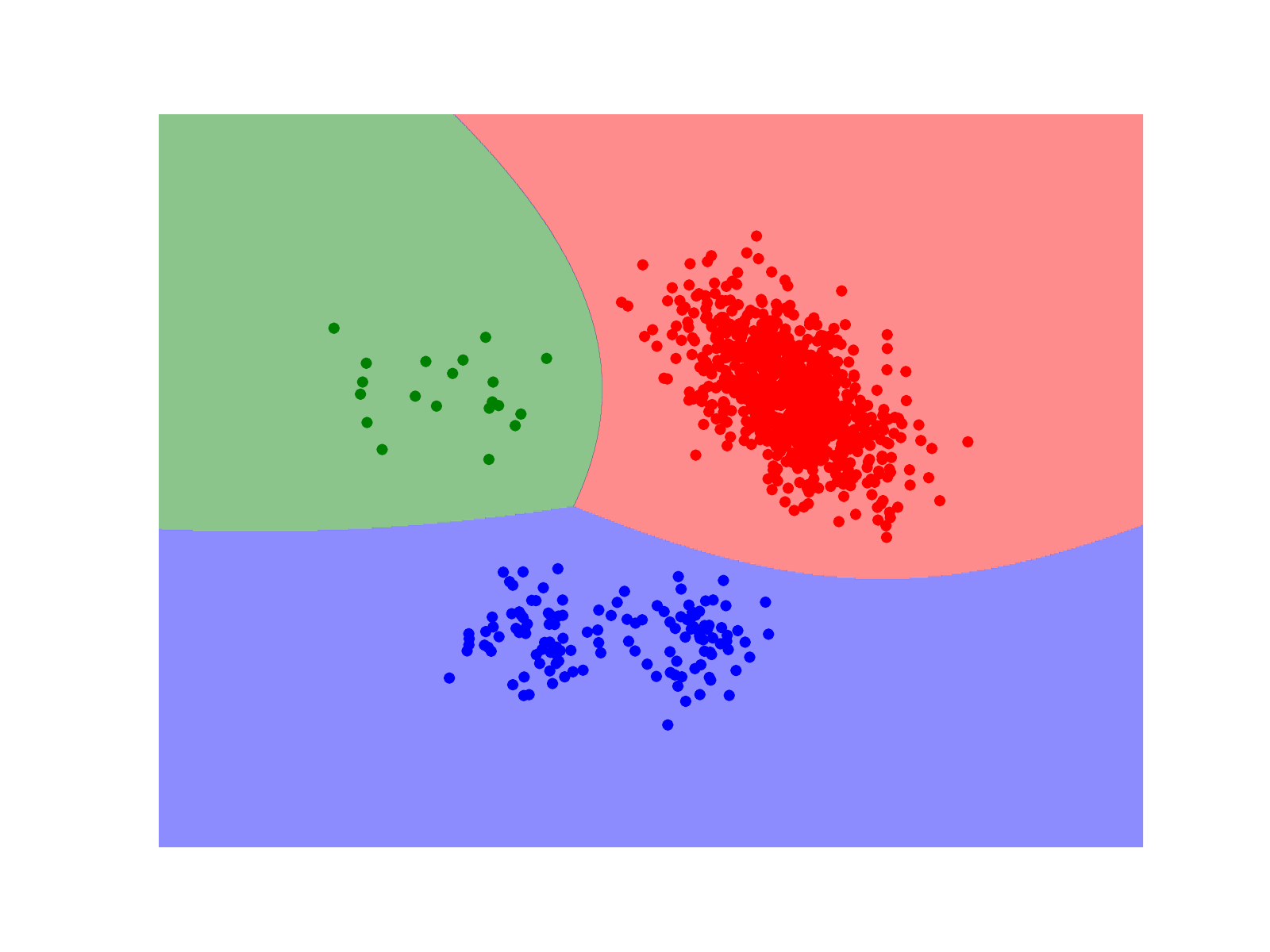}
			\subcaption{II drift}}
	\end{subfigure}\vspace*{3pt}\\
	\caption{Scenario 2 -- global concept drift, dynamic imbalance ratio, and changing class roles.}
	\label{fig:vis2}
\end{figure}

\smallskip
\noindent \textbf{Scenario 2: Global concept drift, dynamic imbalance ratio, and changing class roles.} Here we extend Scenario 1 by adding the third learning difficulty -- changing class roles. Now the imbalance ratio is subject to more significant changes and as a result classes may switch roles -- minority may become majority and vice versa. This is especially challenging to track in a multi-class case, where relationships among classes are more complex. Drift detectors have difficulties with keeping any reliable statistics coming from classes that rapidly change their roles. This may lead to frequently switching bias towards whichever class is currently the most frequent one. This is depicted in Fig.~\ref{fig:vis2}.

\begin{figure}[h]
	\centering
	\begin{subfigure}{0.33\linewidth}{
			\includegraphics[width=\linewidth,trim=2cm 2cm 2cm 2cm,clip]{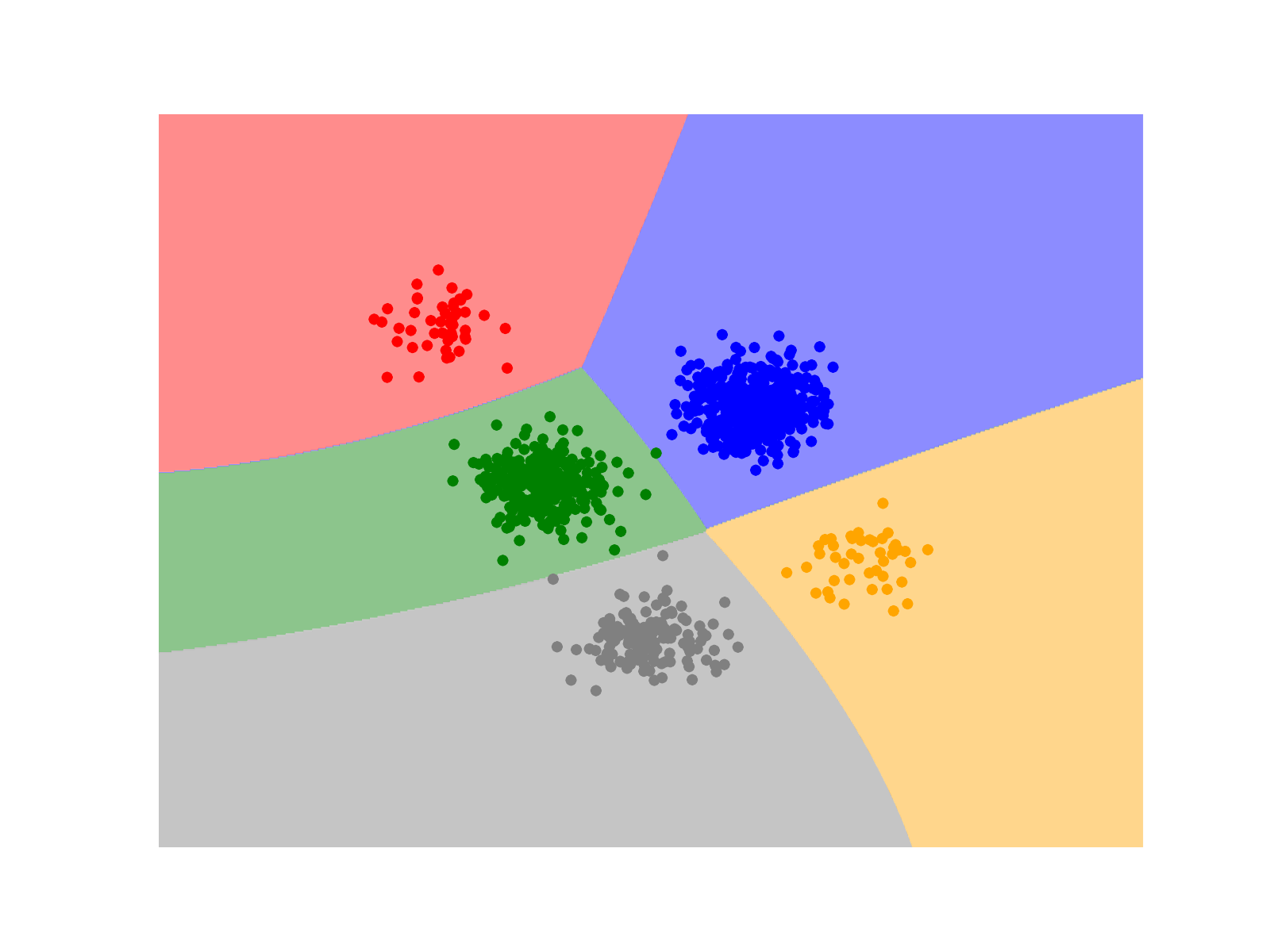}
			\subcaption{Before drift}}
	\end{subfigure}\hspace*{1pt}%
	\begin{subfigure}{0.33\linewidth}
		{\includegraphics[width=\linewidth,trim=2cm 2cm 2cm 2cm,clip]{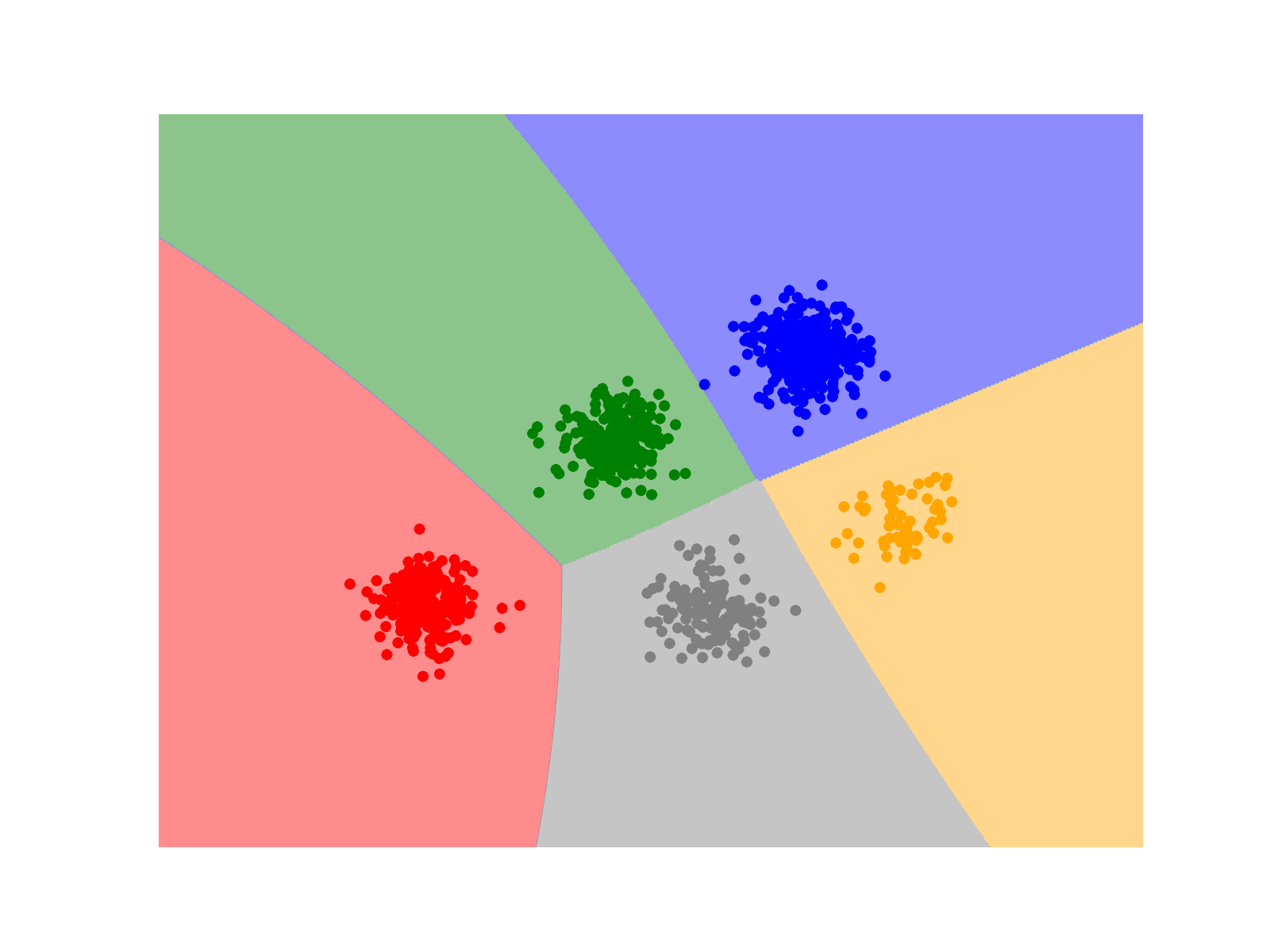}
			\subcaption{I drift}}
	\end{subfigure}\hspace*{1pt}%
	\begin{subfigure}{0.33\linewidth}
		{\includegraphics[width=\linewidth,trim=2cm 2cm 2cm 2cm,clip]{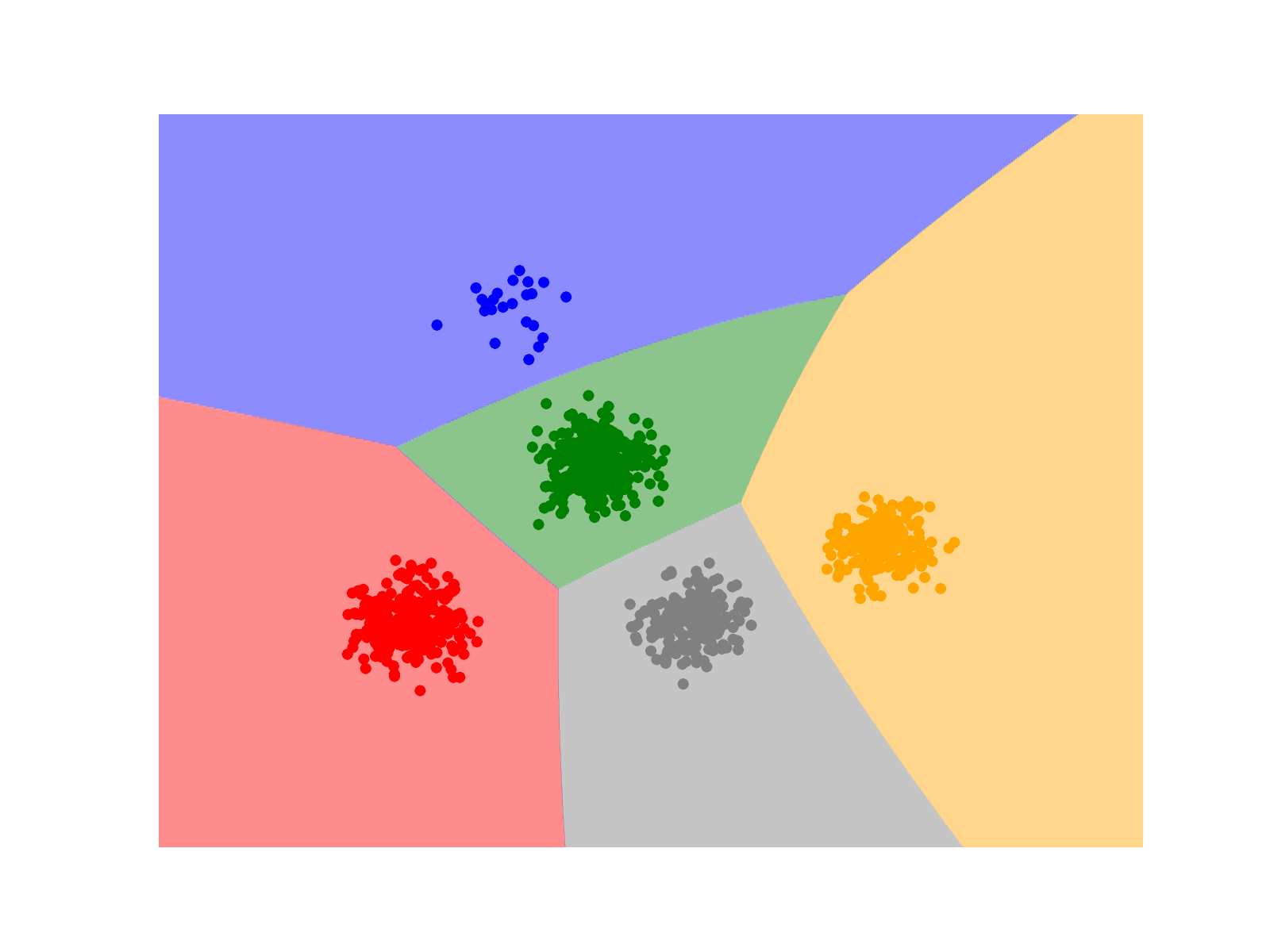}
			\subcaption{II drift}}
	\end{subfigure}\vspace*{3pt}%
	\caption{Scenario 3 -- local concept drift, dynamic imbalance ratio, and changing class roles.}
	\label{fig:vis3}
\end{figure}

\smallskip
\noindent \textbf{Scenario 3: Local concept drift, dynamic imbalance ratio, and changing class roles.} This is the most challenging scenario that retains dynamic imbalance ratio and changing class roles from Scenario 2, but moves from global concept drift to local one. That means in a given moment only a subset of classes (or even a single one) may be affected by a real concept drift, while the remaining ones are subject to no changes or a virtual concept drift that does not impact decision boundaries (see Sec. 2). In such a setting we should not only be able to tell if drift takes place but also which classes are affected. It is a big step towards understanding the dynamics of concept drift and offering classifier adaptation to specific regions of decision space (leading to savings in time and computational resources). This is the most challenging scenario for concept drift detectors, as changes happening in minority classes will remain unnoticed when a detector is biased towards the majority class. This is depicted in Fig.~\ref{fig:vis3}.

\smallskip
\noindent \textbf{Real-world problems affected by multi-class imbalance and concept drift.} The three defined scenarios are not only interesting from the theoretical point of view but also directly transfer to a plethora of real-world applications. In cybersecurity, we deal with multiple types of attacks that appear with varying frequencies (multi-class extremely imbalanced problems). Some of those attacks will dynamically change over time to bypass new security settings, while legal transactions will not be affected by such concept drift. In computer vision, target detection focuses on finding few specific targets, differentiating them from the information coming from a much bigger background. Targets may change their nature over time, being subject to variations, or even camouflage. In natural language processing, we must deal with constantly evolving wording/slang utilized by various minority groups, where changes in those groups will happen independently.

\section{Restricted Boltzmann Machine for imbalanced drift detection}
	\label{sec:rrb}

	\smallskip
	\noindent \textbf{Overview of the proposed method.} We introduce a novel concept drift detector for multi-class imbalanced data streams, implemented as a Restricted Boltzmann Machine (RBM-IM) with leveraged robustness to skewed distributions via dedicated loss function. It is a fully trainable drift detector, capable of autonomous adaptation to the current state of a stream, imbalance ratios, and class roles, without relying on user-defined thresholds. 
	
	\subsection{Skew-insensitive Restricted Boltzmann Machine}
	\label{sec:rbm}
	
	\noindent \textbf{RBM-IM neural network architecture.}  Restricted Boltzmann Machines (RBMs) are generative two-layered neural networks \cite{Ramasamy:2020} constructed using the $\mathbf{v}$ layer of $V$ visible neurons and the $\mathbf{h}$ layer of $H$ hidden neurons:
	
	\begin{equation}
	\begin{split}
	\mathbf{v} = [v_1,\cdots, v_V] \in \{0,1\}^V,\\
	\mathbf{h} = [h_1,\cdots, h_H] \in \{0,1\}^H
	\end{split} 
	\label{eq:rbm1}
	\end{equation}
	
	We deal with supervised continual learning from data streams (as defined in Sec.~2), thus we need to extend this two-layer RBM architecture with the third $\mathbf{z}$ layer for class representation. It is implemented as a continuous encoding, meaning that each neuron in $\mathbf{z}$ will return its real-valued support for each analyzed class (thus being responsible for the classification process). By $\mathbf{m}_z$ we denote the vector of RBM outputs with support returned by the $z$-th neuron for the $m$-th class. This allows to define $\mathbf{z}$, known also as the class layer or the softmax layer:
	
	\begin{equation}
	\mathbf{z} = [z_1, \cdots, z_Z] \in {\mathbf{m}_1,\cdots, \mathbf{m}_Z}.
	\end{equation}
	
	\noindent This class layer uses the softmax function to estimate the probabilities of activation of each neuron in $\mathbf{z}$. 
	
	RBMs do not have connections between units in the same layer, which holds for $\mathbf{v}$, $\mathbf{h}$, and $\mathbf{z}$. Neurons in the visible layer $\mathbf{v}$ are connected with neurons in the hidden layer $\mathbf{h}$, and neurons in $\mathbf{h}$ are connected with those in the class layer $\mathbf{z}$. The weight assigned to a connection between the $i$-th visible neuron $v_i$ and the $j$-th hidden neuron $h_j$ is denoted as $w_{ij}$, while the weight assigned to a connection between the $j$-th hidden neuron $h_j$ and the $k$-th class neuron $z_k$ is denoted as $u_{jk}$. This is used to define the RBM energy function:
	
	\begin{equation}
	\begin{split}
	E(\mathbf{v},\mathbf{h},\mathbf{z}) = - \sum_{i=1}^V v_i a_i - \sum_{j=1}^H h_j b_j - \sum_{k=1}^Z z_k c_k \\
- \sum_{i=1}^V \sum_{j=1}^H v_i h_j w_{ij} - \sum_{j=1}^H \sum_{k=1}^Z  h_j z_k u_{jk},
	\end{split}
\label{eq:rbm2}
	\end{equation}
	
	\noindent where $a_i, b_j,$ and $c_k$ are biases introduced to $\mathbf{v}, \mathbf{h}$, and $\mathbf{z}$ respectively. Energy formula $E(\cdot)$ for state $[\mathbf{v},\mathbf{h},\mathbf{z}]$ is used to calculate the probability of RBM of being in a given state (i.e., assuming certain weight values), using the Boltzmann distribution:
	
	\begin{equation}
	P(\mathbf{v},\mathbf{h},\mathbf{z}) = \frac{\exp \left( -E(\mathbf{v},\mathbf{h},\mathbf{z})\right)}{F},
	\label{eq:rbm3}
	\end{equation}
	
	\noindent where $F$ is a partition function allowing to normalize the probability $P(\mathbf{v},\mathbf{h},\mathbf{z})$ to 1.
	
 Hidden neurons in $\mathbf{h}$ are independent and use features given by the visible layer $\mathbf{v}$. The activation probability of the $j$-th given neuron $h_j$ can be calculated as follows:
	
	\begin{equation}
	\begin{split}
	P(h_j |\mathbf{v},\mathbf{z}) = \frac{1}{1 + \exp\left(-b_j - \sum_{i=1}^V v_i w_{ij} - \sum_{k=1}^Z z_k u_{jk}\right)} \\ 
	= \sigma \left( b_j + \sum_{i=1}^V v_i w_{ij} + \sum_{k=1}^Z z_k u_{jk}\right),
	\end{split} 
	\label{eq:rbm4}
	\end{equation}
	
	\noindent where $\sigma(\cdot) = 1/(1+\exp(- \cdot))$ stands for a sigmoid function.
	
	The same assumption may be made for neurons in the visible layer $\mathbf{v}$, when values of neurons in the hidden layer $\mathbf{h}$ are known. This allows us to calculate the activation probability of the $i$-th visible neuron as:
	
	\begin{equation}
	\begin{split}
	P(v_i |\mathbf{h}) = \frac{1}{1 + \exp\left(-a_i - \sum_{j=1}^H h_j w_{ij}\right)} \\
= \sigma \left( a_i + \sum_{j=1}^H h_j w_{ij} \right),
\end{split}
	\label{eq:rbm5}
	\end{equation}
	
	\noindent where one must note that given $\mathbf{h}$, the activation probability of neurons in $\mathbf{v}$ does not depend on $\mathbf{z}$. The activation probability of class layer (i.e., decision which class the object should be assigned to) is calculated using the softmax function:
	
	\begin{equation}
	P(\mathbf{z} = \mathbf{1}_k |\mathbf{h}) = \frac{\exp \left( - c_k - \sum_{j=1}^H h_j u_{jk}\right)}{\sum_{l=1}^Z \exp\left( - c_l - \sum_{j=1}^H h_j u_{jl}\right)},
	\label{eq:rbm6}
	\end{equation}
	
	\noindent where $k \in [1, \cdots, Z]$ and $k \neq l$.
	
	\smallskip
	\noindent \textbf{RBM training procedure.} As RBM is a neural network model, we may train it using a loss function $L(\cdot)$ minimization with any gradient descent method. Standard RBM most commonly uses the negative log-likelihood of both external layers $\mathbf{v}$ and $\mathbf{z}$. However, our RBM-IM architecture must be designed to handle multiple imbalanced classes. Therefore, we need to modify this loss function to make RBM-IM skew-insensitive. We will achieve this by using the effective number of samples approach \cite{Cui:2019} that measures the contributions of instances in each class. This allows us to formulate a class-balanced negative log-likelihood loss for RBM-IM:
	
	\begin{equation}
	L(\mathbf{v},\mathbf{z}) = - \frac{1 - \beta}{1 - \beta^{x}_{m}}\log\left( P(\mathbf{v},\mathbf{z}) \right),
	\label{eq:rbm7}
	\end{equation}
	
	\noindent where $\beta^{x}_{m}$ stands for the contribution of $x$-th instance to the $m$-th class. By taking each independent weight $w_{ij}$, we may now calculate the gradient of the loss function:
	
	\begin{equation}
	\begin{split}
	\nabla L(w_{ij}) = \frac{\delta L(\mathbf{v},\mathbf{z})}{\delta w_{ij}} = \sum_{\mathbf{v},\mathbf{h},\mathbf{z}} P(\mathbf{v},\mathbf{h},\mathbf{z}) v_i h_j \\
- \sum_{\mathbf{h}} P(\mathbf{h}|\mathbf{v},\mathbf{z})  v_i h_j.
\end{split}
	\label{eq:rbm8}
	\end{equation}
	
	\noindent This equation allows us to calculate the loss function gradient for a single instance. However, as we use RBM as a drift detector, we must be able to capture the evolving properties of a data stream. If we based our change detection on variations induced by a single new instance, we would be highly sensitive to even the smallest noise ratio. Therefore, our RBM-based drift detector must be able to work with a batch of the most recent instances in order to capture the current stream characteristics. We propose to define RBM-IM model for learning on mini-batches of instances. This will offer significant speed-up when compared to traditional batch learning used in data streams. For a mini-batch of $n$ instances arriving in $t$ time $\mathbf{M}_t = {x_1^t,\cdots, x_n^t}$, we can rewrite the gradient from Eq.~\ref{eq:rbm8} using expected values with loss function:
	
	\begin{equation}
	\frac{\delta L(\mathbf{M}_t)}{\delta w_{ij}} = E_{\text{model}}[v_i h_j] - E_{\text{data}}[v_i h_j],
	\label{eq:rbm9}
	\end{equation}
	
	\noindent where $E_{\text{data}}$ is the expected value over the current mini-batch of instances and $E_{\text{model}}$ is the expected value from the current state of RBM-IM. Of course, we cannot trace directly the value of $E_{\text{model}}$ (as this would require an immediate oracle access to ground truth), therefore we must approximate it using Contrastive Divergence with $k$ Gibbs sampling steps to reconstruct the input data (CD-$k$):
	
	\begin{equation}
	\frac{\delta L(\mathbf{M}_t)}{\delta w_{ij}} \approx E_{\text{recon}}[v_i h_j] - E_{\text{data}}[v_i h_j].
	\label{eq:rbm10}
	\end{equation}
	
	After processing the $t$-th mini-batch $\mathbf{M}_t$, we can update the wights in RBM-IM using any gradient descent method as follows:
	
	\begin{equation}
	w_{ij}^{t+1} = w_{ij}^{t} - \eta \left( E_{\text{recon}}[v_i h_j] - E_{\text{data}}[v_i h_j]\right),
	\label{eq:rbm11}
	\end{equation}
	
	\noindent where $\eta$ stands for the learning rate of the RBM-IM neural network (responsible for the speed of model update and forgetting of old information). The way to update the $a_i$, $b_j$, and $c_k$ biases, as well as weights $u_{jk}$ is analogous to Eq.~\ref{eq:rbm11} and can be expressed as:
	
	\begin{equation}
	a_{i}^{t+1} = a_{i}^{t} - \eta \left( E_{\text{recon}}[v_i] - E_{\text{data}}[v_i]\right),
	\label{eq:rbm12}
	\end{equation}
	
	\begin{equation}
	b_{j}^{t+1} = b_{j}^{t} - \eta \left( E_{\text{recon}}[h_j] - E_{\text{data}}[h_j]\right),
	\label{eq:rbm13}
	\end{equation}
	
	\begin{equation}
	c_{k}^{t+1} = c_{k}^{t} - \eta \left( E_{\text{recon}}[z_k] - E_{\text{data}}[z_k]\right),
	\label{eq:rbm14}
	\end{equation}
	
	\begin{equation}
	u_{jk}^{t+1} = u_{jk}^{t} - \eta \left( E_{\text{recon}}[h_j z_k] - E_{\text{data}}[h_j z_k]\right).
	\label{eq:rbm15}
	\end{equation}
	
	\subsection{Drift detection with RBM-IM}
	\label{sec:drd}
	
	While RBM-IM is a skew-insensitive generative neural network model, we can use it as an explicit drift detector. The RBM-IM model stores compressed characteristics of the distribution of data it was trained on. By using any similarity measure between the data prototypes and properties of newly arrived instances, one may evaluate if there are any changes in the distribution. This allows us to use RBM-IM as a drift detector. Our model uses an embedded similarity measure for monitoring the state of a stream and the level to which the newly arrived instances differ from the previously observed concepts. RBM-IM tracks the similarity measure for every single class independently, using the class layer continuous outputs. RBM-IM is a fully trainable and self-adaptive drift detector, capable not only of capturing the trends of changes in each class independently (versus state-of-the-art drift detectors that monitor changes in all classes with an aggregated measure), but also of learning and adapting to the current state of a stream, class imbalance ratios, and class roles. This makes it a highly attractive approach for handling multi-class imbalanced streams with various learning difficulties discussed in Sec.~4.
	
	\smallskip
	\noindent \textbf{Measuring data similarity}. In order to evaluate the similarity of newly arrived instances to old concepts stored in RBM-IM, we will use the reconstruction error metric. We can calculate it online for each new instance, by inputting a newly arrived $d$-dimensional instance $S_n = [x_1^n, \cdots, x_d^n, y^n]$ to the $\mathbf{v}$ layer of RBM. Then values of neurons in $\mathbf{v}$ are calculated to reconstruct the feature values. Finally, class layer $\mathbf{z}$ is activated and used to reconstruct the class label. This allows us to keep track of the reconstruction error for each class independently, offering per-class drift detection capabilities. We can denote the reconstructed vector for $m$-th class as:
	
	\begin{equation}
	\tilde{S}_n^m = [\tilde{x}_1^n, \cdots, \tilde{x}_d^n, \tilde{y}_1^n, \cdots, \tilde{y}_Z^n],
	\label{eq:rbm16}
	\end{equation}
	
	\noindent where the reconstructed vector features and labels are taken from probabilities calculated using the hidden layer:
	
	\begin{equation}
	\tilde{x}_i^n = P (v_i|h),
	\label{eq:rbm17}
	\end{equation}
	
	\begin{equation}
	\tilde{y}_k^n = P (z_k|h).
	\label{eq:rbm18}
	\end{equation}
	
	\noindent The $\mathbf{h}$ layer is taken from the conditional probability, in which the $\mathbf{v}$ layer is identical to the input instance:
	
	\begin{equation}
	\mathbf{h} \sim P(\mathbf{h}|\mathbf{v} = x^n, \mathbf{z} = \mathbf{1}_{y_n}).
	\label{eq:rbm19}
	\end{equation}
	
	\noindent This allows us to write the reconstruction error in a form of the mean squared error between the true and reconstructed instance for the $m$-th class:
	
	\begin{equation}
	R(S_n^m) = \sqrt{\sum_{i=1}^d (x_i^n - \tilde{x}_i^n)^2 + \sum_{k=1}^Z(\mathbf{1}^{y_n}_k - \tilde{y}_k^n)^2}.
	\label{eq:rbm19}
	\end{equation}
	
	For the purpose of obtaining a stable concept drift detector, we do not look for a change in distribution over a single instance, but for the change over the newly arriving mini-batch of instances. Therefore, we need to calculate the average reconstruction error over the recent mini-batch of data for the $m$-th class:
	
	\begin{equation}
	R(\mathbf{M}_t^m) = \frac{1}{n} \sum_{m=1}^n R(x_m^t).
	\label{eq:rbm20}
	\end{equation}
	
	\smallskip
	\noindent \textbf{Adapting reconstruction error to drift detection.} In order to make the reconstruction error a practical measure for detecting the presence of concept drift, we propose to measure the evolution of this measure (i.e., its trends) over arriving mini-batches of instances. The analysis of the trends is done for each class independently, allowing us to effectively detect local concept drifts. We achieve this by using the well-known sliding window technique that will move over the arriving mini-batches. Let us denote the trend of reconstruction error for the $m$-th class over time as $Q_r(t)^m$ and calculate it using the following equation:
	
	\begin{equation}
	Q_r(t)^m = \frac{\bar{n^m}_t \bar{TR}_t - \bar{T}_t \bar{R}_t}{\bar{n}_t \bar{T^2}_t - (\bar{T}_t)^2 }.
	\label{eq:rbm21}
	\end{equation}
	
	\noindent The trend over time can be computed using a simple linear regression, with the terms in Eq.~\ref{eq:rbm21} being simply sums over time as follows:
	
	\begin{equation}
	\bar{TR}_t = \bar{TR}_{t-1} + t R(\mathbf{M}_t^m),
	\label{eq:rbm22}
	\end{equation}
	
	\begin{equation}
	\bar{T}_t = \bar{T}_{t-1} + t,
	\label{eq:rbm23}
	\end{equation}
	
	\begin{equation}
	\bar{R}_t = \bar{R}_{t-1} + R(\mathbf{M}_t^m),
	\label{eq:rbm24}
	\end{equation}
	
	\begin{equation}
	\bar{T^2}_t = \bar{T^2}_{t-1} + t^2,
	\label{eq:rbm25}
	\end{equation}
	
	\noindent where $\bar{TR}_0 = 0$, $\bar{T}_0 = 0$, $\bar{R}_0 = 0$, and $\bar{T^2}_0 = 0$. We capture those statistics for each class using a sliding window of size $W$. Instead of using a manually set size, which is inefficient for drifting data streams, we propose to use a self-adaptive window size \cite{Bifet:2007}. This eliminates the need for manual tuning of the window size that is used for drift detection. To allow flexible learning from various sizes of mini-batches, we must consider a case where $t > W$. Here, we must compute the terms for the trend regression using the following equations:
	
	\begin{equation}
	\bar{TR}_t = \bar{TR}_{t-1} + t R(\mathbf{M}_t) - (t - w)R(\mathbf{M}_{t-W}^m),
	\label{eq:rbm26}
	\end{equation}
	
	\begin{equation}
	\bar{T}_t = \bar{T}_{t-1} + t - (t - W),
	\label{eq:rbm27}
	\end{equation}
	
	\begin{equation}
	\bar{R}_t = \bar{R}_{t-1} + R(\mathbf{M}_t) - R(\mathbf{M}_{t-W}^m),
	\label{eq:rbm28}
	\end{equation}
	
	\begin{equation}
	\bar{T^2}_t = \bar{T^2}_{t-1} + t^2 - (t - W)^2.
	\label{eq:rbm29}
	\end{equation}
	
	\noindent The required number of instances $\bar{n}_t^m$ to compute the trend of $Q_r(t)^m$ for $m$-th class as time $t$ is given as follows:
	
	\begin{equation}
	\bar{n}_t = 
	\begin{cases}
	t & \quad \text{if } t \leq W\\
	W & \quad \text{if } t > W.
	\end{cases}	
	\label{eq:rbm30}
	\end{equation}

\noindent \textbf{Drift detection.} The above Eq.~\ref{eq:rbm21} allows us to compute the trends for every analyzed mini-batch of data. In order to detect the presence of drift we need to have capability of checking if the new mini-batch differs significantly from the previous one for each analyzed class. Our RBM-IM uses Granger causality test \cite{Sun:2008} on trends from subsequent mini-batches of data for each class $Q_r(\mathbf{M}_t^m)$ and $Q_r(\mathbf{M}_{t+1}^m)$. This is a statistical test that determines whether one trend is useful in forecasting another. As we deal with non-stationary processes we perform the variation of Granger causality test based on first differences \cite{Mahjoub:2020}. Accepted hypothesis means that it is assumed that there exist Granger causality relationship between $Q_r(\mathbf{M}_t^m)$ and $Q_r(\mathbf{M}_{t+1}^m)$, which means there is no concept drift on the $m$-th class. If the hypothesis is rejected, RBM-IM signals the presence of concept drift on the $m$-th class. 

\section{Experimental study}
\label{sec:exp}

In this section we present the experimental study used to evaluate the quality of RBM-IM. It was carefully designed to offer an in-depth analysis of the proposed method and gain insights into its behavior in various multi-class imbalanced data stream scenarios. We tailored this study to answer the following research questions.

\begin{itemize}

\item \textbf{RQ1:} Does RBM-IM offer better concept drift detection than state-of-the-art drift detectors designed for standard data streams?  

\item \textbf{RQ2:} Does RBM-IM offer better concept drift detection than state-of-the-art skew-insensitive drift detectors designed for imbalanced data streams? 

\item \textbf{RQ3:} What is the capability of RBM--IBM to detect local drifts that affect a subset of minority classes?

\item \textbf{RQ4:} What robustness to increasing imbalance ratio is offered by RBM-IM?

\end{itemize}

\noindent All methods and experiments were implemented in MOA environment \cite{Bifet:2010moa} and run on Intel Core i7-8365u with 64GB DDR4 RAM.

\subsection{Data stream benchmarks}

For the purpose of this experimental study, we selected 24 benchmark data streams: 12 come from real-world domains and 12 were generated artificially using the MOA environment \cite{Bifet:2010moa}. Such a diverse mix allowed us to evaluate the effectiveness of RBM-IM over a plethora of scenarios. Using artificial data streams allows us to control the specific nature of drift and class imbalance, as well as to inject local concept drift into selected minority classes. Artificial data streams use a dynamic imbalance ratio that both increases and decreases over time.  Real-world streams offer challenging problems that are characterized by a mix of different learning difficulties. Properties of the data stream benchmarks are given in Tab.~\ref{tab:data}. We report the highest imbalance ratio among all the classes, i.e., the ratio between the biggest and the smallest class.

\begin{table}[h!]
	\centering
	\caption{Properties of real-world (top) and artificial (bottom) imbalanced data stream benchmarks.}
	\label{tab:data}
	\resizebox{0.49\textwidth}{!}{
		\begin{tabular}{lrrrrr}
			\toprule
			Dataset & Instances & Features & Classes & IR & Drift \\ 
			\midrule
			Activity-Raw & 1 048 570 & 3 & 6 & 128.93 & yes\\
			Connect4 & 67 557 & 42 & 3 & 45.81 & unknown\\
			Covertype &  581 012 & 54 & 7 & 96.14 & unknown\\
			Crimes & 878 049 & 3 & 39 & 106.72 & unknown\\
			DJ30 & 138 166 & 8 & 30 & 204.66 & yes\\
			EEG & 14 980 & 14 & 2 & 29.88 & yes\\
			Electricity & 45 312 & 8 & 2 & 17.54 & yes\\
            	Gas & 13 910 & 128 & 6 & 138.03 & yes\\
            	Olympic & 271 116 & 7 & 4 & 66.82 & unknown\\
            	Poker & 829 201 & 10 & 10 & 144.00 & yes\\
            	IntelSensors & 2 219 804 & 5 & 57 & 348.26 & yes\\
            	Tags &164 860 & 4 & 11 & 194.28 & unknown\\
			\midrule
			Aggrawal5 & 1 000 000 & 20 & 5 & 50.00 & incremental\\
			Aggrawal10 & 1 000 000 & 40 & 10 & 80.00 & incremental\\
			Aggrawal20 &  2 000 000 & 80 & 20 & 100.00 & incremental\\
			Hyperplane5 & 1 000 000 & 20 & 5 & 100.00 & gradual\\
			Hyperplane10 & 1 000 000 & 40 & 10 & 200.00 & gradual\\
			Hyperplane20 & 2 000 000 & 80 & 20 & 300.00 & gradual\\
			RBF5 & 1 000 000 & 20 & 5 & 100.00 & sudden\\
            	RBF10 & 1 000 000 & 40 & 10 & 200.00 & sudden\\
            	RBF20& 2 000 000 & 80 & 20 & 300.00 & sudden\\
            	RandomTree5 &1 000 000 & 20 & 5 & 100.00 & sudden\\
            	RandomTree10 & 1 000 000 & 40 & 10 & 200.00 & sudden\\
            	RandomTree20 & 2 000 000 & 80 & 20 & 300.00 & sudden\\
			\bottomrule
		\end{tabular}
	}
\end{table}

\subsection{Setup}

\noindent \textbf{Reference concept drift detectors.} As reference methods to the proposed RBM-IM, we have selected three state-of-the-art concept drift detectors for standard data: WSTD \cite{Barros:2018w}, RDDM \cite{Barros:2017}, and FHDDM \cite{Pesaranghader:2016}; as well as two state-of-the-art drift detectors for imbalanced data streams: PerfSim and DDM-OCI. Parameters of all the six drift detectors are given in Tab.~\ref{tab:ddp}.
	
		\begin{table}[h!]
		\centering
		\caption{Examined drift detectors and their parameters.}
\scalebox{0.65}{
		\begin{tabular}{lll}
			\toprule
			Abbr. & Name & Parameters \\ 
			\midrule
	    	WSTD \cite{Barros:2018w} & Wilcoxon Rank Sum Test       & sliding window size $\omega \in \{25,50,75,100\}$\\
	    	&Drift Detection                                   & warning significance $\alpha_w \in \{0.01,0.03,0.05,0.07\}$\\
	    	&								    & drift significance $\alpha_d \in \{0.001,0.003,0.005,0.007\}$\\  
	    	& 							        & max. no of old instances $\min \in \{1000,2000,3000,4000\}$\\
    	RDDM \cite{Barros:2017} & Reactive Drift Detection     & warning threshold $\alpha_w \in \{0.90,0.92,0.95,0.98\}$\\
	    	&                                   & drift threshold $\alpha_d \in \{0.80,0.85,0.90.0.95\}$\\
	    	&								    & min. no. of errors $e \in \{10,30,50,70\}$\\   
	    	& 							        & min. no. of instances $\min \in \{3000,5000,7000,9000\}$\\
	    	&                                   & max. no. of instances $\max \in \{10000,20000,30000,40000\}$\\
	    	&								    & warning limit $wL \in \{800,1000,1200,1400\}$\\   
FHDDM \cite{Pesaranghader:2016}& Fast Hoeffding Drift Detection & sliding window size $\omega \in \{25,50,75,100\}$\\
	    		 &								  & allowed error $\delta \in \{0.000001,0.00001,0.0001,0.001\}$ \\ 
		PerfSim \cite{Antwi:2012} & Performance Similarity     & differentiation weights $\lambda \in \{0.1,0.2,0.3,0.4\}$\\
				 & 								& min. no. of errors $n = \{10,30,50,70\}$\\
		DDM--OCI \cite{Wang:2020} & Drift Detection Method        & warning threshold $\alpha_w \in \{0.90,0.92,0.95,0.98\}$\\
				 & for online class imbalance                             & drift threshold $\alpha_d \in \{0.80,0.85,0.90.0.95\}$\\
				 &								& min. no. of errors $e \in \{10,30,50,70\}$\\
			\midrule
			RBM-IM & RBM Drift Detection& mini--batch size $\mathbf{M} \in \{25,50,75,100\}$\\
			         & for imbalanced data streams									  &	visible neurons $\mathbf{V} = $ no. of features\\ 
		           	&									  &	hidden neurons $\mathbf{H} \in \{0.25\mathbf{V},0.5\mathbf{V},0.75\mathbf{V},\mathbf{V}\}$ \\ 
			         &									  &	class neurons $\mathbf{Z} = $ no. of classes\\ 
			         &									  & learning rate $\eta \in \{0.01,0.03,0.05,0.07\}$\\
			         &                                    & Gibbs sampling steps $k \in \{1,2,3,4\}$\\
			\bottomrule
			\label{tab:ddp}
		\end{tabular}
}
	\end{table}

	\noindent \textbf{Parameter tuning.} In order to offer a fair and thorough comparison, we performed parameter tuning for every drift detector and for every data stream benchmark. As we deal with a streaming scenario, we used self hyper-parameter tuning \cite{Veloso:2018} that is based on the online Nelder-Mead optimization.

	\noindent \textbf{Base classifier.} In order to ensure fairness when comparing the examined drift detectors they all use Adaptive Cost-Sensitive Perceptron Trees \cite{Krawczyk:2017ecml} as a base classifier. This is a skew-insensitive and efficient classifier capable of handling both binary and multi-class imbalanced data streams, but is strongly dependent on an attached concept drift detection component. Therefore, it offers an excellent backbone for our experiments, allowing us to directly measure how a given drift detector impacts the classification quality. 

	\noindent\textbf{RBM-IM training.} Our drift detector uses the first instance batch to train itself at the beginning of the stream processing. It continuously updates itself in an online fashion together with the base classifier.
	
	\noindent \textbf{Evaluation metrics.} As we deal with multi-class imbalanced and drifting data streams, we evaluated the examined algorithms using prequential multi-class AUC \cite{Wang:2020} and prequential multi-class G-mean \cite{Korycki:2020}. 
	
	\noindent\textbf{Windows.} We used a window size $W = 1000$ for calculating the prequential metrics. ADWIN self-adapting window was used for both RBM-IM and reference drift detectors to alleviate the need for manual window size tuning \cite{Korycki:2020}. 
	
	\noindent\textbf{Statistical analysis.} We used the Friedman ranking test with Bonferroni-Dunn post-hoc and Bayesian signed test \cite{Benavoli:2017} for statistical significance over multiple comparison with significance level $\alpha = 0.05$. 

	\noindent\textbf{Drift injection.} For experiment 2, we inject local concept drift starting with the smallest minority class and then add classes according to their increasing size. This allows us to consider most difficult scenarios, where smallest classes are affected by the local concept drift and thus most likely to be neglected.

\begin{table*}[h!]
	\centering
	\caption{Results according to pmAUC, pmGM and update times per batch for the examined concept drift detectors.}
	\label{tab:res}
	\resizebox{0.98\textwidth}{!}{
		\begin{tabular}{lccccccrcccccc}
			\toprule
			Dataset & \multicolumn{6}{c}{pmAUC} & & \multicolumn{6}{c}{pmGM}  \\
			 & WSTD & RDDM & FHDDM & PerfSim & DDM--OCI & RBM-IM  & & WSTD & RDDM & FHDDM & PerfSim & DDM--OCI & RBM-IM \\
			\cmidrule(l){1-7}\cmidrule(l){8-14}
			Activity-Raw & 45.43 & 46.23& 48.45 & 72.81 & 74.29&\textbf{79.92} & &51.06 & 54.10&55.82 &76.11 & 78.59& \textbf{82.04} \\
			Connect4 & 54.19&53.48 &55.27 &64.19 &69.10 & \textbf{75.04}& & 55.03 & 55.39 & 56.29& 66.08&70.21 &\textbf{77.92}  \\
			Covertype & 33.19 & 34.12& 35.72& 41.24 & 40.58 & \textbf{53.98} & &32.45 & 33.10& 35.98& 40.19& 41.02& \textbf{54.02}  \\
			Crimes &19.93 & 20.04& 22.11& 28.56& 30.02& \textbf{64.59}& & 21.88& 23.92& 26.01& 30.99& 32.07& \textbf{69.58} \\
			DJ30 &26.94 & 25.98 & 26.02&34.11 &33.98 &\textbf{59.04} & & 27.45 &27.11 &28.73 & 36.71 & 35.48& \textbf{61.29} \\
			EEG & 58.14 & 59.98 & 62.29 & 70.08 & \textbf{74.22}& 72.03 & &59.85 & 60.98 & 64.67& 72.93& \textbf{77.29}& 74.13 \\
			Electricity & 68.94 & 72.10& 73.45& 80.04& \textbf{83.20}& 79.39 & &70.45 &75.90 &77.28 & 83.92 & \textbf{85.44}& 81.99 \\
            	Gas &48.83 & 47.23 & 46.92 & 63.59& \textbf{67.54}& 64.20& & 50.05& 49.54& 49.17& 65.98& \textbf{70.02}& 66.13 \\
            	Olympic &72.98 & 70.34& 74.53& 80.08& 83.19& \textbf{87.01}& & 73.95 & 71.91 & 76.02 & 83.19& 86.88 &\textbf{89.24}  \\
            	Poker &72.11 & 69.65& 72.98& 84.65& 87.91& \textbf{91.03}& & 74.46 & 70.97 & 74.52 & 87.11 & 89.34 & \textbf{93.06}  \\
            	IntelSensors &9.45 & 11.45 & 13.99& 36.23& 37.08& \textbf{58.10}& & 10.02 & 13.01 & 14.38& 37.82& 38.03& \textbf{60.39} \\
            	Tags & 30.45 &28.67 & 29.45& \textbf{42.68}& 40.18& 39.04 & & 33.10 & 30.08& 31.14 & \textbf{45.28} & 43.21& 41.02  \\
			\cmidrule(l){1-7}\cmidrule(l){8-14}
			Aggrawal5 & 78.34& 77.45& 80.41& 84.92& 88.34 & \textbf{90.38} & & 77.19 & 79.02 & 80.93 & 85.99 & 90.02 & \textbf{93.01}  \\
			Aggrawal10 & 70.12& 68.34& 70.23& 74.99& 78.32& \textbf{88.02} &  &71.04& 70.16& 71.88 & 75.38 & 79.14 & \textbf{90.49}  \\
			Aggrawal20 &55.62 & 56.23& 58.93& 65.76& 66.98& \textbf{83.87}& &56.45 &57.22 &59.39 & 66.28& 67.57&\textbf{85.09}  \\
			Hyperplane5 & 62.05& 63.66& 62.07&70.45 &73.98 & \textbf{75.06}& &65.39 & 67.20 & 66.14 & 74.82 & 78.05 & \textbf{81.80}  \\
			Hyperplane10 & 53.56& 54.37 & 54.02& 63.74 & 66.59 &\textbf{72.30} & & 56.93 & 59.14 & 57.92 &  66.72& 70.56& \textbf{78.03} \\
			Hyperplane20 & 40.04 & 38.45& 42.19& 50.10& 57.67& \textbf{66.48}& & 42.06 & 41.99& 40.86& 52.19& 59.37& \textbf{68.27} \\
			RBF5 &80.18 &78.56 &82.40 & 90.48& 92.36& \textbf{92.78}& & 83.47 & 81.59& 84.99& 92.12 & 94.82 &\textbf{94.97}  \\
            	RBF10 & 69.45& 67.84& 73.29&82.19 & 84.48 &\textbf{88.82} & & 72.19 & 70.48 & 76.44 & 85.11 & 87.81  & \textbf{90.26}  \\
            	RBF20 & 53.18 & 52.88 & 54.01 & 70.24 & 71.93 & \textbf{83.08} & & 55.98 & 54.90 & 57.73 & 73.89 &  74.84& \textbf{85.30}  \\
            	RandomTree5 & 45.29 & 47.21 & 47.93 & 58.90 & 64.32 & \textbf{67.98} & & 46.12 & 48.52 & 49.11 & 60.05 & 66.30 & \textbf{69.93}  \\
            	RandomTree10 & 31.63 & 33.19& 35.02& 50.02& 53.87& \textbf{63.01}& & 32.79 & 33.90 & 36.14 & 51.58 & 55.20 & \textbf{64.97}  \\
            	RandomTree20 &19.83 & 20.04 & 21.38 & 36.29 & 43.22& \textbf{59.42}& & 20.02 & 20.88 &22.94 & 38.01 & 44.87& \textbf{60.33}  \\
			\cmidrule(l){1-7}\cmidrule(l){8-14}
			ranks & 5.46 & 4.78 & 3.84 & 2.97 & 2.56 & 1.39 & & 5.80 & 5.05 & 4.15 & 2.45 & 2.29 & 1.26 \\
			\cmidrule(l){1-7}\cmidrule(l){8-14}
                 Avg. test time [s] & 17.26$\pm$3.11 & 18.11$\pm$4.72 & 16.54$\pm$2.98 & 8.92$\pm$3.07 & 9.78$\pm$4.14 & 6.28$\pm$1.08 & &  &  &  &  & &  \\
			Avg. update time [s] & 0.02$\pm$0.01 & 0.08$\pm$0.02 & 0.11$\pm$0.05 & 19.83$\pm$6.98 & 18.54$\pm$7.82 & 12.22$\pm$0.92 & &  &  &  &  &  &  \\
			\bottomrule
		\end{tabular}
	}
\end{table*}

\subsection{Experiment 1: Drift detectors comparison}

\begin{figure}[h]
	\centering
	\tiny
	\resizebox{\columnwidth}{!}{
		\begin{tikzpicture}
		%axis
		\draw (1,0) -- (6.5,0);
		\foreach \x in {1,2,3,4,5,6} {
			\draw (\x, 0) -- ++(0,.1) node [below=0.15cm,scale=0.75] {\tiny \x};
			\ifthenelse{\x < 7}{\draw (\x+.5, 0) -- ++(0,.03);}{}
		}
		% coordinates
		\coordinate (c0) at (1.26,0);
		\coordinate (c1) at (2.29,0);
		\coordinate (c2) at (2.45,0);
		\coordinate (c3) at (4.15,0);
		\coordinate (c4) at (5.05,0);
		\coordinate (c5) at (5.80,0);
		
		% labels
		\node (l0) at (c0) [above right=.15cm and 0.1cm, align=center, scale=0.7] {\tiny RBM-IM};
		\node (l1) at (c1) [above right=.4cm and 0.1cm, align=center, scale=0.7] {\tiny DDM-OCI};
		\node (l2) at (c2) [above right=.15cm and 0.1cm, align=center, scale=0.7] {\tiny PerfSim};
		\node (l3) at (c3) [above right=.15cm and 0.08cm, align=center, scale=0.7] {\tiny FHDDM};
		\node (l4) at (c4) [above right=.4cm and 0.1cm, align=center, scale=0.7] {\tiny RDDM};
		\node (l5) at (c5) [above right=.15cm and 0.1cm, align=center, scale=0.7] {\tiny WSTD};
		
		\fill[fill=gray,fill opacity=0.5] (1.3,-0.08) rectangle (1.3+0.77,0.08);
		
		% connectors
		\foreach \x in {0,...,5} {
			\draw (l\x) -| (c\x);
		};
		\end{tikzpicture}
	}
	\caption{The Bonferroni-Dunn test (pmGM).}
	\label{fig:bon2}
\end{figure}
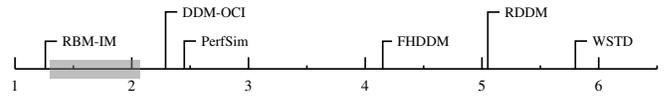

\begin{figure}[h]
	\centering
			\includegraphics[width=0.49\linewidth]{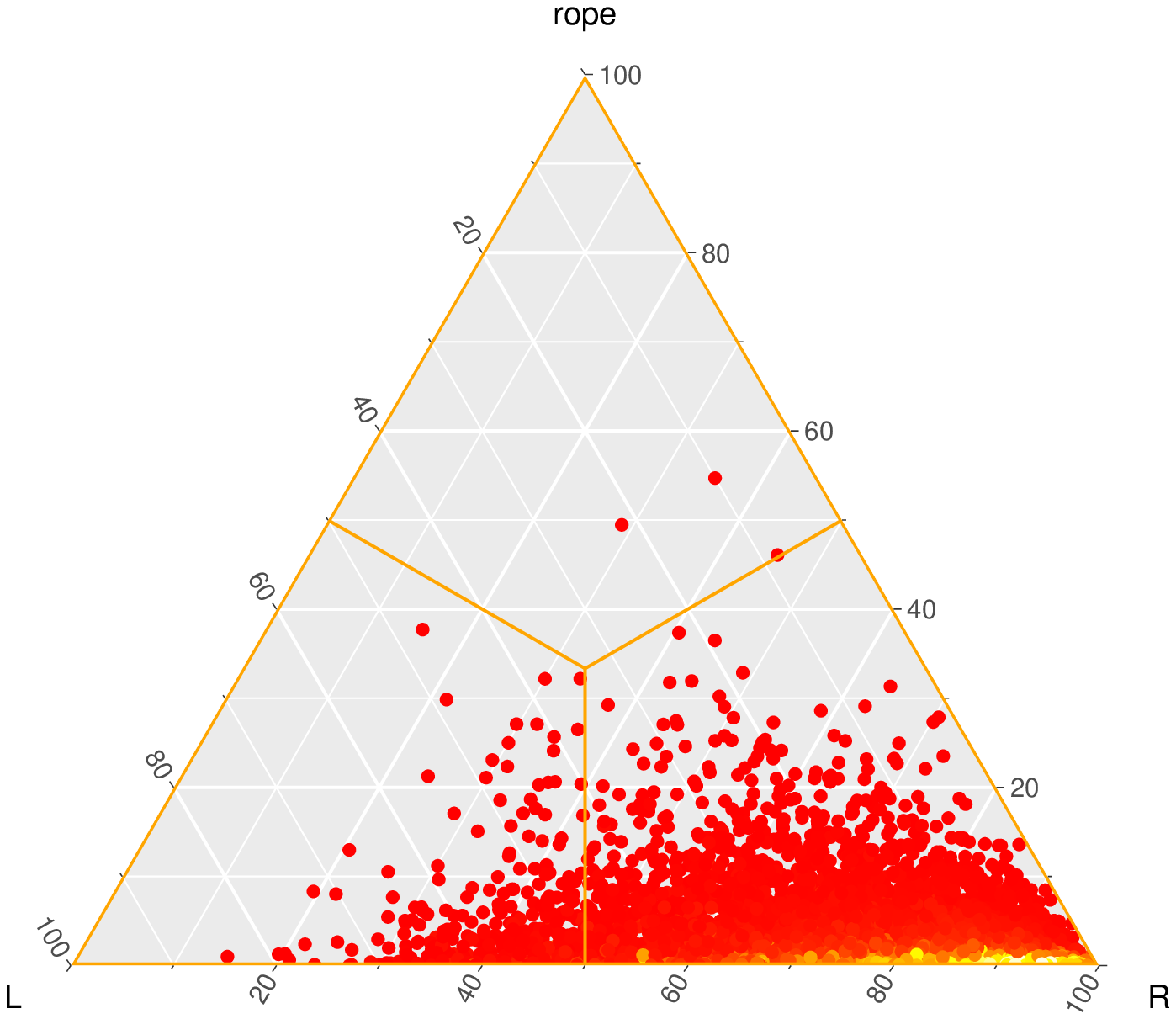}
			\includegraphics[width=0.49\linewidth]{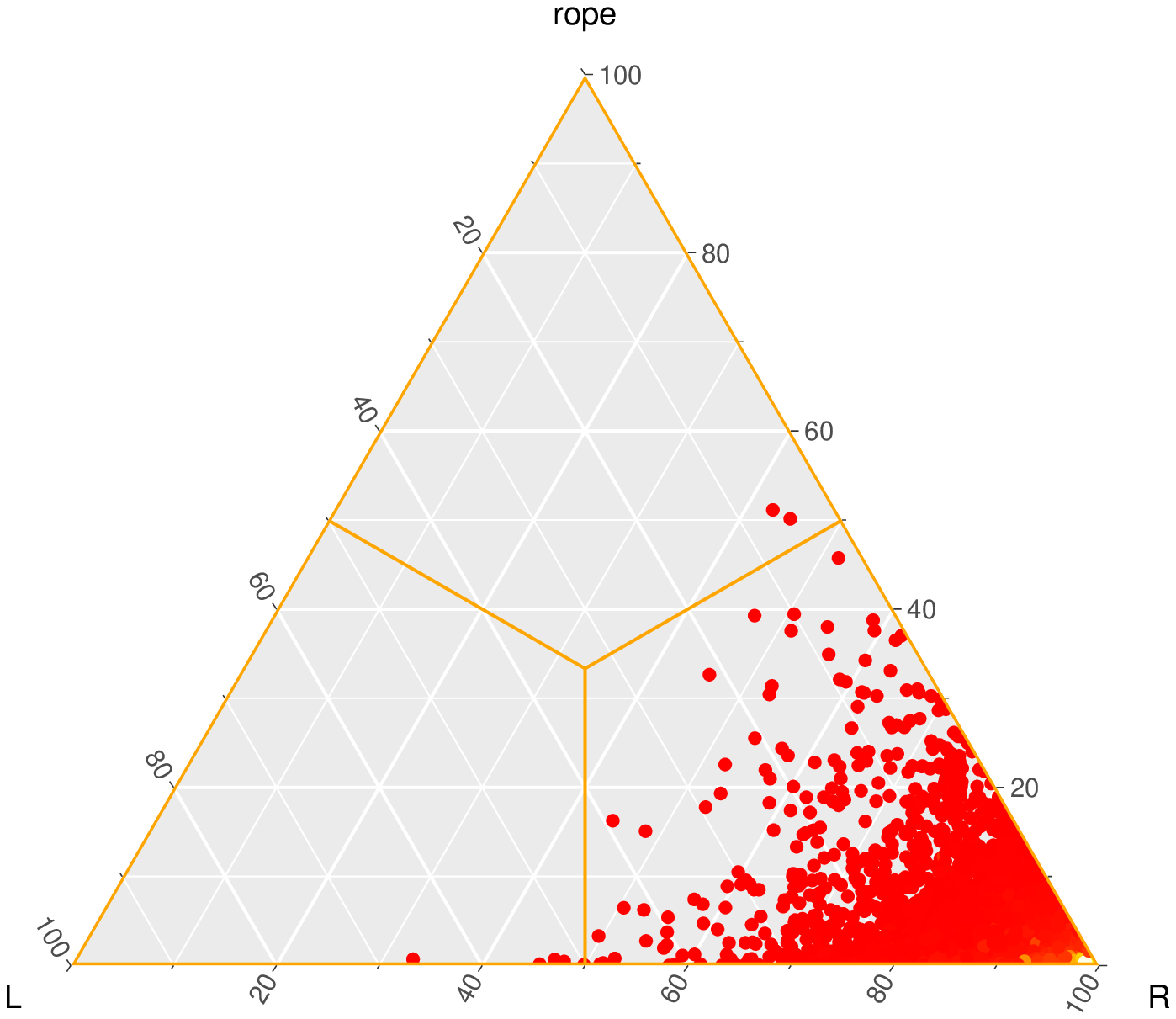}
	\caption{Visualizations of the Bayesian signed test for comparison between PerfSim and RBM-IM for pmAUC (left) and pmGM (right).}
	\label{fig:bt1}
\end{figure}

\begin{figure}[h]
	\centering
			\includegraphics[width=0.49\linewidth]{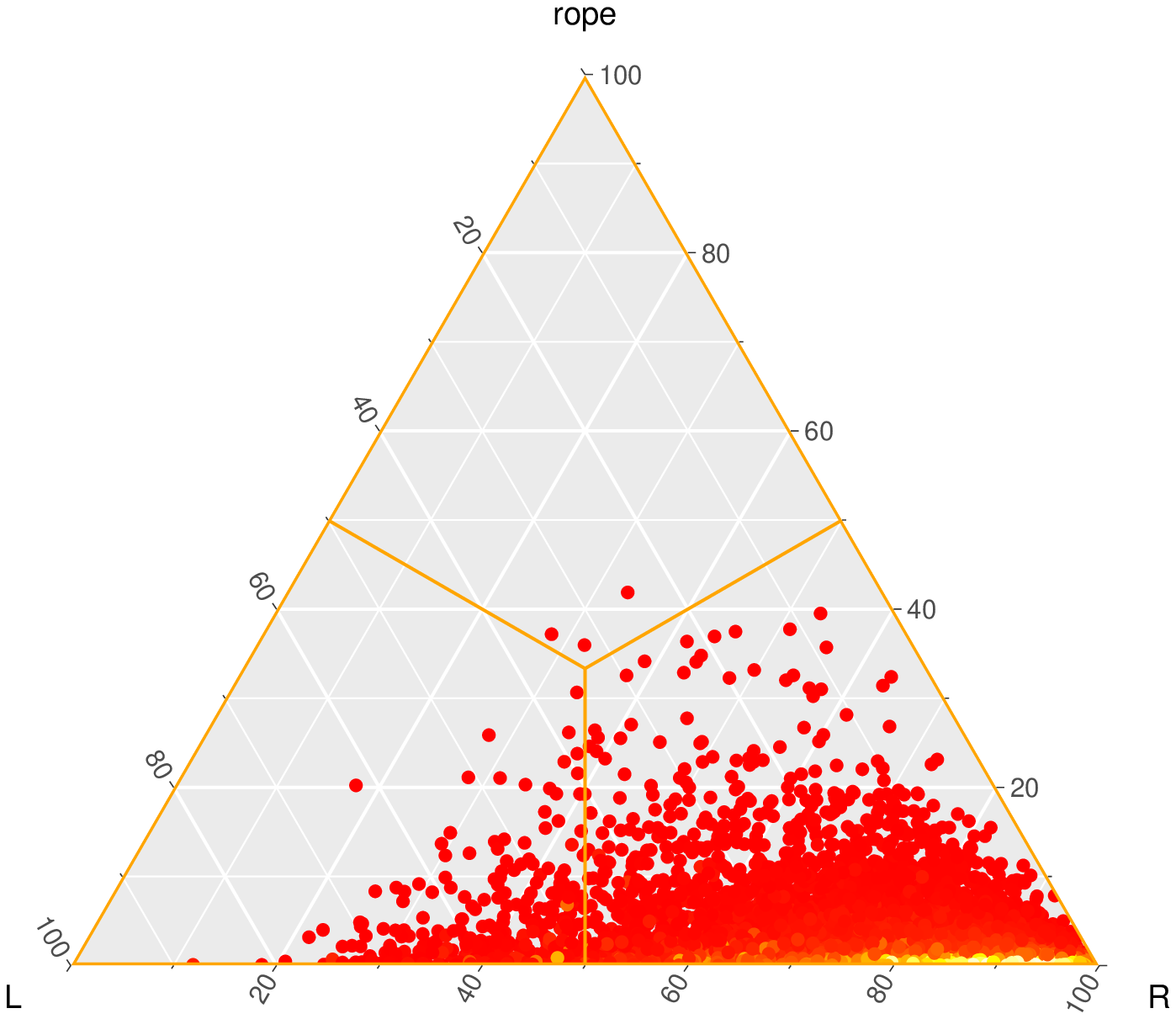}
			\includegraphics[width=0.49\linewidth]{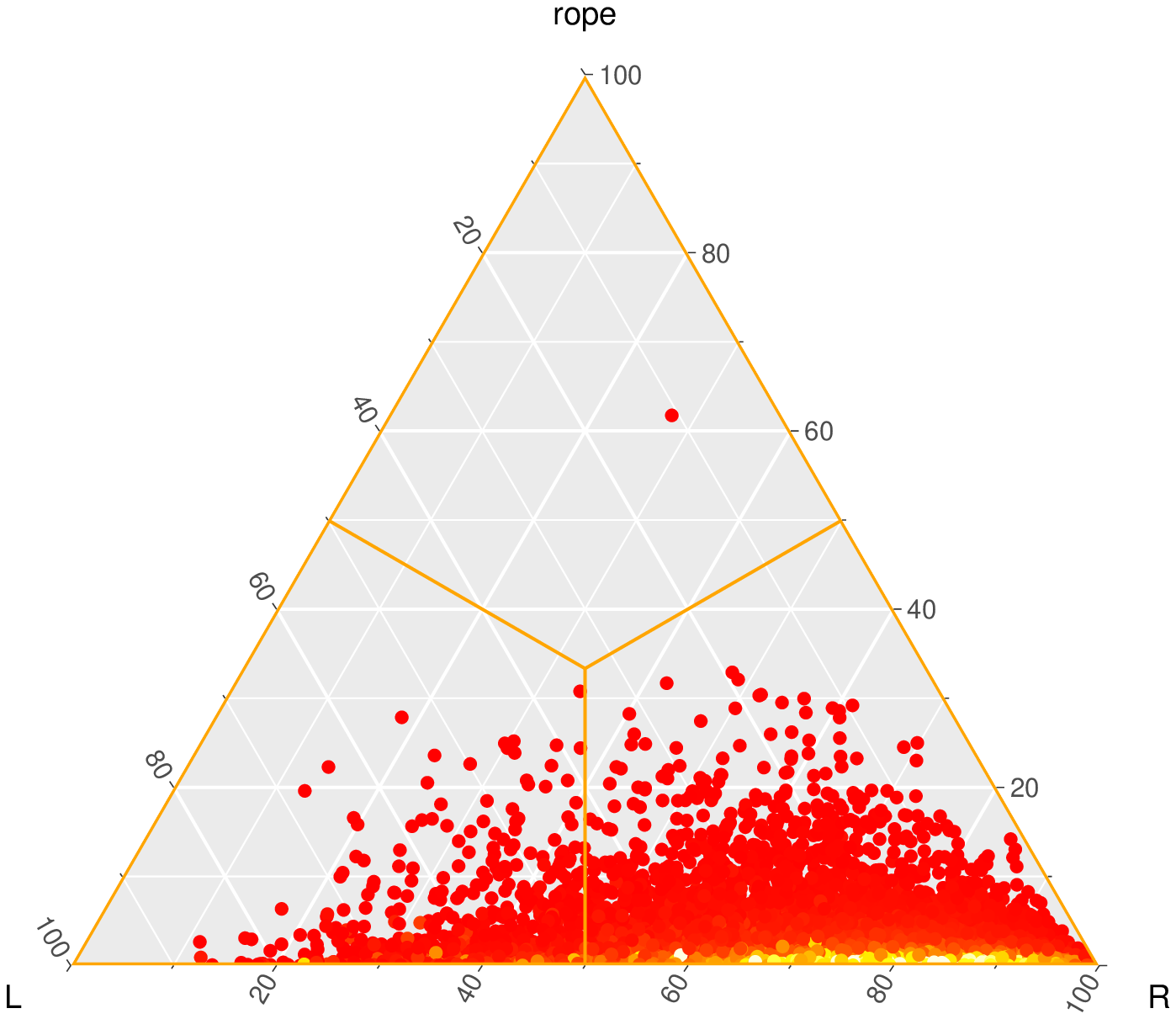}
	\caption{Visualizations of the Bayesian signed test for comparison between DDM-OCI and RBM-IM for pmAUC (left) and pmGM (right).}
	\label{fig:bt2}
\end{figure}

The first experiment was designed to analyze the behavior of the six examined drift detectors under two different metrics measured on all 24 benchmark data streams. This will allow us to evaluate how competitive is RBM-IM as compared with the state-of-the-art reference methods. Results according to pmAUC and pmGM are given in Tab.~\ref{tab:res}, while Fig.~\ref{fig:bon1} and~\ref{fig:bon2} depict the outcomes of the post-hoc statistical tests of significance. Fig.~\ref{fig:bt1} and~\ref{fig:bt2} present visualizations of the Bayesian signed test for pairwise comparisons with two best performing reference detectors.

\begin{figure}[h]
	\centering
	\tiny
	\resizebox{\columnwidth}{!}{
		\begin{tikzpicture}
		%axis
		\draw (1,0) -- (6.5,0);
		\foreach \x in {1,2,3,4,5,6} {
			\draw (\x, 0) -- ++(0,.1) node [below=0.15cm,scale=0.75] {\tiny \x};
			\ifthenelse{\x < 7}{\draw (\x+.5, 0) -- ++(0,.03);}{}
		}
		% coordinates
		\coordinate (c0) at (1.39,0);
		\coordinate (c1) at (2.56,0);
		\coordinate (c2) at (2.97,0);
		\coordinate (c3) at (3.84,0);
		\coordinate (c4) at (4.78,0);
		\coordinate (c5) at (5.46,0);
		
		% labels
		\node (l0) at (c0) [above right=.15cm and 0.1cm, align=center, scale=0.7] {\tiny RBM-IM};
		\node (l1) at (c1) [above right=.4cm and 0.1cm, align=center, scale=0.7] {\tiny DDM-OCI};
		\node (l2) at (c2) [above right=.15cm and 0.1cm, align=center, scale=0.7] {\tiny PerfSim};
		\node (l3) at (c3) [above right=.15cm and 0.08cm, align=center, scale=0.7] {\tiny FHDDM};
		\node (l4) at (c4) [above right=.4cm and 0.1cm, align=center, scale=0.7] {\tiny RDDM};
		\node (l5) at (c5) [above right=.15cm and 0.1cm, align=center, scale=0.7] {\tiny WSTD};
		
		\fill[fill=gray,fill opacity=0.5] (1.43,-0.08) rectangle (1.45+1.04,0.08);
		
		% connectors
		\foreach \x in {0,...,5} {
			\draw (l\x) -| (c\x);
		};
		\end{tikzpicture}
	}
	\caption{The Bonferroni-Dunn test (pmAUC).}
	\label{fig:bon1}
\end{figure}
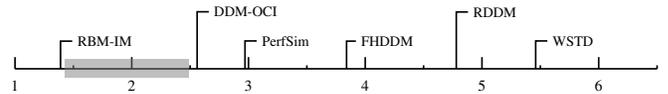

\smallskip 
\noindent \textbf{Comparison with standard drift detectors.} The standard drift detectors return unsatisfactory performance for all of the examined multi-class imbalanced data streams. This shows that the metrics collected by them are unsuitable to monitor skewed data streams. This also indicates that drift detectors, despite not being actually trainable models, are still prone to class imbalance. Despite the fact that the underlying classifier used was designed for imbalanced data streams, it could not offer accurate predictions when being feed incorrect information from the drift detectors. Especially in the case of data sets with a high number of classes (such as Crimes, DJ20, IntelSensor, or the artificial ones) standard drift detectors returned performance only slightly above a random guess. Those detectors were not capable of capturing changes affecting at the same time multiple class distributions and imbalance ratios. RBM-IM alleviated those limitations, while displaying comparable computational complexity.

\smallskip 
\noindent \textbf{Answer to RQ1:} Yes, RBM-IM offers significant improvements over standard drift detectors when applied to monitoring multi-class imbalanced data streams. Standard detectors cannot handle both a high number of classes and simultaneous changes in distributions and imbalance ratios. This shows that we need to have dedicated drift detectors for such difficult scenarios. 

\smallskip 
\noindent \textbf{Comparison with skew-insensitive drift detectors.} Skew-insensitive detectors performed significantly better when compared with their standard counterparts. However, for most of the real-world benchmarks and for all the artificial ones they still could not compete with RBM-IM. The only four data sets on which they returned a slightly better performance were EEG, Electricity, Gas, and Tags. All of them are relatively small and have a low number of classes. Especially the former factor might have had a strong impact on RBM-IM. As this is a trainable drift detector, it probably suffered from the problem of underfitting when learning from small data streams. This could be potentially alleviated by combining RBM-IM with transfer learning or instance exploitation techniques, which we will investigate in our future works. For all the remaining 20 data stream benchmarks RBM-IM outperformed in a statistically significant manner both PerfSim and DDM-OCI. This can be contributed to the compressed information about the current concept for each class stored within the RBM-IM structure, which allowed for a significantly more informative analysis of the changing properties of incoming instances.

\smallskip 
\noindent \textbf{Answer to RQ2:} Yes, RBM-IM is capable of outperforming state-of-the-art skew-insensitive drift detectors, while additionally offering faster detection and update times. This is especially visible on data sets with a high number of classes, where monitoring simple performance measures is not enough to accurately and timely detect occurrences of drifts. Additionally, by being a trainable detector RBM-IM can better adapt to changes in data streams, allowing fine-tuned encapsulation of the definition of what currently is considered as a temporal concept.

\subsection{Experiment 2: Detection of local concept drifts}

\begin{figure}[h!]
	\centering
			\includegraphics[width=0.32\linewidth]{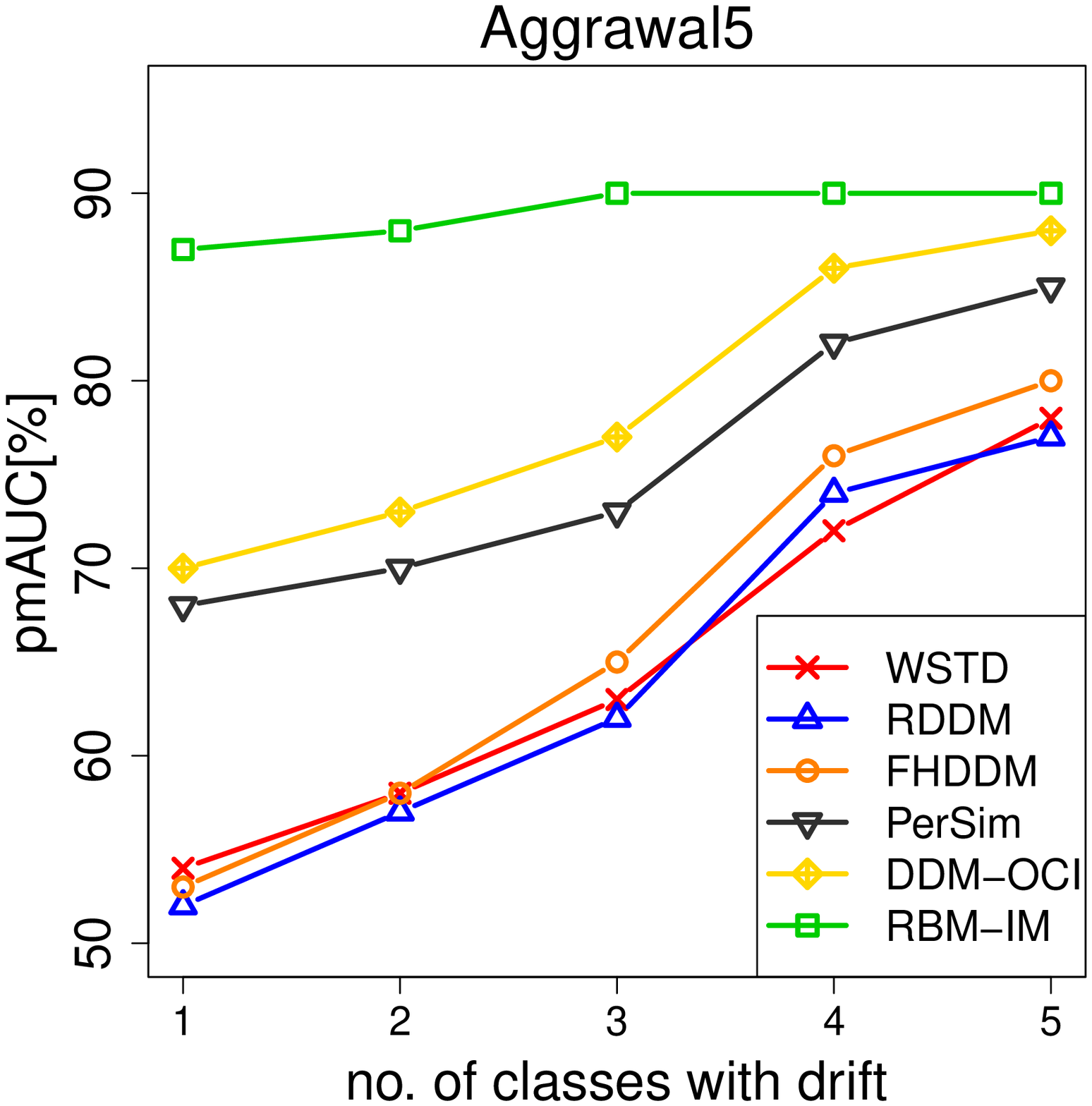}
			\includegraphics[width=0.32\linewidth]{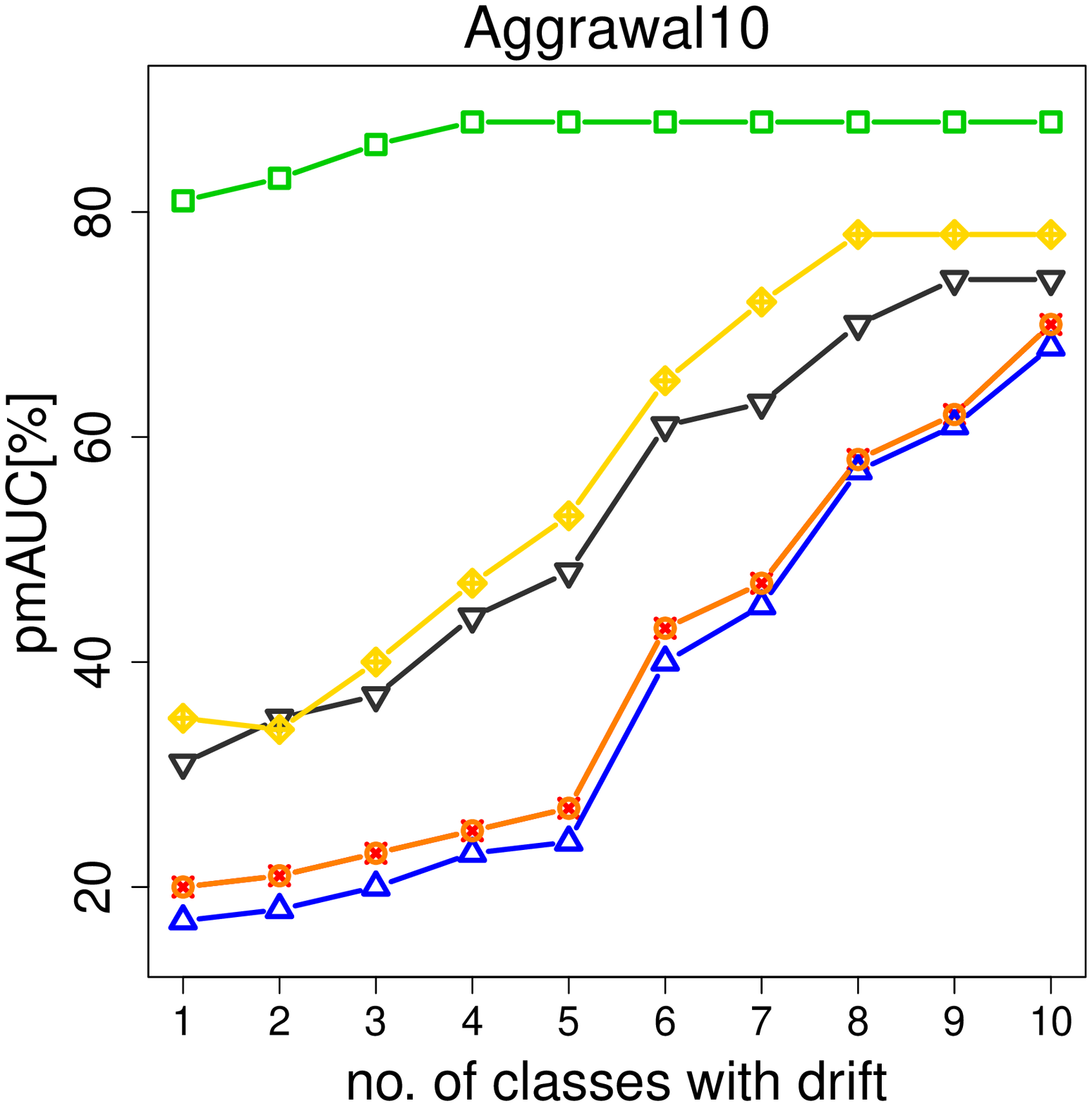}
			\includegraphics[width=0.32\linewidth]{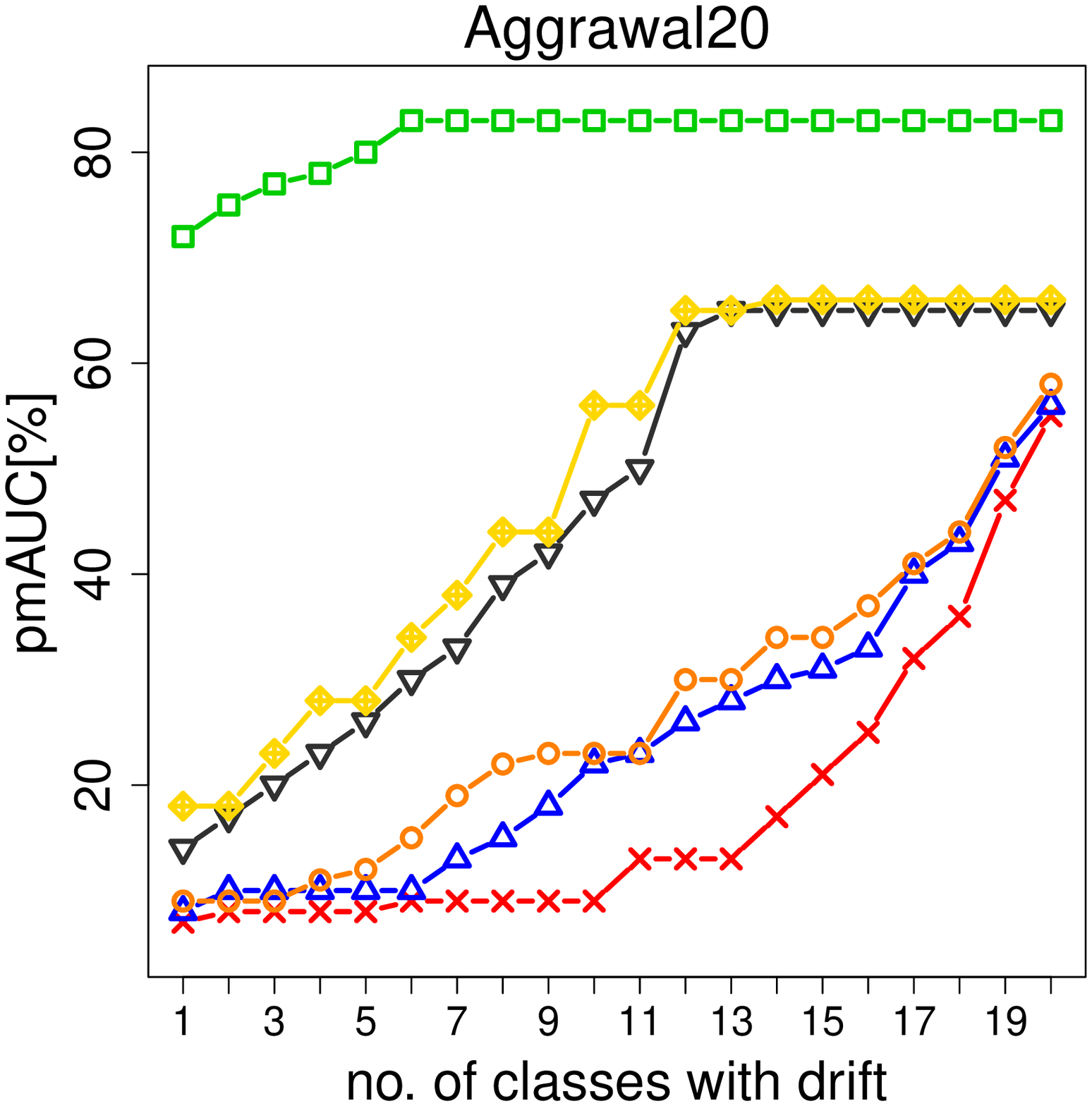}
			\includegraphics[width=0.32\linewidth]{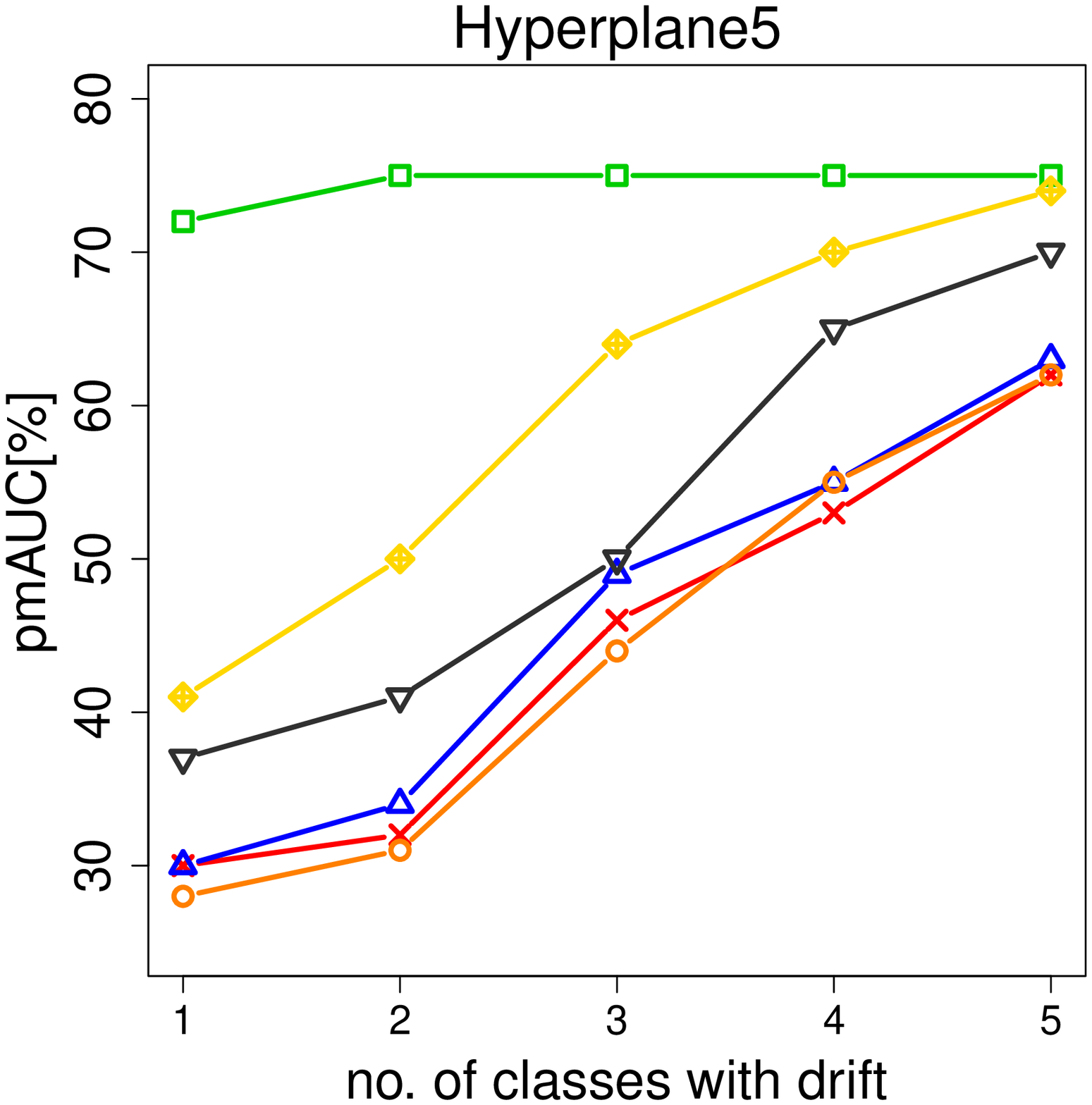}
			\includegraphics[width=0.32\linewidth]{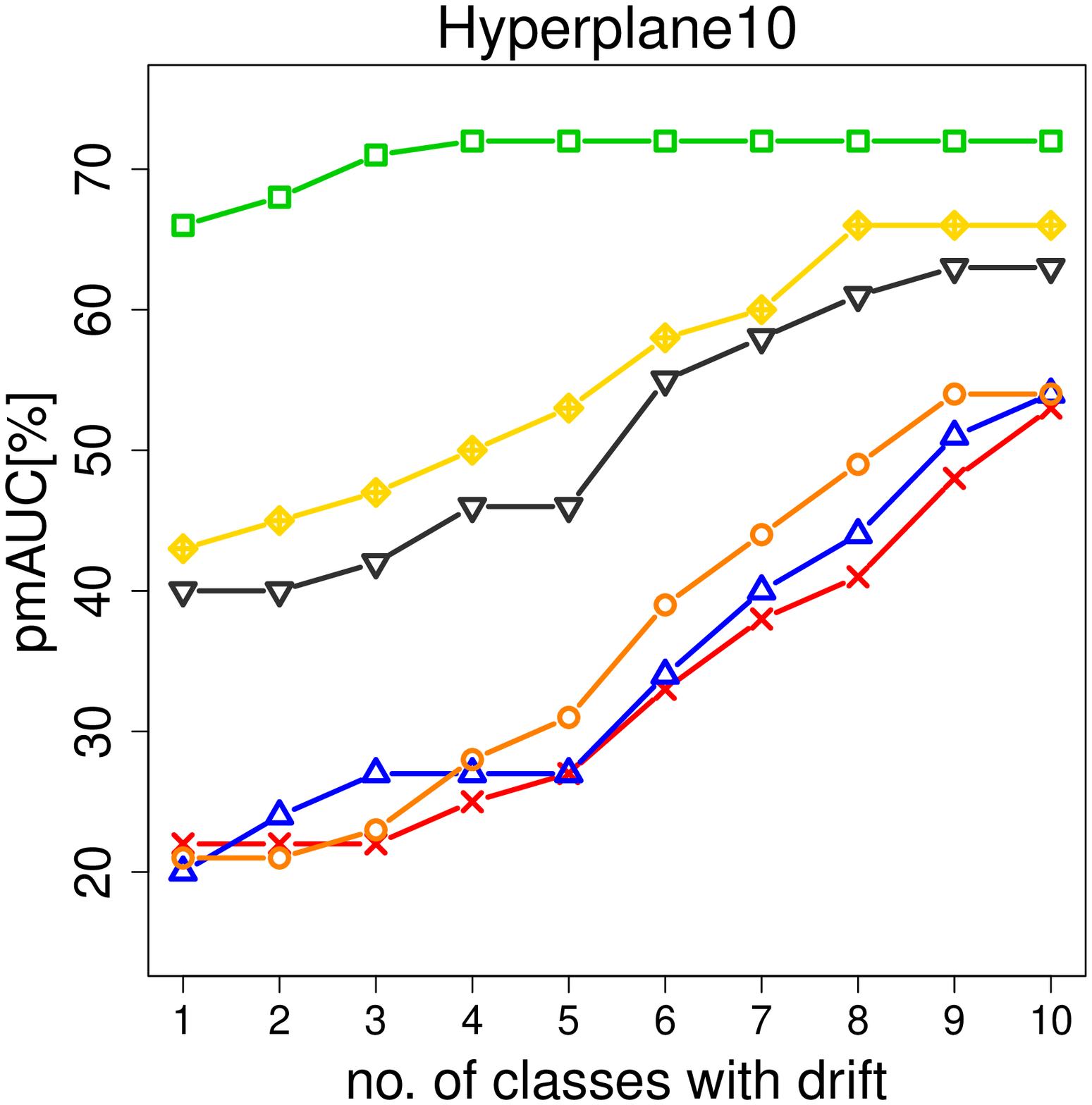}
			\includegraphics[width=0.32\linewidth]{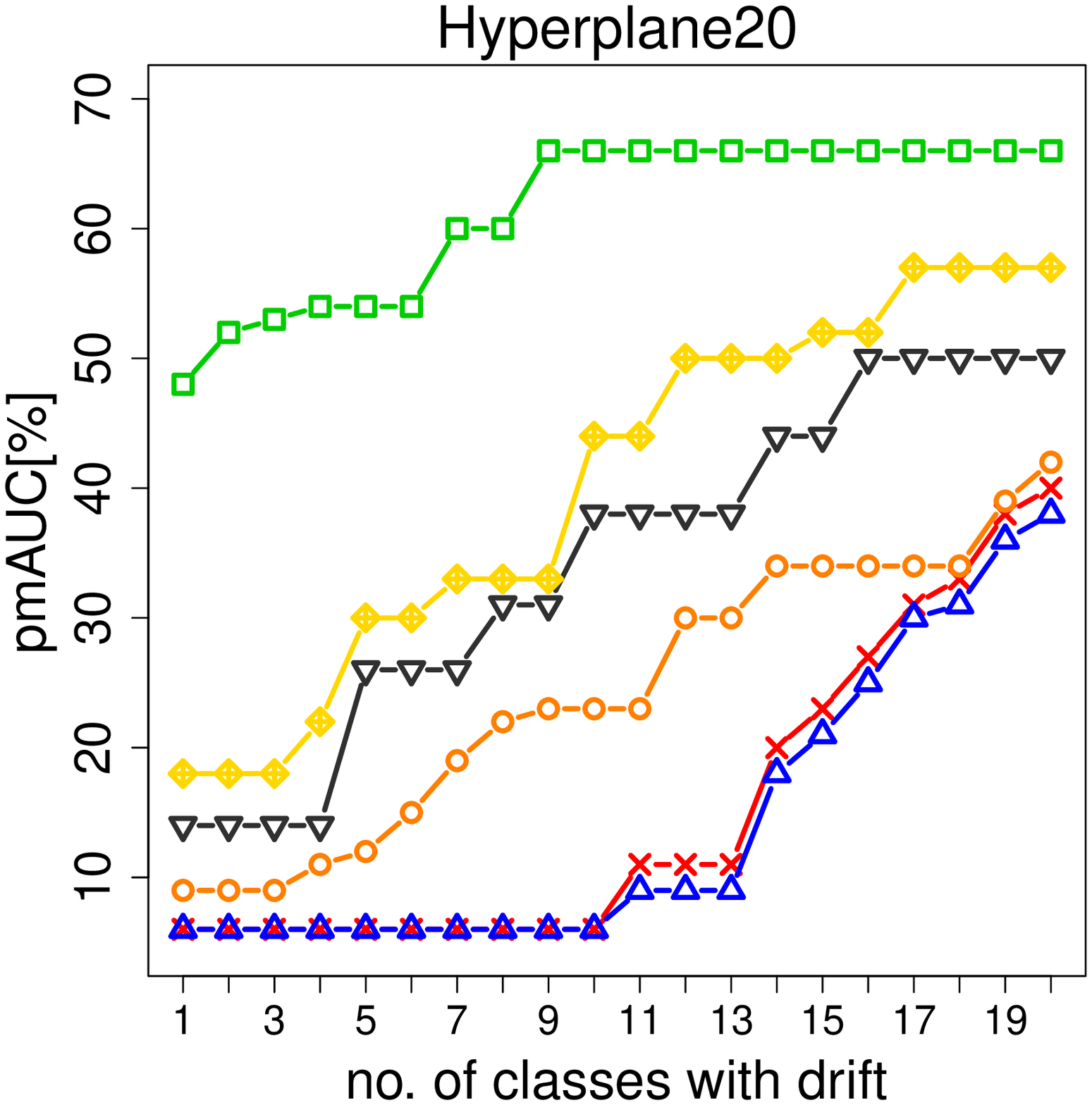}
			\includegraphics[width=0.32\linewidth]{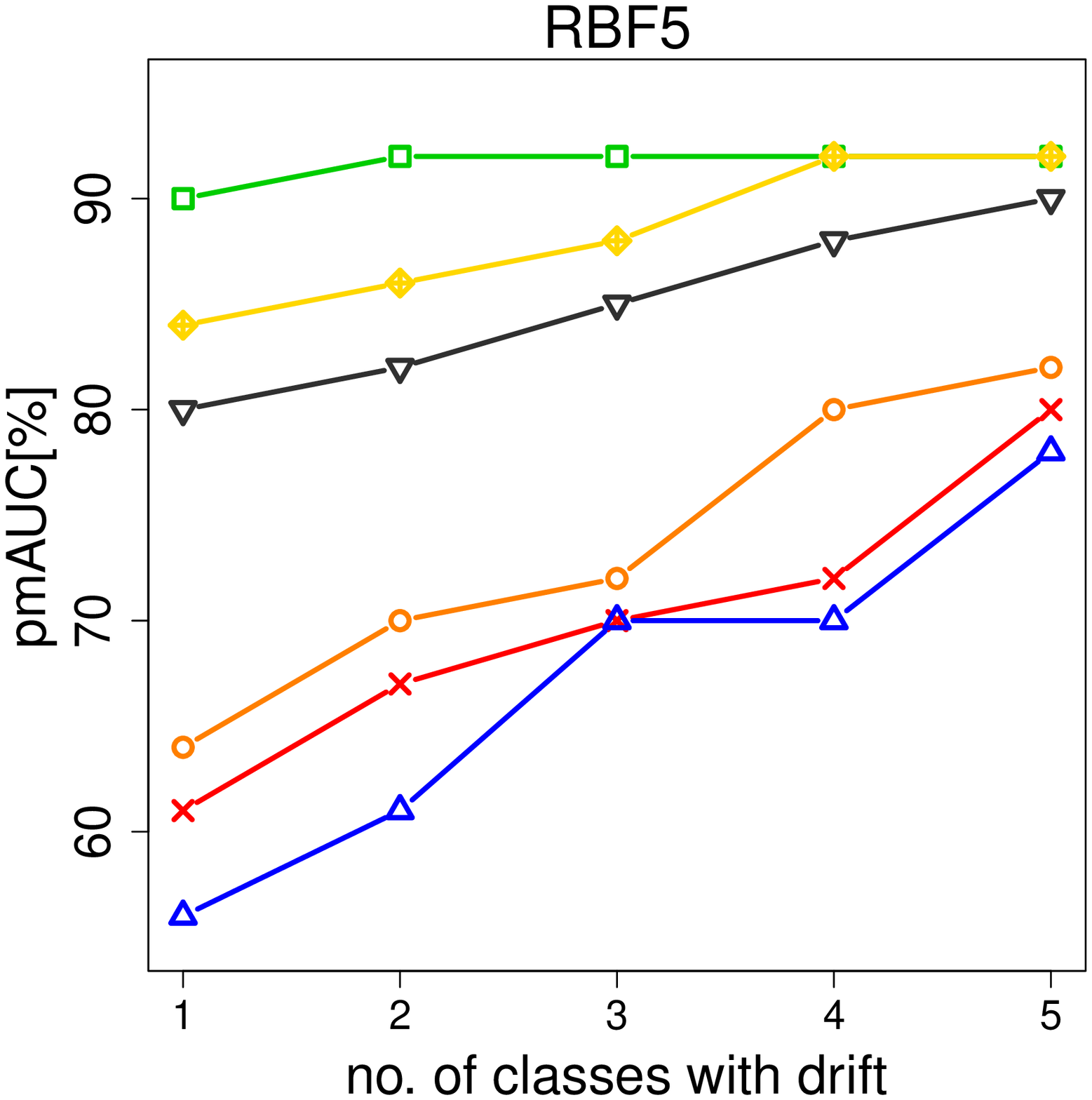}
			\includegraphics[width=0.32\linewidth]{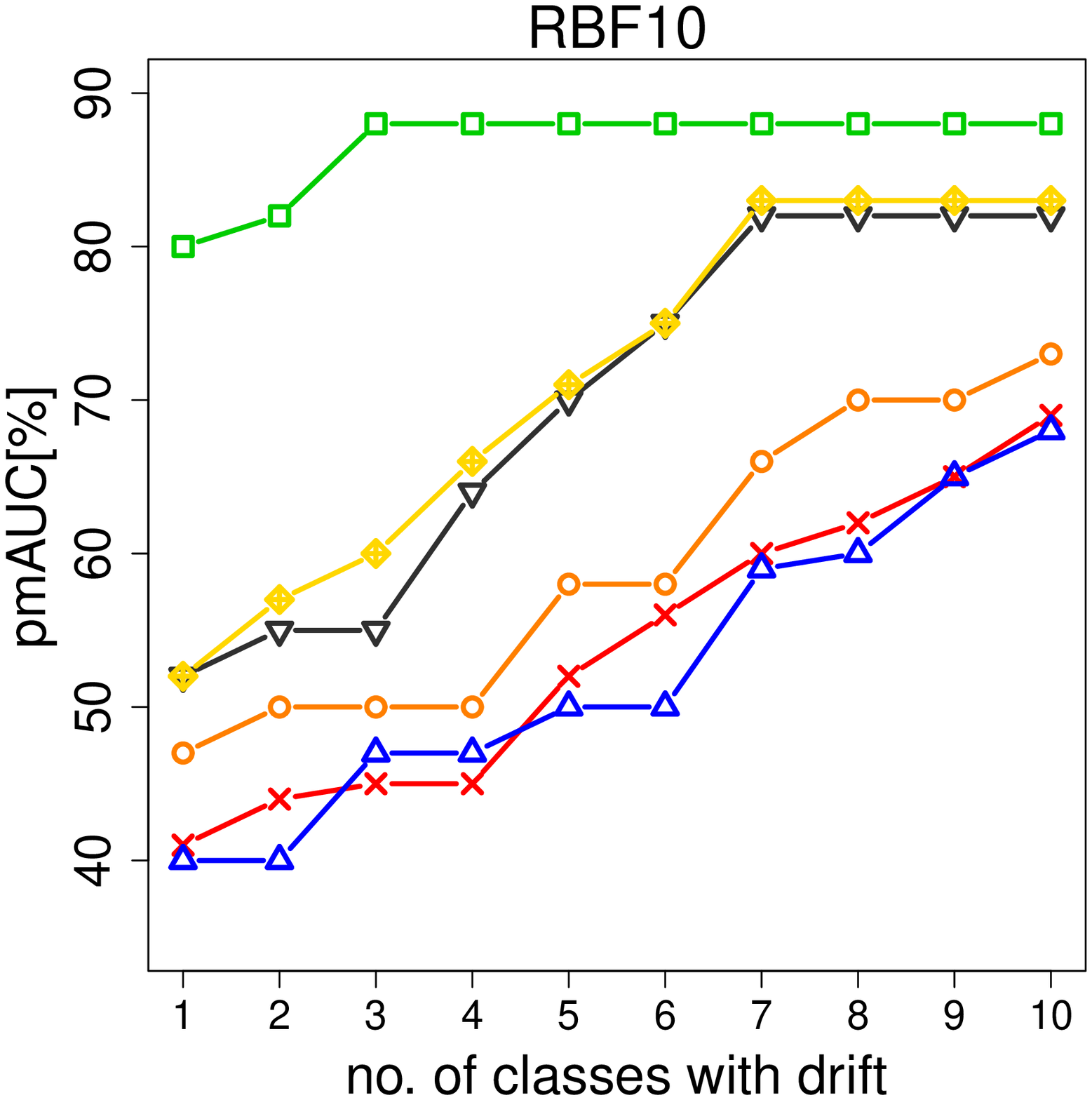}
			\includegraphics[width=0.32\linewidth]{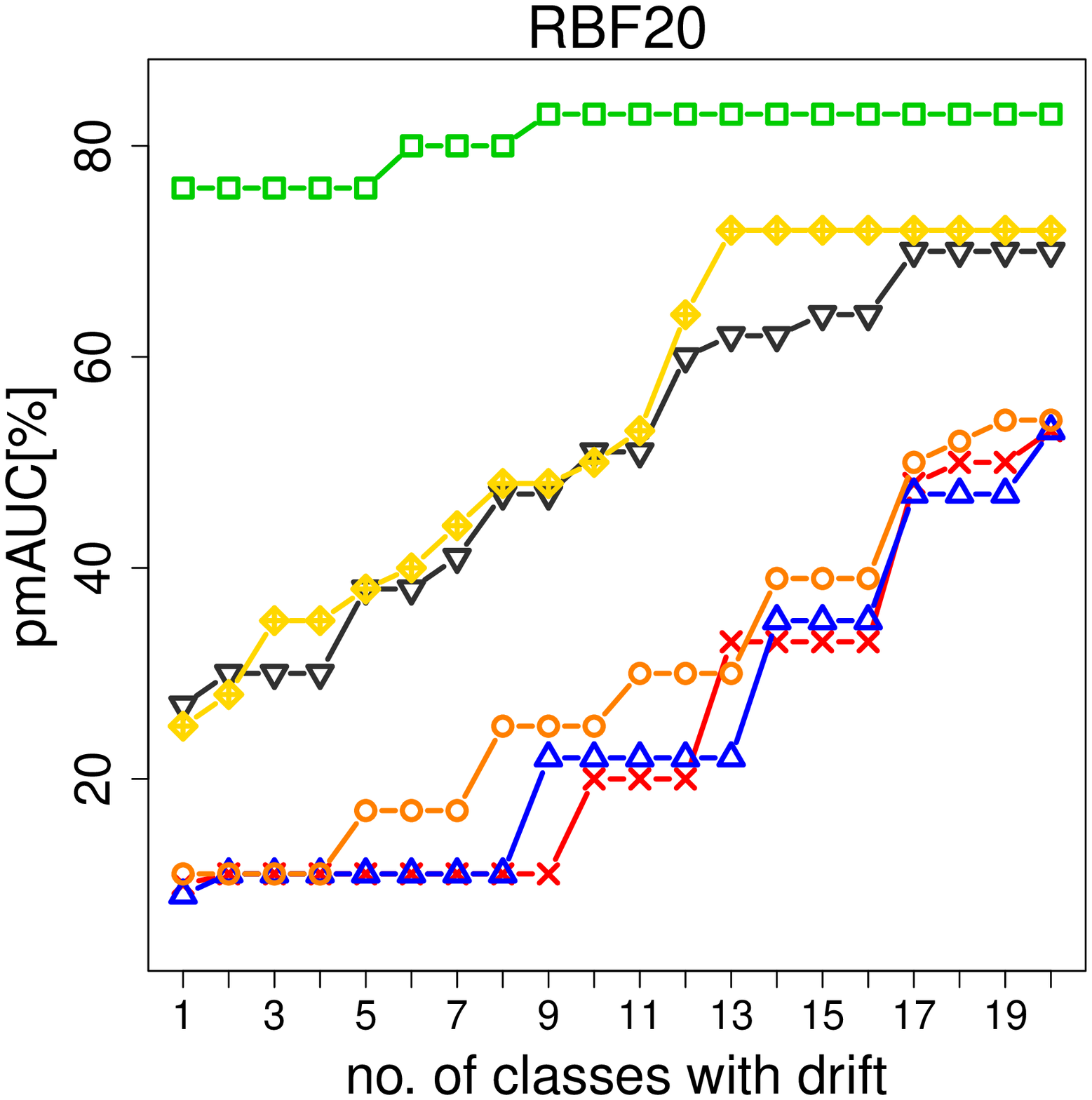}
			\includegraphics[width=0.32\linewidth]{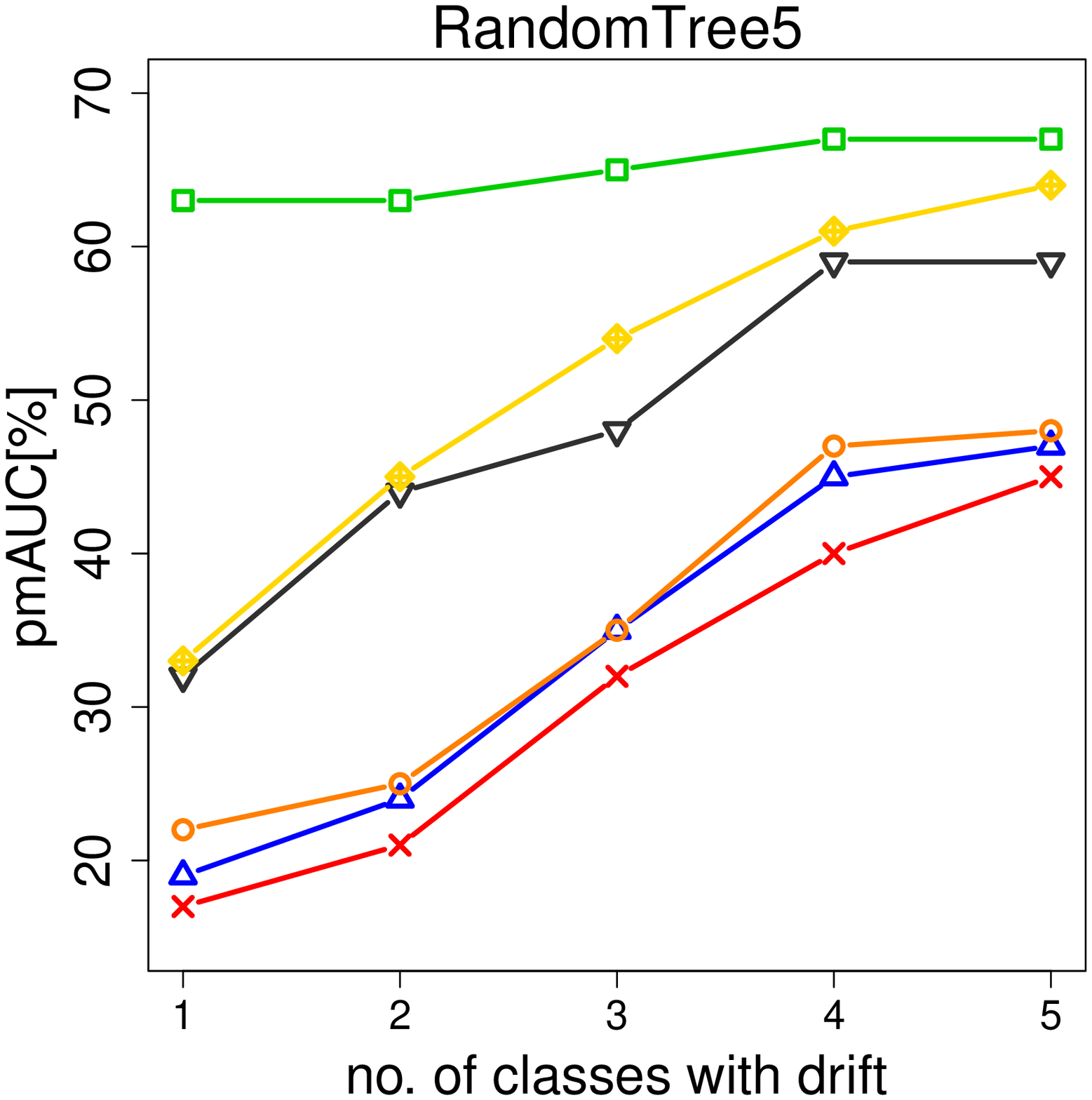}
			\includegraphics[width=0.32\linewidth]{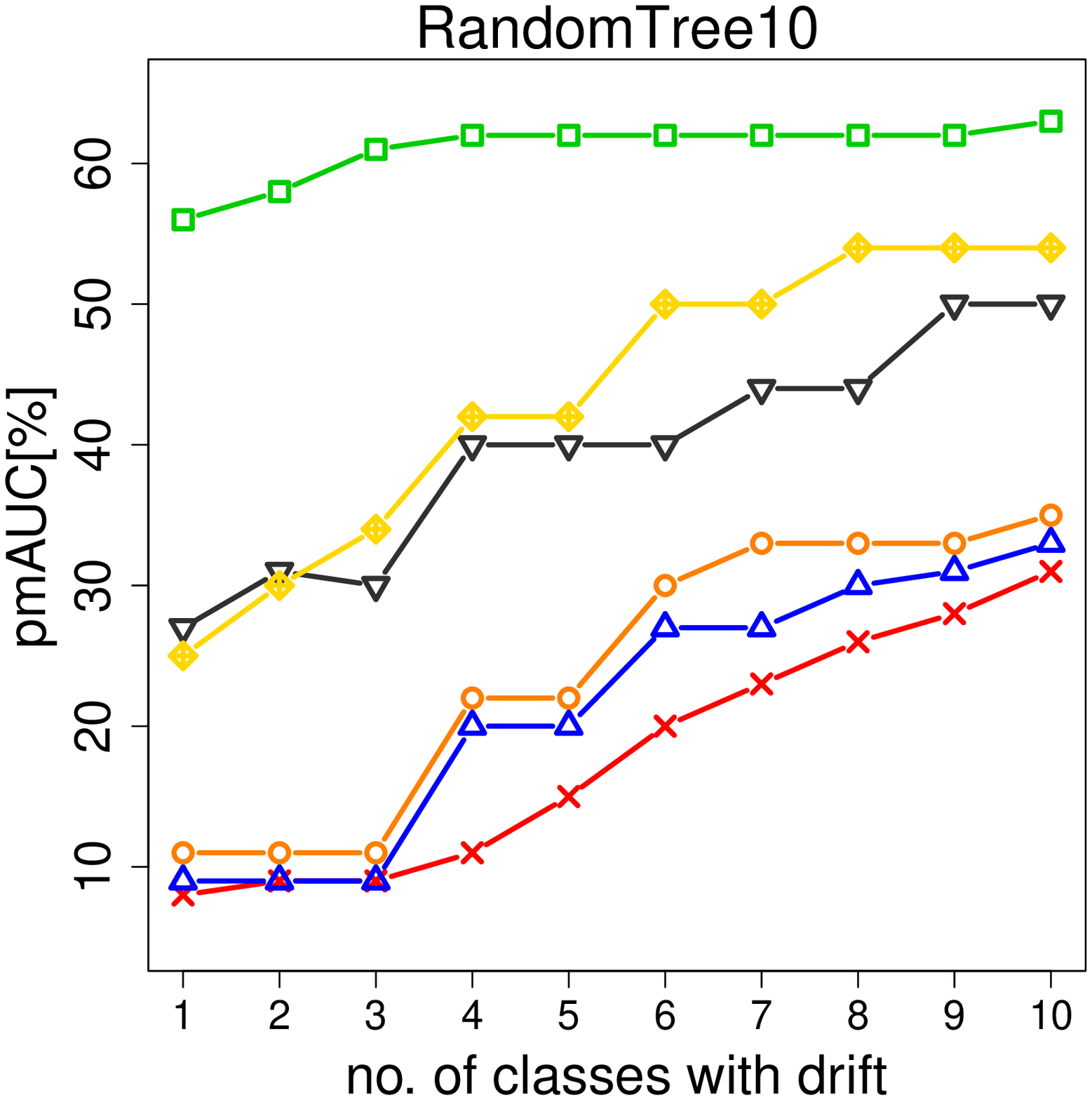}
			\includegraphics[width=0.32\linewidth]{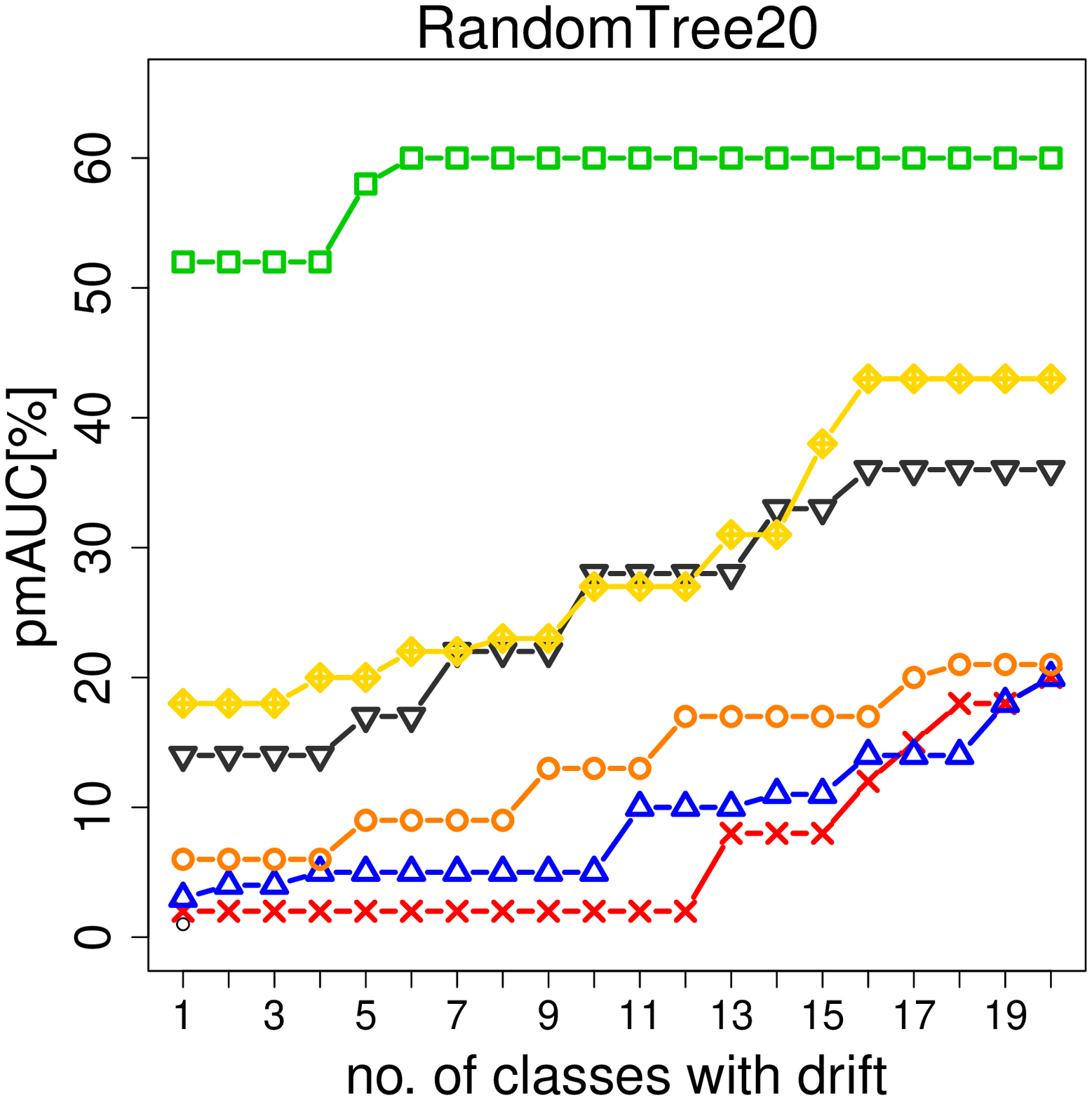}
	\caption{Relationship between pmAUC and the number of classes affected by the local drift for the artificial benchmarks. The lower the number of classes subject to concept drift, the more difficult its detection.}
	\label{fig:cla}
\end{figure}

This experiment was designed to understand if and how the examined drift detectors can handle the appearance of local concept drifts on top of changing imbalance ratios and class roles (see Sec. 3 -- Scenario 3 for more details). We carried this experiment only on artificial benchmarks, as they allowed us to directly inject concept drift into a selected number of classes. We evaluated how the performance of drift detectors change with the decrease of the number of classes being affected by the concept drift. For each of the 12 benchmark data streams, we created scenarios where from 1 to $M$ classes are being affected by the drift, the $M$ case standing for every single class in the stream being subject to the concept drift. Fig.~\ref{fig:cla} depicts the behavior of all the six drift detectors under various levels of the local concept drift for the pmAUC metric. We do not show plots for pmGM as they have very similar characteristics and would not provide any additional insights. Please note that the smaller number of classes that are subject to concept drift, the more difficult its detection becomes.

\smallskip 
\noindent \textbf{Comparison with standard drift detectors.} Unsurprisingly, standard detectors completely failed when facing the task of local drift detection. When the number of classes subject to concept drift dropped below 80\%, we could see significant drops in their pmAUC. When the number of affected classes dropped below 50\%, all three detectors started to completely ignore the presence of any drift. This crucially impacted the underlying classifier that lost any adaptation capabilities, as drift detectors were never signaling any change being present. Such results clearly support our earlier statement that standard drift detectors cannot handle local changes, as statistics they monitor relate to the entire stream, not specific classes. Furthermore, in the case of imbalanced multi-class drifting streams, the underlying bias toward the majority class had a strong impact on those statistics. This damaged the reactivity of those detectors to an even greater degree, as changes happening in minority classes were obscured by static properties of the majority class. 

\smallskip 
\noindent \textbf{Comparison with skew-insensitive drift detectors.} This experiment showed the weak side of the skew-insensitive drift detectors published so far. While they can display some robustness to changing class ratios and global concept drift, they did not perform significantly better than standard detectors when facing local drifts. For more than 90\% of classes being affected by drift, both PerfSim and DDM-OCI returned satisfactory performance. Their quality started degrading when less than 70\% of classes were being affected, reaching the lowest plateau for less than 30\% of classes being affected. This shows that despite the fact of monitoring some performance metrics for each class (like DDM-OCI monitors recall) they do not extract strong enough properties of those classes to properly detect local drifts. Only when the majority of classes become subject to concept drift those detectors can pick up local changes.

\smallskip 
\noindent \textbf{RBM-IM sensitivity to local drifts.} RBM-IM displayed an excellent sensitivity to local drifts, even when they affected only a single class. This observation holds for any data set, any imbalance ratio, and any total number of classes. This can be contributed to the effectiveness of the reconstruction error, used as a change detection metric, combined with storing compressed information about each class independently, and being able to compare reconstruction error for each class individually. This allows RBM-IM to detect local drifts that at a given moment affect any number of classes.  

\smallskip 
\noindent \textbf{Answer to RQ3:} RBM-IM is the only drift detector among the examined ones that can correctly detect local concept drifts, even when they affect only a single minority class. This allows to gain a better understanding of what is the exact nature of changes affecting the data stream and which classes should be more carefully analyzed to discover useful knowledge. This RBM-IM's capability of offering at the same time global and local concept drift detection is a crucial step towards explainable drift detection and gaining deeper insights into dynamics behind data streams, especially those imbalanced.

\subsection{Experiment 3: Robustness to changing imbalance ratio}

\begin{figure}[h]
	\centering
			\includegraphics[width=0.32\linewidth]{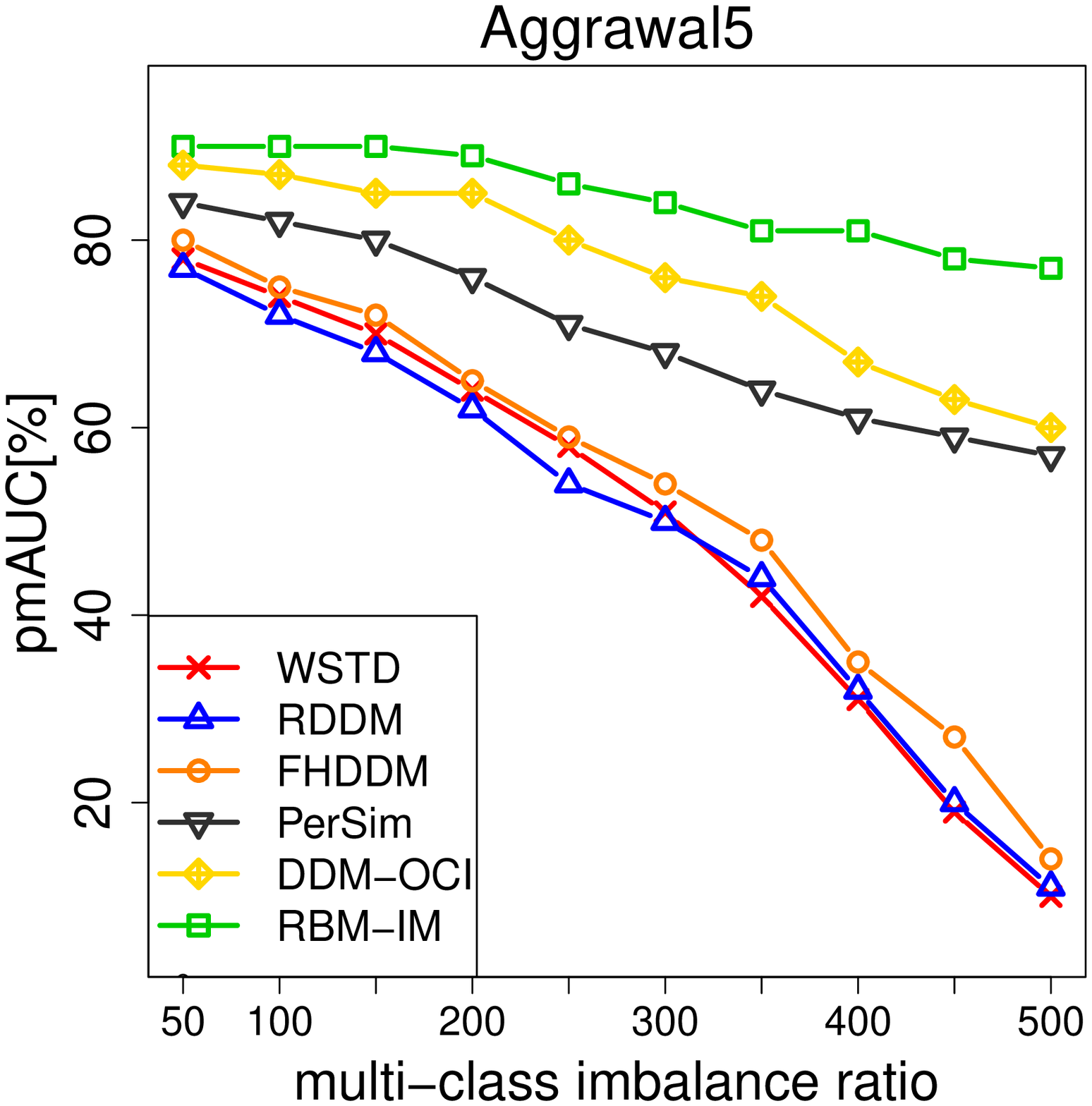}
			\includegraphics[width=0.32\linewidth]{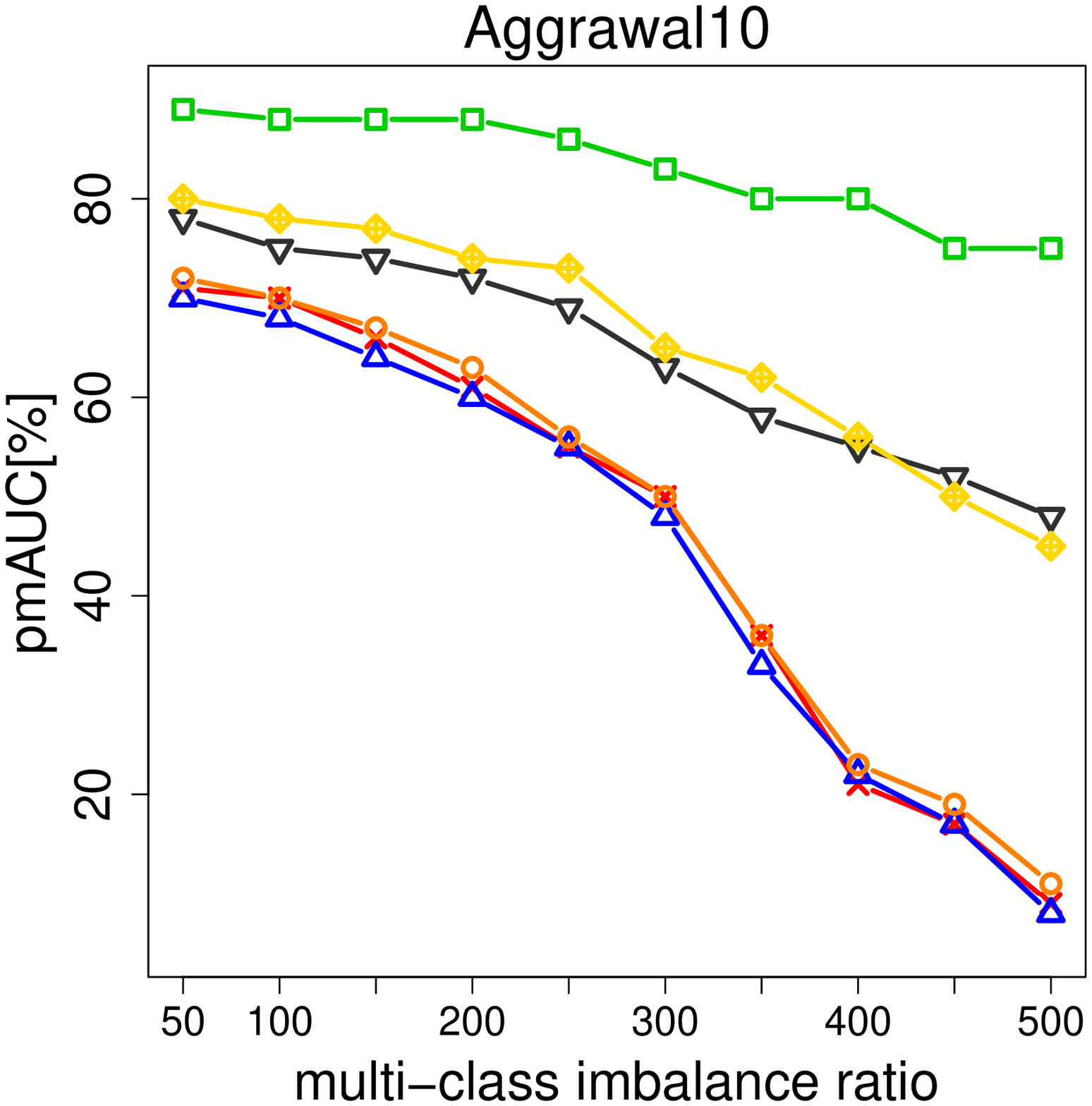}
			\includegraphics[width=0.32\linewidth]{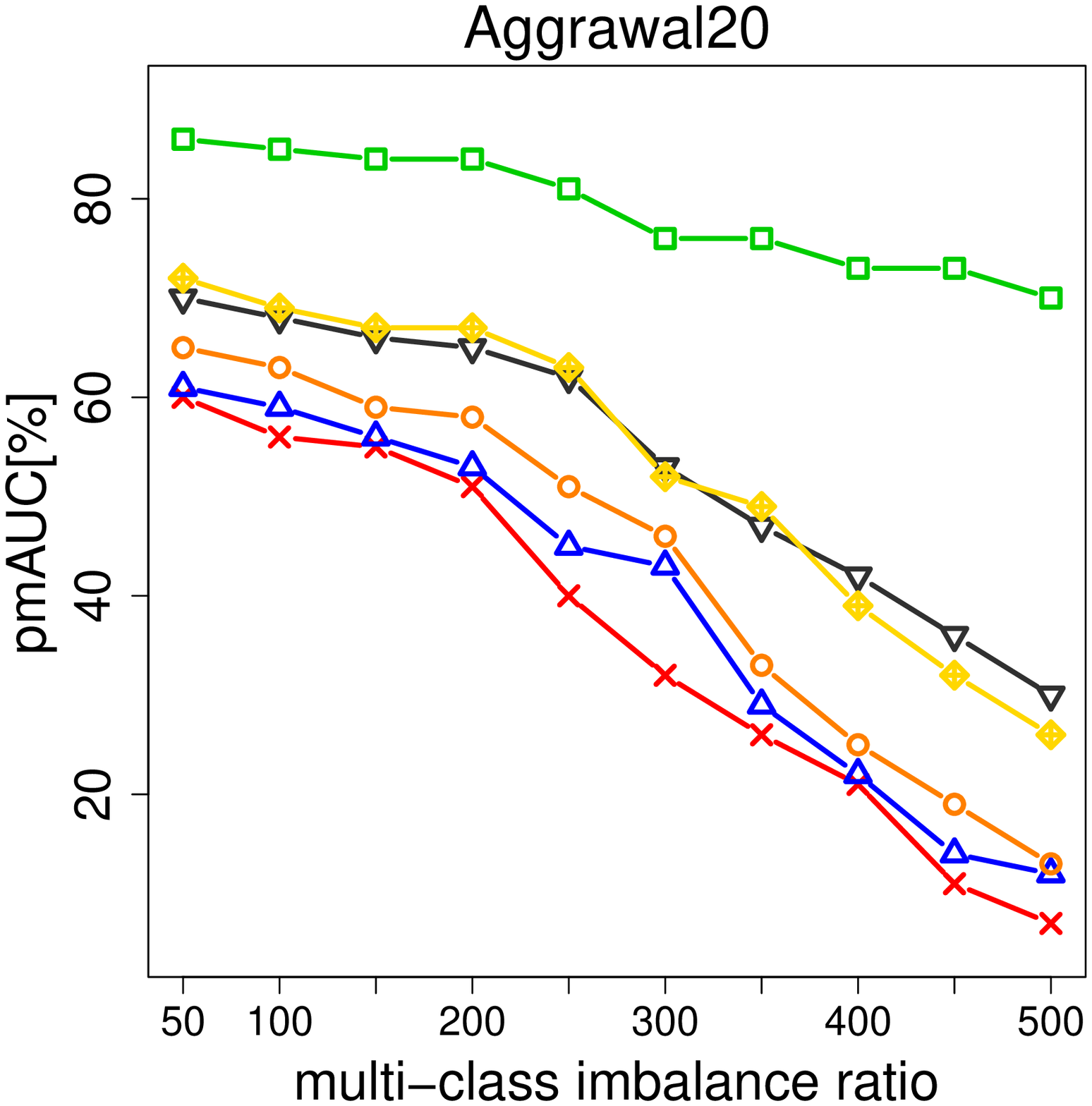}
			\includegraphics[width=0.32\linewidth]{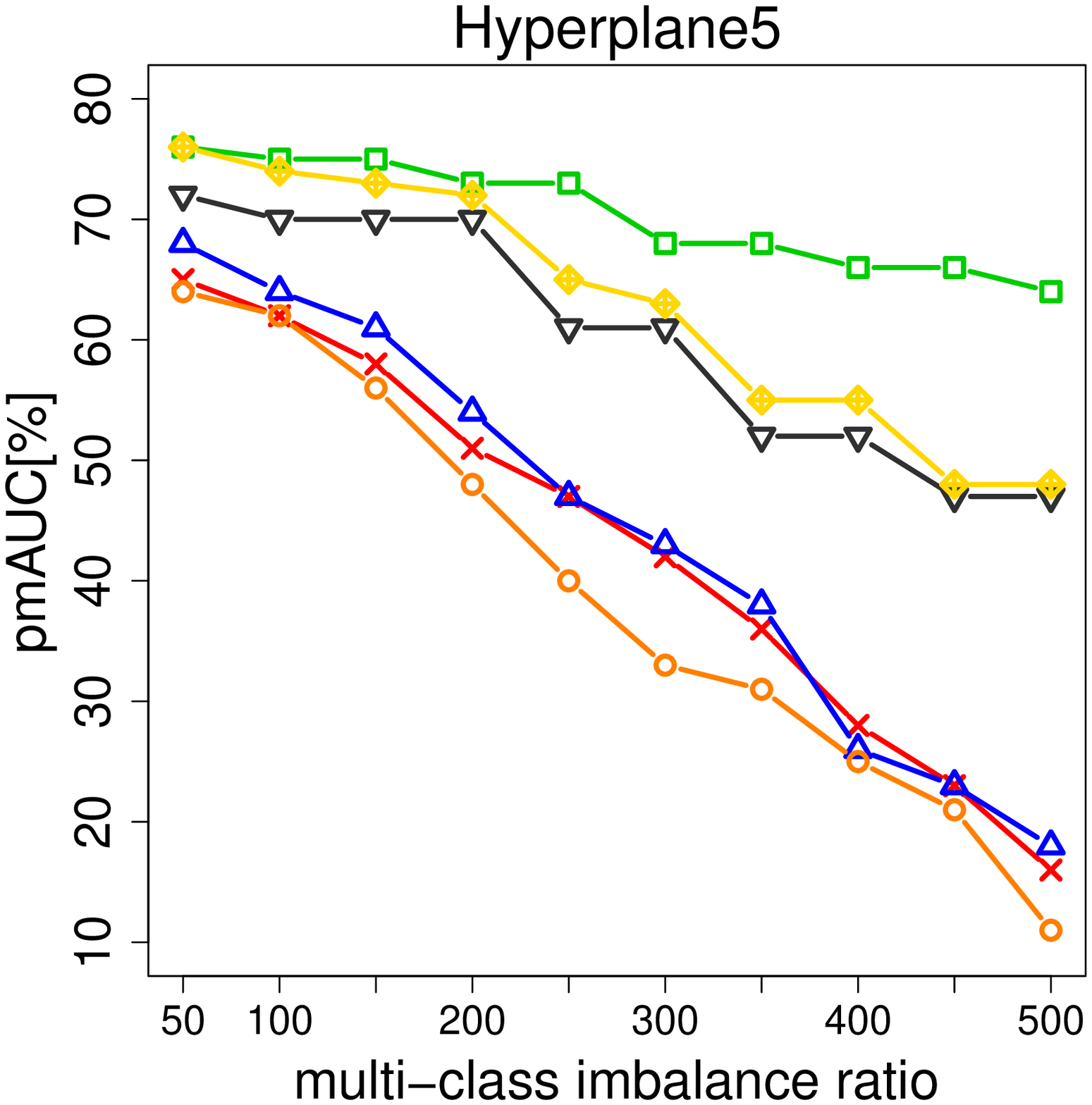}
			\includegraphics[width=0.32\linewidth]{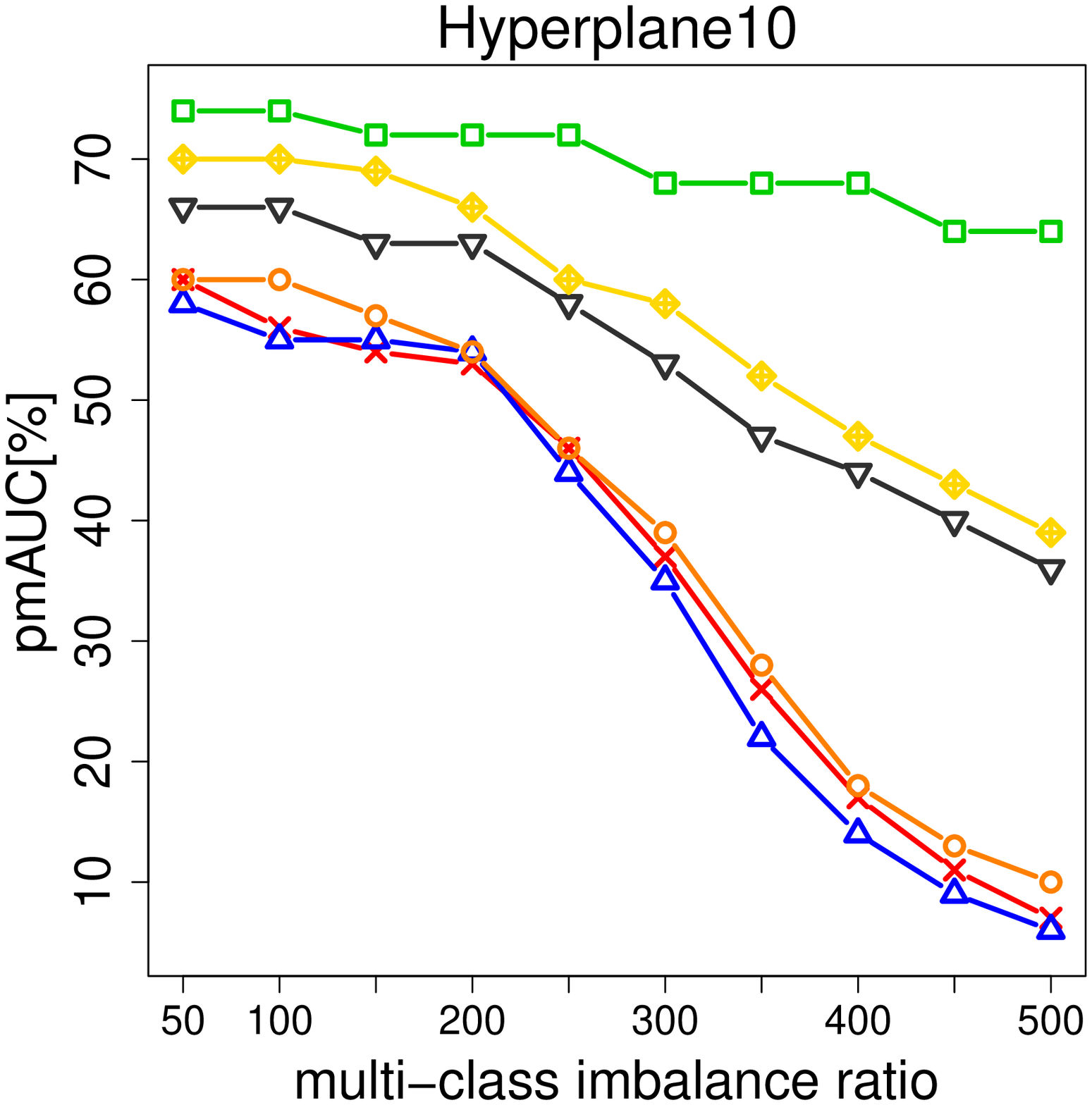}
			\includegraphics[width=0.32\linewidth]{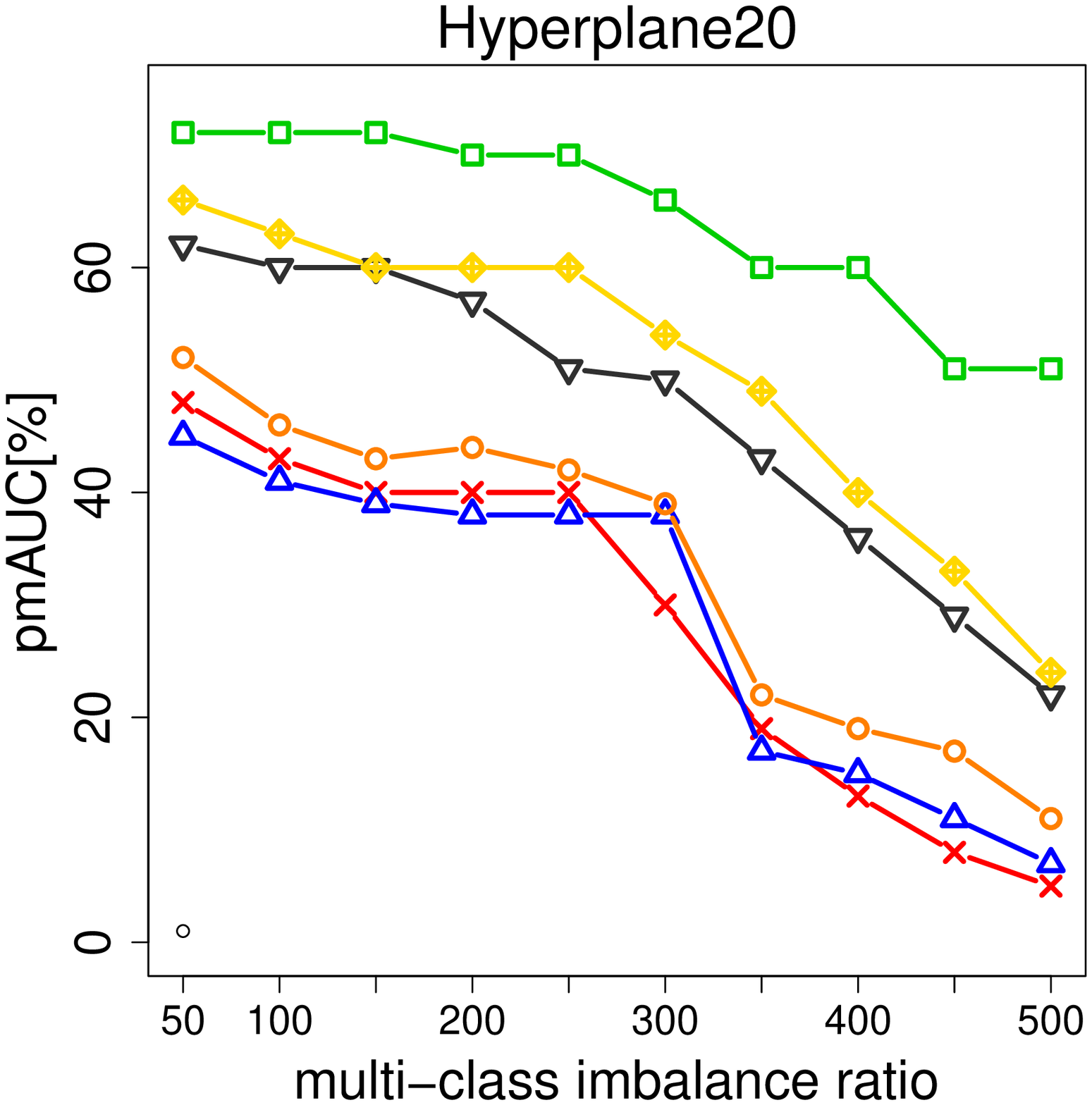}
			\includegraphics[width=0.32\linewidth]{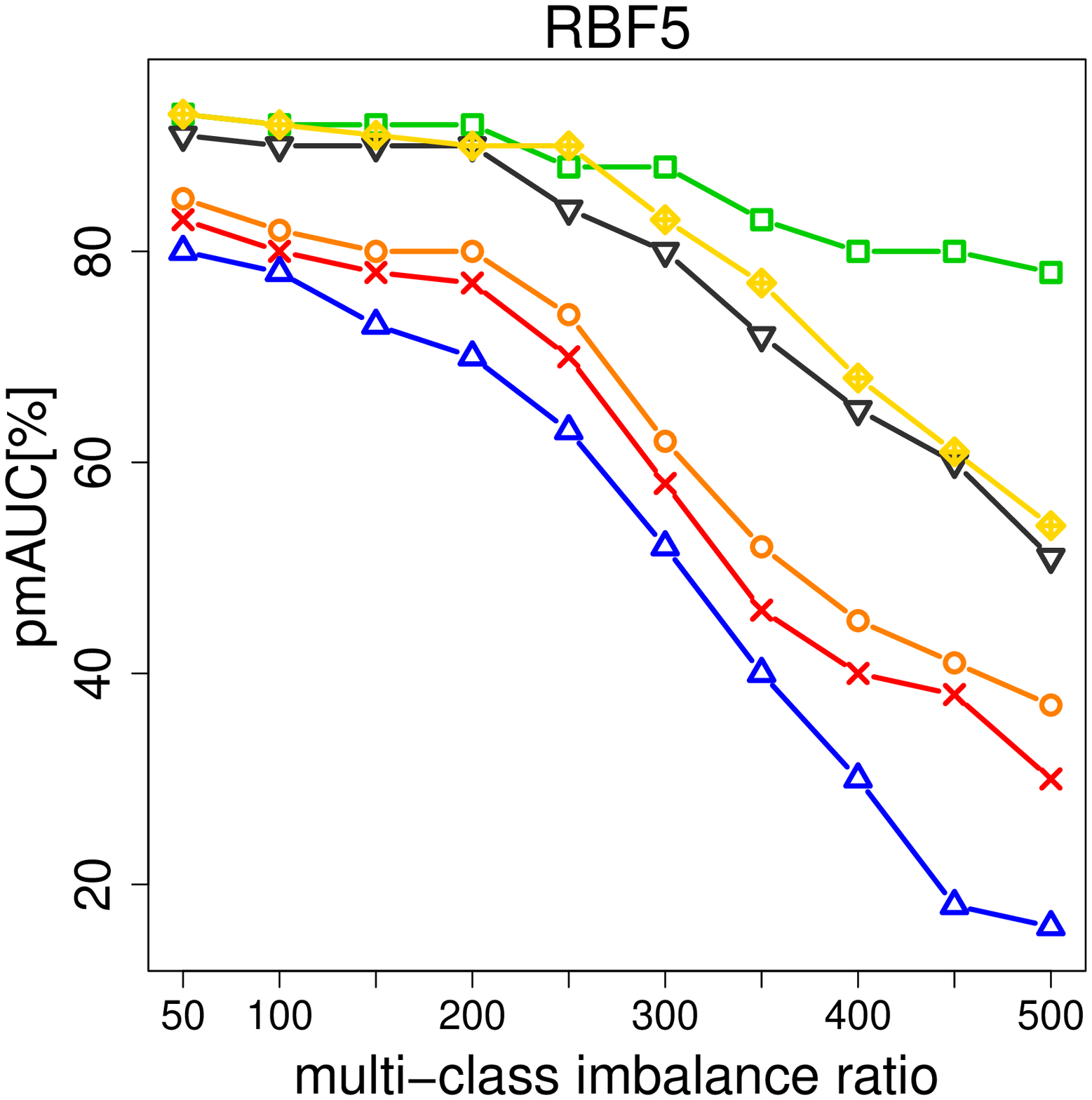}
			\includegraphics[width=0.32\linewidth]{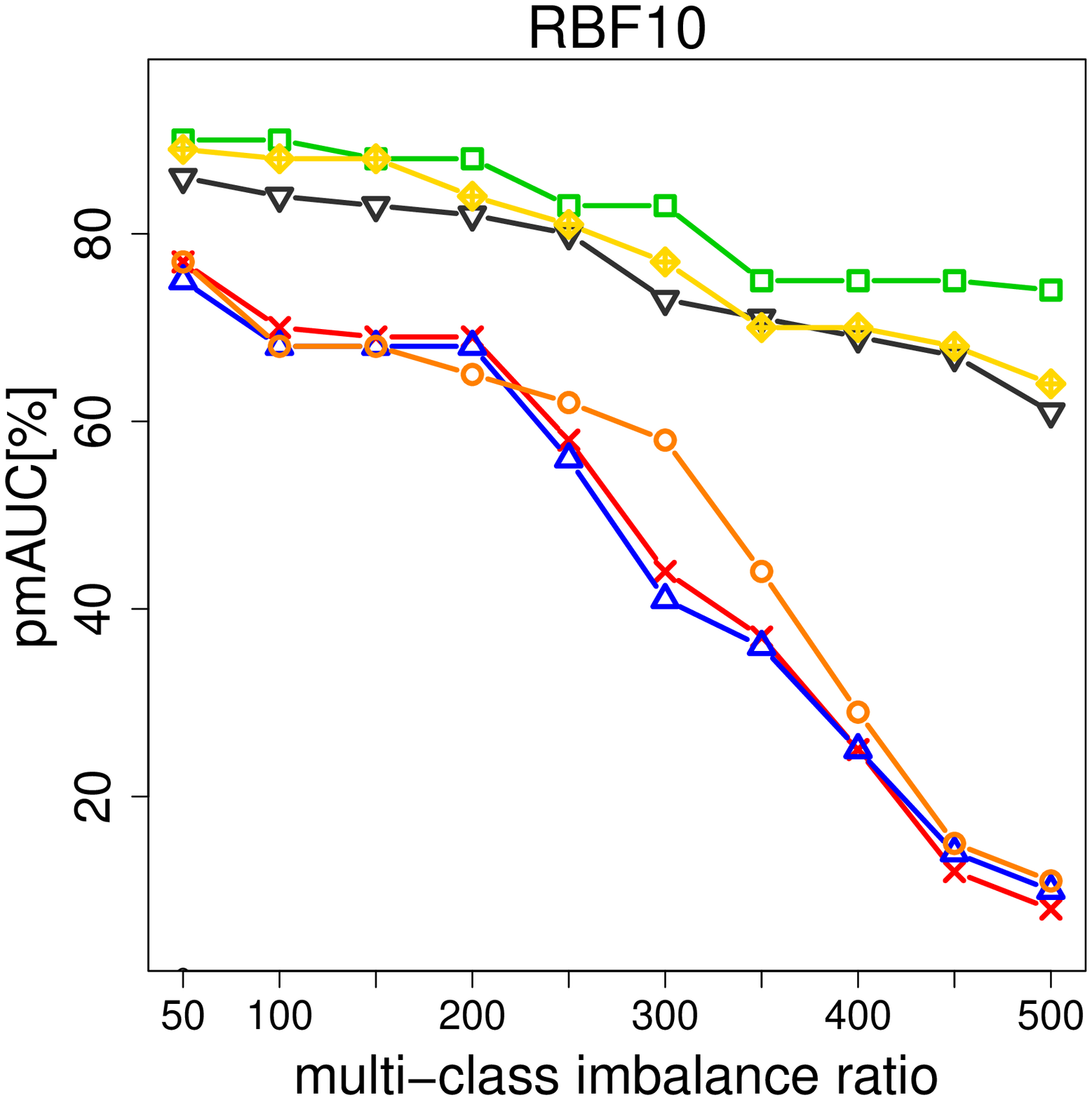}
			\includegraphics[width=0.32\linewidth]{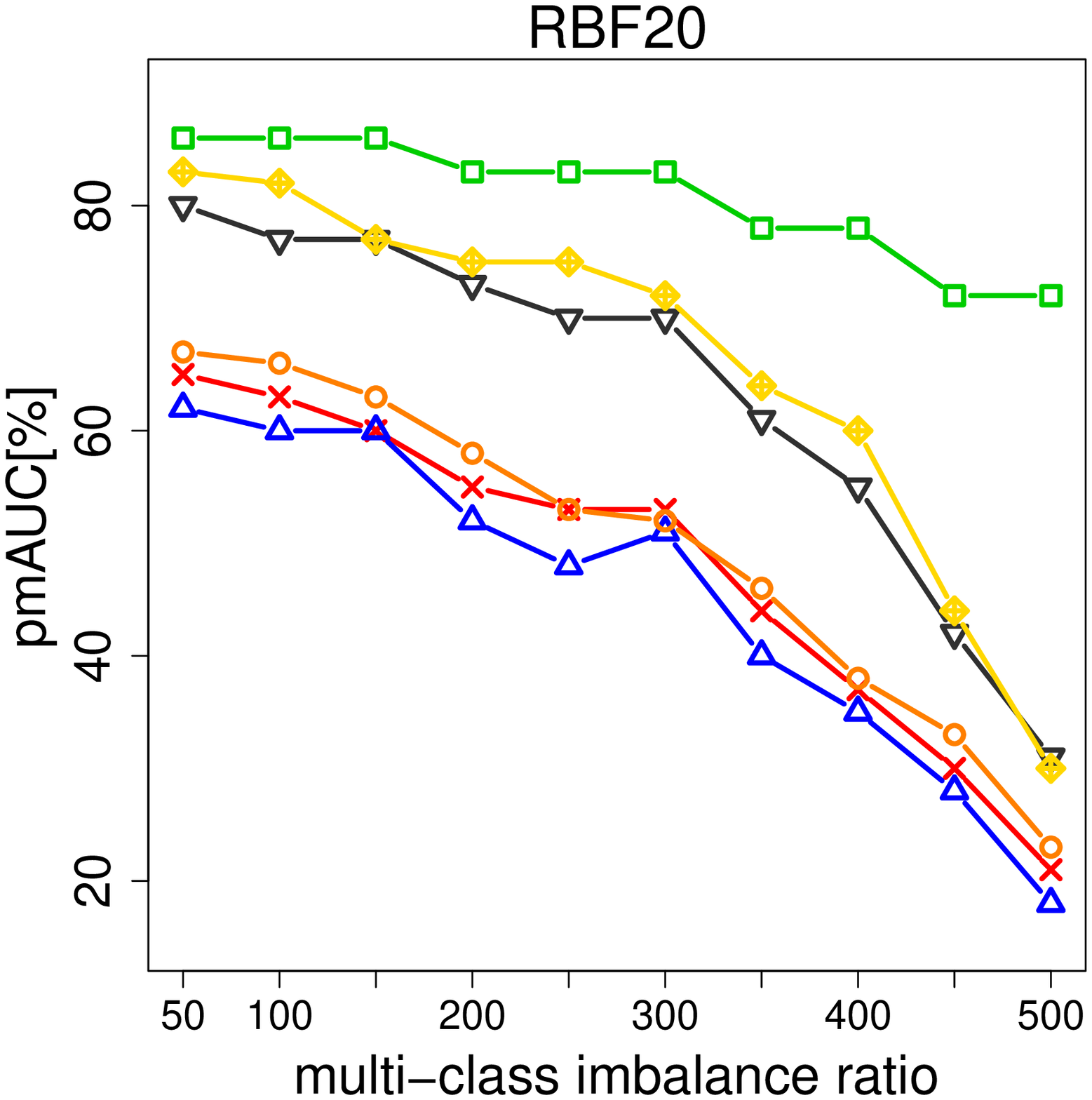}
			\includegraphics[width=0.32\linewidth]{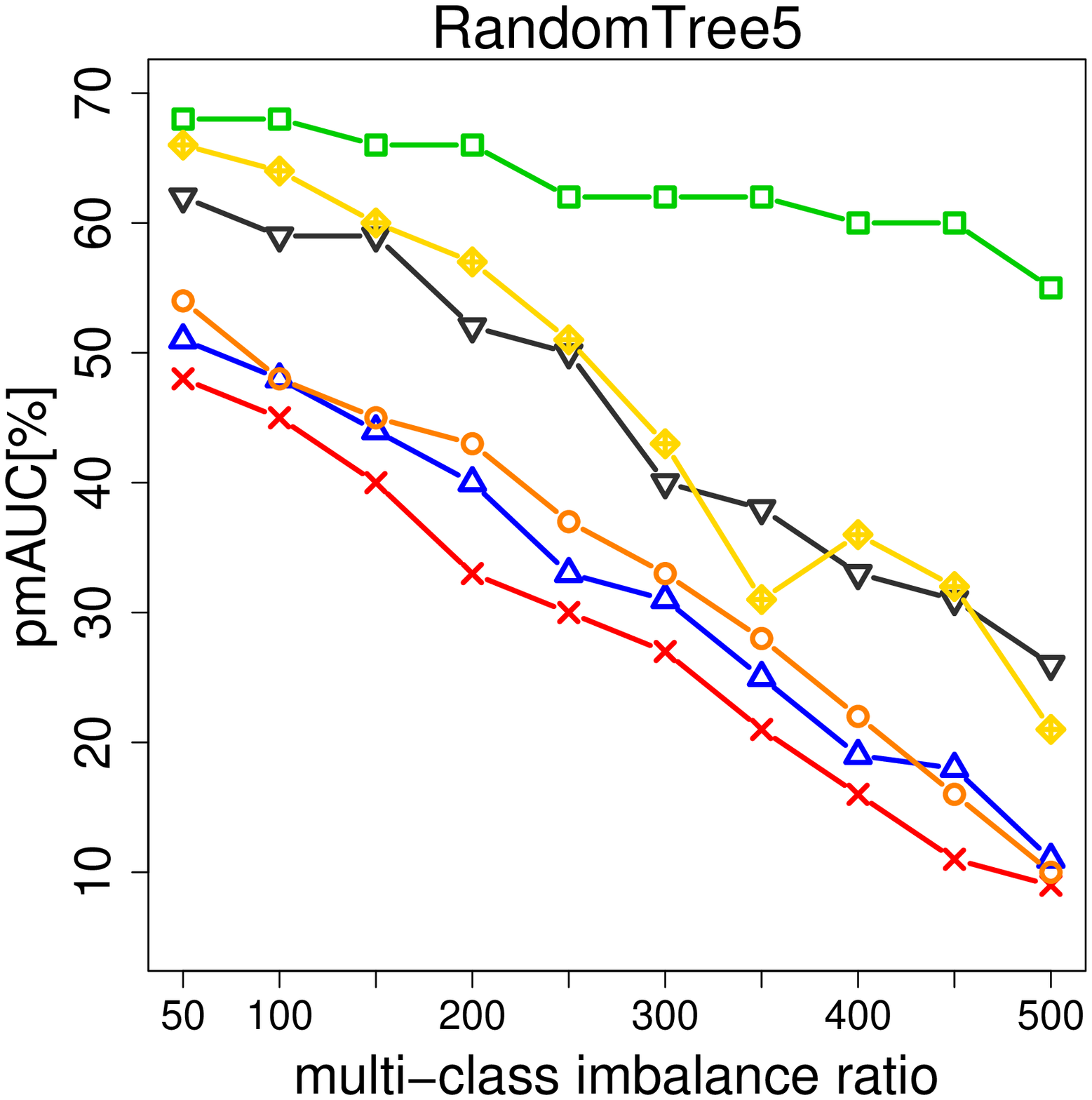}
			\includegraphics[width=0.32\linewidth]{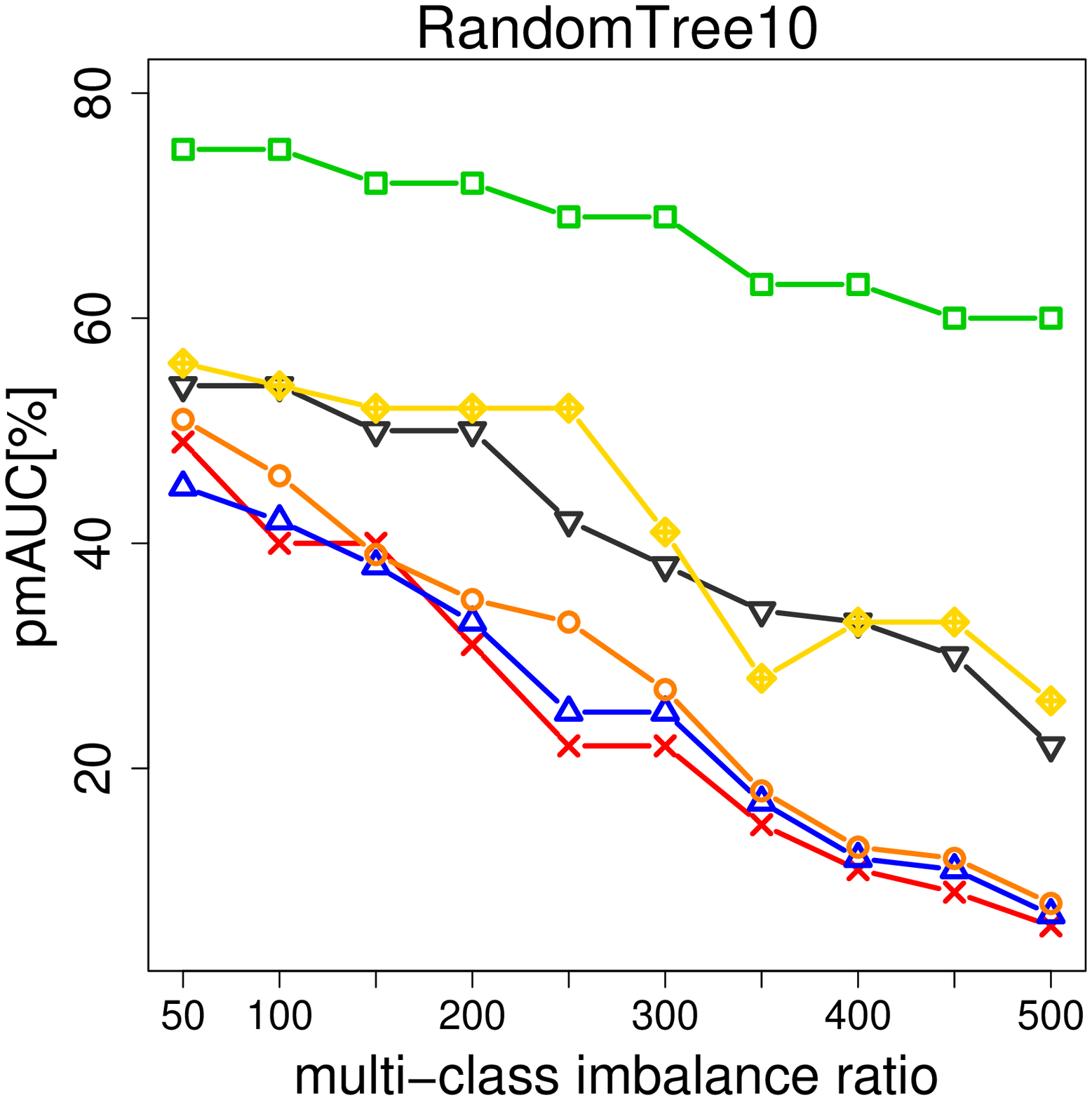}
			\includegraphics[width=0.32\linewidth]{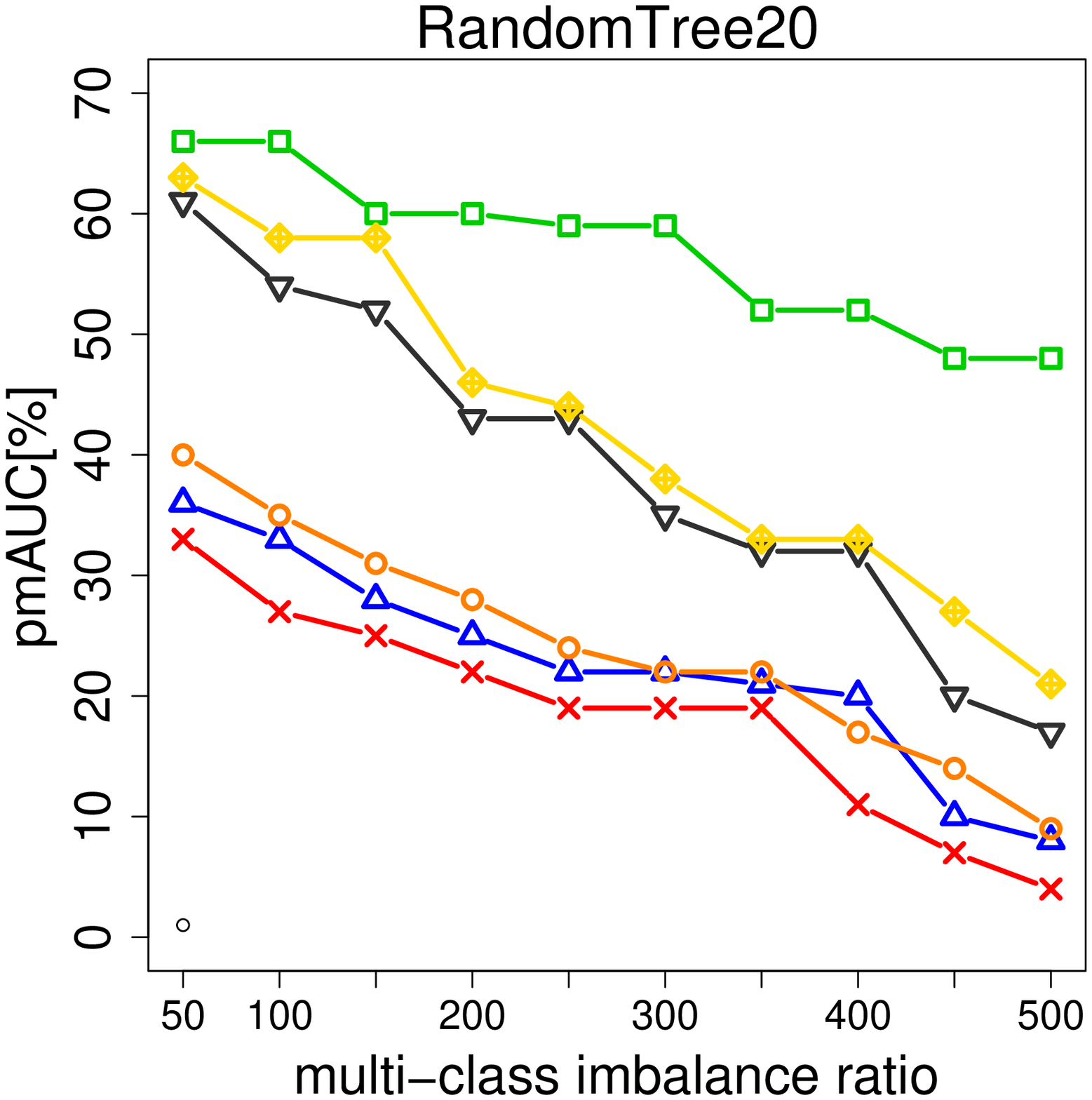}
	\caption{Relationship between pmAUC and changing imbalance ratio for the artificial benchmarks. The higher the imbalance ratio, the higher the disproportions among multiple classes.}
	\label{fig:ir}
\end{figure}

The third experiment was designed for evaluating the robustness of the examined drift detectors to changing imbalance ratio, especially for extremely imbalanced cases (IR $>$ 400). This will allow us to test the flexibility and trustworthiness of skew-insensitive mechanisms used in the detectors and to see how reliable they are. For each of 12 benchmark data streams, we created scenarios in which we generate varying imbalance ratios from 50 to 500. Fig.~\ref{fig:ir} depicts the behavior of the six drift detectors under various levels of class imbalance for the pmAUC metric. We do not show plots for pmGM as, analogously to the previous experiment, they have very similar characteristics.

\smallskip 
\noindent \textbf{Analyzing robustness to changing imbalance ratios.} As expected the standard drift detectors cannot handle any class imbalance ratios and do not return any acceptable results, omitting drift detection. This can be seen in the extremely poor performance of the underlying classifier that stopped being updated and could not handle new incoming concepts. Two reference skew-insensitive detectors maintain an acceptable robustness to small and medium imbalance ratios (IR $<$ 200), but start to critically fail with further increasing IR. At extreme levels of IR they performance becomes similar to standard detectors. This shows that none of the existing detectors can handle high imbalance ratios in multi-class data streams. RBM-IM offers excellent and stable robustness, filling the gap and providing a sought-after robust drift detection approach. We can contribute this to a combination of the used loss function and the ability of RBM-IM to continually learn from the stream. This is a massive advantage, as all other drift detectors are using some preset rules for deciding if the drift is present or not. RBM-IM can learn the current distribution in a skew-insensitive manner, making its drift detection much more accurate and not affected by the imbalance ratio.

\smallskip 
\noindent \textbf{Answer to RQ4:} RBM-IM offers excellent robustness to various levels of dynamic imbalance ratio in multi-class scenarios. Due to its trainable nature, RBM-IM is capable of quickly adapting to the current state of any stream and re-aligning its own structure regarding class ratios and class roles. This is the only drift detector displaying robustness to extremely high levels of class imbalance (IR $>$ 400).

\section{Lessons learned}
\label{sec:les}

Let us now present a short summary of insights and conclusions that were drawn from both the theoretical and experimental parts of this paper.

\smallskip
\noindent \textbf{Unified view on challenges in imbalanced multi-class data streams.} Continual learning from non-stationary and skewed multiple distributions is a challenging topic that requires more attention from the research community. It offers an excellent field for developing and evaluating novel learning algorithms, while calling for enhancing our models with various valuable robust characteristics. Three mutually complementary scenarios were identified by us, each dealing with different learning difficulties embedded in the nature of data. %One must remember that in multi-class imbalanced streams both distributions, imbalance ratios, and class role may change over time and our models must be capable of swift adaptation to such evolving learning scenario. 

\smallskip
\noindent \textbf{Advantages of trainable drift detector.} To the best of our knowledge, the existing state-of-the-art drift detectors are realized as external modules that track some properties of the stream and use them to decide if a drift should be detected or not. However, those models use static rules for determining the degree of change that constitutes drift presence. This significantly limits them in capturing unique properties of each concept and thus may negatively impact their reactivity to changes. We propose to use a trainable drift detector that can extract and store the most important characteristics of the current state of the stream and use them to make an informative and guided decision on deciding whether the underlying classifier should be retrained or not. 

\smallskip
\noindent \textbf{Handling global and local drifts.} Most of the works in drift detection focus on detecting global drifts that affect the entire stream. Detectors gather information from every single instance and use those statistics to make a decision. However, this makes them less sensitive to local drifts that affect only certain classes. The situation becomes even more challenging when combined with multi-class imbalanced distributions. Here, local drifts affecting the minority classes would go unnoticed, as gathered statistics will be biased towards the majority classes. This shows the importance of monitoring each individual class for local changes. %This also provides us with valuable insights into the nature of concept drifts and helps us understand dynamics of changes. RBM-IM stores information about each class independently in its model, allowing for precise drift detection even if it affects only a single minority class. 

\smallskip
\noindent \textbf{Impact of class imbalance on drift detection.} Not enough attention has been given to the interplay between the concept drift and class imbalance. We observed that imbalanced distributions will directly affect each drift detector in two possible ways: (i) enhancing the presence of small changes in the majority classes; and (ii) diminishing the importance of changes in the minority classes. The former problem is caused by statistics gathered from more abundant classes that will dominate the detector and thus may cause false alarms, as even small changes will be magnified by the sheer disproportion among classes. The latter problem is caused by the minority classes not contributing enough to the drift detector statistics and thus not being able to trigger it to cause an alarm. %We showed that by enhancing RBM-IM with a skew-insensitive loss function we are able to handle a high range of imbalance ratios in multi-class data streams.

\section{Conclusions and future works}
\label{sec:con}

In this work, we have discussed an important area of learning from multi-class imbalanced data streams under concept drift. We proposed a unifying taxonomy of challenges that may be encountered when learning from such data, and identified three realistic scenarios representing various types of learning difficulties. This was the first complete attempt to understand and organize challenges arising in this area of machine learning. We introduced RBM-IM, a novel and trainable drift detector for monitoring changes for continual learning from multi-class imbalanced data streams. Our research was motivated by an apparent lack of drift detection methods designed for skewed multi-class and evolving streams. We developed our drift detector on the basis of the Restricted Boltzmann Machine neural network with a skew-sensitivities loss function. 

%We used it to store compressed information about each class independently and use the reconstruction error over mini-batches of data to detect concept drift per class. This, combined with the loss function robust to imbalanced data, allowed RBM-IM to be highly sensitive and reactive to local concept drifts that affect only a small subset of minority classes.

In our future works, we plan to combine RBM-IM with techniques for handling underfitting (to make it applicable to small data streams), as well as make it robust to adversarial concept drifts that may be injected by a malicious party as a poisoning attack.

\bibliographystyle{IEEEtran}
\bibliography{refs}

\begin{thebibliography}{10}
\providecommand{\url}[1]{#1}
\csname url@samestyle\endcsname
\providecommand{\newblock}{\relax}
\providecommand{\bibinfo}[2]{#2}
\providecommand{\BIBentrySTDinterwordspacing}{\spaceskip=0pt\relax}
\providecommand{\BIBentryALTinterwordstretchfactor}{4}
\providecommand{\BIBentryALTinterwordspacing}{\spaceskip=\fontdimen2\font plus
\BIBentryALTinterwordstretchfactor\fontdimen3\font minus
  \fontdimen4\font\relax}
\providecommand{\BIBforeignlanguage}[2]{{%
\expandafter\ifx\csname l@#1\endcsname\relax
\typeout{** WARNING: IEEEtran.bst: No hyphenation pattern has been}%
\typeout{** loaded for the language `#1'. Using the pattern for}%
\typeout{** the default language instead.}%
\else
\language=\csname l@#1\endcsname
\fi
#2}}
\providecommand{\BIBdecl}{\relax}
\BIBdecl

\bibitem{Parisi:2019}
G.~I. Parisi, R.~Kemker, J.~L. Part, C.~Kanan, and S.~Wermter, ``Continual
  lifelong learning with neural networks: {A} review,'' \emph{Neural Networks},
  vol. 113, pp. 54--71, 2019.

\bibitem{Sahoo:2018}
D.~Sahoo, Q.~Pham, J.~Lu, and S.~C.~H. Hoi, ``Online deep learning: Learning
  deep neural networks on the fly,'' in \emph{Proceedings of the Twenty-Seventh
  International Joint Conference on Artificial Intelligence, {IJCAI} 2018, July
  13-19, 2018, Stockholm, Sweden}, J.~Lang, Ed.\hskip 1em plus 0.5em minus
  0.4em\relax ijcai.org, 2018, pp. 2660--2666.

\bibitem{Ditzler:2015}
G.~Ditzler, M.~Roveri, C.~Alippi, and R.~Polikar, ``Learning in nonstationary
  environments: {A} survey,'' \emph{{IEEE} Comput. Intell. Mag.}, vol.~10,
  no.~4, pp. 12--25, 2015.

\bibitem{Krawczyk:2017}
B.~Krawczyk, L.~L. Minku, J.~Gama, J.~Stefanowski, and M.~Wozniak, ``{Ensemble
  learning for data stream analysis: {A} survey},'' \emph{Inf. Fusion},
  vol.~37, pp. 132--156, 2017.

\bibitem{Chandra:2018}
S.~Chandra, A.~Haque, H.~Tao, J.~Liu, L.~Khan, and C.~C. Aggarwal, ``Ensemble
  direct density ratio estimation for multistream classification,'' in
  \emph{34th {IEEE} International Conference on Data Engineering, {ICDE} 2018,
  Paris, France, April 16-19, 2018}.\hskip 1em plus 0.5em minus 0.4em\relax
  {IEEE} Computer Society, 2018, pp. 1364--1367.

\bibitem{Wang:2019}
Z.~Wang, Z.~Kong, S.~Chandra, H.~Tao, and L.~Khan, ``Robust high dimensional
  stream classification with novel class detection,'' in \emph{35th {IEEE}
  International Conference on Data Engineering, {ICDE} 2019, Macao, China,
  April 8-11, 2019}.\hskip 1em plus 0.5em minus 0.4em\relax {IEEE}, 2019, pp.
  1418--1429.

\bibitem{Krawczyk:2016}
B.~Krawczyk, ``Learning from imbalanced data: open challenges and future
  directions,'' \emph{Prog. Artif. Intell.}, vol.~5, no.~4, pp. 221--232, 2016.

\bibitem{Wang:2018}
S.~Wang, L.~L. Minku, and X.~Yao, ``A systematic study of online class
  imbalance learning with concept drift,'' \emph{{IEEE} Trans. Neural Networks
  Learn. Syst.}, vol.~29, no.~10, pp. 4802--4821, 2018.

\bibitem{Fernandez:2018}
\BIBentryALTinterwordspacing
A.~Fern{\'{a}}ndez, S.~Garc{\'{\i}}a, M.~Galar, R.~C. Prati, B.~Krawczyk, and
  F.~Herrera, \emph{Learning from Imbalanced Data Sets}.\hskip 1em plus 0.5em
  minus 0.4em\relax Springer, 2018. [Online]. Available:
  \url{https://doi.org/10.1007/978-3-319-98074-4}
\BIBentrySTDinterwordspacing

\bibitem{Masegosa:2020}
A.~R. Masegosa, A.~M. Mart{\'{\i}}nez, D.~Ramos{-}L{\'{o}}pez, H.~Langseth,
  T.~D. Nielsen, and A.~Salmer{\'{o}}n, ``{Analyzing concept drift: {A} case
  study in the financial sector},'' \emph{Intell. Data Anal.}, vol.~24, no.~3,
  pp. 665--688, 2020.

\bibitem{Lu:2019}
J.~Lu, A.~Liu, F.~Dong, F.~Gu, J.~Gama, and G.~Zhang, ``{Learning under Concept
  Drift: {A} Review},'' \emph{{IEEE} Transactions on Knowledge and Data
  Engineering}, vol.~31, no.~12, pp. 2346--2363, 2019.

\bibitem{Goldenberg:2020}
I.~Goldenberg and G.~I. Webb, ``{PCA-based drift and shift quantification
  framework for multidimensional data},'' \emph{Knowledge and Information
  Systems}, vol.~62, no.~7, pp. 2835--2854, 2020.

\bibitem{Oliveira:2019}
G.~H. F.~M. Oliveira, L.~L. Minku, and A.~L.~I. Oliveira, ``{{GMM-VRD:} {A}
  Gaussian Mixture Model for Dealing With Virtual and Real Concept Drifts},''
  in \emph{International Joint Conference on Neural Networks, {IJCNN} 2019
  Budapest, Hungary, July 14-19, 2019}.\hskip 1em plus 0.5em minus 0.4em\relax
  {IEEE}, 2019, pp. 1--8.

\bibitem{Gama:2006}
J.~Gama and G.~Castillo, ``{Learning with Local Drift Detection},'' in
  \emph{Advanced Data Mining and Applications, Second International Conference,
  {ADMA} 2006, Xi'an, China, August 14-16, 2006, Proceedings}, ser. Lecture
  Notes in Computer Science, vol. 4093.\hskip 1em plus 0.5em minus 0.4em\relax
  Springer, 2006, pp. 42--55.

\bibitem{Barros:2018}
R.~S.~M. de~Barros and S.~G.~T. de~Carvalho~Santos, ``{A large-scale comparison
  of concept drift detectors},'' \emph{Information Sciences}, vol. 451-452, pp.
  348--370, 2018.

\bibitem{Gama:2004}
J.~Gama, P.~Medas, G.~Castillo, and P.~P. Rodrigues, ``Learning with drift
  detection,'' in \emph{Advances in Artificial Intelligence - {SBIA} 2004, 17th
  Brazilian Symposium on Artificial Intelligence, S{\~{a}}o Luis,
  Maranh{\~{a}}o, Brazil, September 29 - October 1, 2004, Proceedings}, ser.
  Lecture Notes in Computer Science, A.~L.~C. Bazzan and S.~Labidi, Eds., vol.
  3171.\hskip 1em plus 0.5em minus 0.4em\relax Springer, 2004, pp. 286--295.

\bibitem{Garcia:2006}
M.~Baena-García, J.~Campo-Avila, R.~Fidalgo-Merino, A.~Bifet, R.~Gavald, and
  R.~Morales-Bueno, ``Early drift detection method,'' in \emph{4th ECML PKDD
  International Workshop on Knowledge Discovery from Data Streams}, 2006, p.
  77–86.

\bibitem{Barros:2017}
R.~S.~M. de~Barros, D.~R. de~Lima~Cabral, P.~M.~G. Jr., and S.~G.~T.
  de~Carvalho~Santos, ``{RDDM:} reactive drift detection method,'' \emph{Expert
  Syst. Appl.}, vol.~90, pp. 344--355, 2017.

\bibitem{Bifet:2007}
A.~Bifet and R.~Gavald{\`{a}}, ``{Learning from Time-Changing Data with
  Adaptive Windowing},'' in \emph{Proceedings of the Seventh {SIAM}
  International Conference on Data Mining, April 26-28, 2007, Minneapolis,
  Minnesota, {USA}}.\hskip 1em plus 0.5em minus 0.4em\relax {SIAM}, 2007, pp.
  443--448.

\bibitem{Blanco:2015}
I.~I.~F. Blanco, J.~del Campo{-}{\'{A}}vila, G.~Ramos{-}Jim{\'{e}}nez, R.~M.
  Bueno, A.~A.~O. D{\'{\i}}az, and Y.~C. Mota, ``Online and non-parametric
  drift detection methods based on hoeffding's bounds,'' \emph{{IEEE} Trans.
  Knowl. Data Eng.}, vol.~27, no.~3, pp. 810--823, 2015.

\bibitem{Pesaranghader:2016}
A.~Pesaranghader and H.~L. Viktor, ``Fast hoeffding drift detection method for
  evolving data streams,'' in \emph{Machine Learning and Knowledge Discovery in
  Databases - European Conference, {ECML} {PKDD} 2016, Riva del Garda, Italy,
  September 19-23, 2016, Proceedings, Part {II}}, ser. Lecture Notes in
  Computer Science, vol. 9852.\hskip 1em plus 0.5em minus 0.4em\relax Springer,
  2016, pp. 96--111.

\bibitem{Barros:2018w}
R.~S.~M. de~Barros, J.~I.~G. Hidalgo, and D.~R. de~Lima~Cabral, ``Wilcoxon rank
  sum test drift detector,'' \emph{Neurocomputing}, vol. 275, pp. 1954--1963,
  2018.

\bibitem{Cano:2020}
A.~Cano and B.~Krawczyk, ``{Kappa Updated Ensemble for drifting data stream
  mining},'' \emph{Machine Learning}, vol. 109, no.~1, pp. 175--218, 2020.

\bibitem{Antwi:2012}
D.~K. Antwi, H.~L. Viktor, and N.~Japkowicz, ``The perfsim algorithm for
  concept drift detection in imbalanced data,'' in \emph{12th {IEEE}
  International Conference on Data Mining Workshops, {ICDM} Workshops,
  Brussels, Belgium, December 10, 2012}.\hskip 1em plus 0.5em minus 0.4em\relax
  {IEEE} Computer Society, 2012, pp. 619--628.

\bibitem{Wang:2020}
S.~Wang and L.~L. Minku, ``{AUC} estimation and concept drift detection for
  imbalanced data streams with multiple classes,'' in \emph{2020 International
  Joint Conference on Neural Networks, {IJCNN} 2020, Glasgow, United Kingdom,
  July 19-24, 2020}.\hskip 1em plus 0.5em minus 0.4em\relax {IEEE}, 2020, pp.
  1--8.

\bibitem{Saadallah:2019}
A.~Saadallah, L.~Moreira{-}Matias, R.~Sousa, J.~Khiari, E.~Jenelius, and
  J.~Gama, ``{BRIGHT} - drift-aware demand predictions for taxi networks
  (extended abstract),'' in \emph{35th {IEEE} International Conference on Data
  Engineering, {ICDE} 2019, Macao, China, April 8-11, 2019}.\hskip 1em plus
  0.5em minus 0.4em\relax {IEEE}, 2019, pp. 2145--2146.

\bibitem{Ramasamy:2020}
S.~Ramasamy, A.~Ambikapathi, and K.~Rajaraman, ``Online {RBM:} growing
  restricted boltzmann machine on the fly for unsupervised representation,''
  \emph{Appl. Soft Comput.}, vol.~92, p. 106278, 2020.

\bibitem{Cui:2019}
Y.~Cui, M.~Jia, T.~Lin, Y.~Song, and S.~J. Belongie, ``Class-balanced loss
  based on effective number of samples,'' in \emph{{IEEE} Conference on
  Computer Vision and Pattern Recognition, {CVPR} 2019, Long Beach, CA, USA,
  June 16-20, 2019}.\hskip 1em plus 0.5em minus 0.4em\relax Computer Vision
  Foundation / {IEEE}, 2019, pp. 9268--9277.

\bibitem{Sun:2008}
X.~Sun, ``Assessing nonlinear granger causality from multivariate time
  series,'' in \emph{Machine Learning and Knowledge Discovery in Databases,
  European Conference, {ECML/PKDD} 2008, Antwerp, Belgium, September 15-19,
  2008, Proceedings, Part {II}}, ser. Lecture Notes in Computer Science, vol.
  5212.\hskip 1em plus 0.5em minus 0.4em\relax Springer, 2008, pp. 440--455.

\bibitem{Mahjoub:2020}
C.~Mahjoub, J.~Bellanger, A.~Kachouri, and R.~L. Bouquin{-}Jeann{\`{e}}s, ``On
  the performance of temporal granger causality measurements on time series: a
  comparative study,'' \emph{Signal Image Video Process.}, vol.~14, no.~5, pp.
  955--963, 2020.

\bibitem{Bifet:2010moa}
A.~Bifet, G.~Holmes, R.~Kirkby, and B.~Pfahringer, ``{{MOA:} Massive Online
  Analysis},'' \emph{Journal of Machine Learning Research}, vol.~11, pp.
  1601--1604, 2010.

\bibitem{Veloso:2018}
B.~Veloso, J.~Gama, and B.~Malheiro, ``Self hyper-parameter tuning for data
  streams,'' in \emph{Discovery Science - 21st International Conference, {DS}
  2018, Limassol, Cyprus, October 29-31, 2018, Proceedings}, ser. Lecture Notes
  in Computer Science, vol. 11198.\hskip 1em plus 0.5em minus 0.4em\relax
  Springer, 2018, pp. 241--255.

\bibitem{Krawczyk:2017ecml}
B.~Krawczyk and P.~Skryjomski, ``Cost-sensitive perceptron decision trees for
  imbalanced drifting data streams,'' in \emph{Machine Learning and Knowledge
  Discovery in Databases - European Conference, {ECML} {PKDD} 2017, Skopje,
  Macedonia, September 18-22, 2017, Proceedings, Part {II}}, ser. Lecture Notes
  in Computer Science, vol. 10535.\hskip 1em plus 0.5em minus 0.4em\relax
  Springer, 2017, pp. 512--527.

\bibitem{Korycki:2020}
L.~Korycki and B.~Krawczyk, ``Online oversampling for sparsely labeled
  imbalanced and non-stationary data streams,'' in \emph{2020 International
  Joint Conference on Neural Networks, {IJCNN} 2020, Glasgow, United Kingdom,
  July 19-24, 2020}.\hskip 1em plus 0.5em minus 0.4em\relax {IEEE}, 2020, pp.
  1--8.

\bibitem{Benavoli:2017}
A.~Benavoli, G.~Corani, J.~Demsar, and M.~Zaffalon, ``Time for a change: a
  tutorial for comparing multiple classifiers through bayesian analysis,''
  \emph{J. Mach. Learn. Res.}, vol.~18, pp. 77:1--77:36, 2017.

\end{thebibliography}

\end{document}